\documentclass[letterpaper]{article} 
\usepackage{aaai2027}  
\usepackage[hyphens]{url}  
\usepackage{graphicx} 
\def\UrlFont{\rm}  
\usepackage{natbib}  
\usepackage{caption} 
\usepackage{algorithm}
\usepackage{algorithmic}
\usepackage{amsmath}
\usepackage{textcomp}
\usepackage{xcolor}
\usepackage{graphicx}
\usepackage[table]{xcolor}
\usepackage{multirow}
\usepackage{amssymb}
\usepackage{adjustbox}
\definecolor{headergray}{RGB}{230,230,230}   
\definecolor{oursgray}{RGB}{220,220,235}     

\usepackage{booktabs}
\usepackage{makecell}
\usepackage{array}
\usepackage{newfloat}
\usepackage{listings}
\DeclareCaptionStyle{ruled}{labelfont=normalfont,labelsep=colon,strut=off} 
\floatstyle{ruled}
\newfloat{listing}{tb}{lst}{}
\floatname{listing}{Listing}

\usepackage{booktabs}

\title{AnchorVLA: Bridging Discrete Decisions and Continuous Trajectories for Vision-Language-Action Planning}

\author{
    Qi Liu\textsuperscript{\rm 1,2,3}\equalcontrib,
    Yabei Li\textsuperscript{\rm 4}\equalcontrib,
    Heng Zhang\textsuperscript{\rm 5},
    Lei He\textsuperscript{\rm 1,2}\corresponding
}

\affiliations{
    \textsuperscript{\rm 1}School of Vehicle and Mobility, Tsinghua University, Beijing 100084, China\\
    \textsuperscript{\rm 2}State Key Laboratory of Intelligent Green Vehicle and Mobility, Tsinghua University, Beijing 100084, China\\
    \textsuperscript{\rm 3}School of Computer Science and Engineering, Southeast University, Nanjing 210096, China\\
    \textsuperscript{\rm 4}Meituan Inc.\\
    \textsuperscript{\rm 5}Dongfeng Motor Corporation\\
}

\nocopyright
\begin{document}

\maketitle

\begin{figure}[t]
    \centering
    \includegraphics[width=0.95\columnwidth]{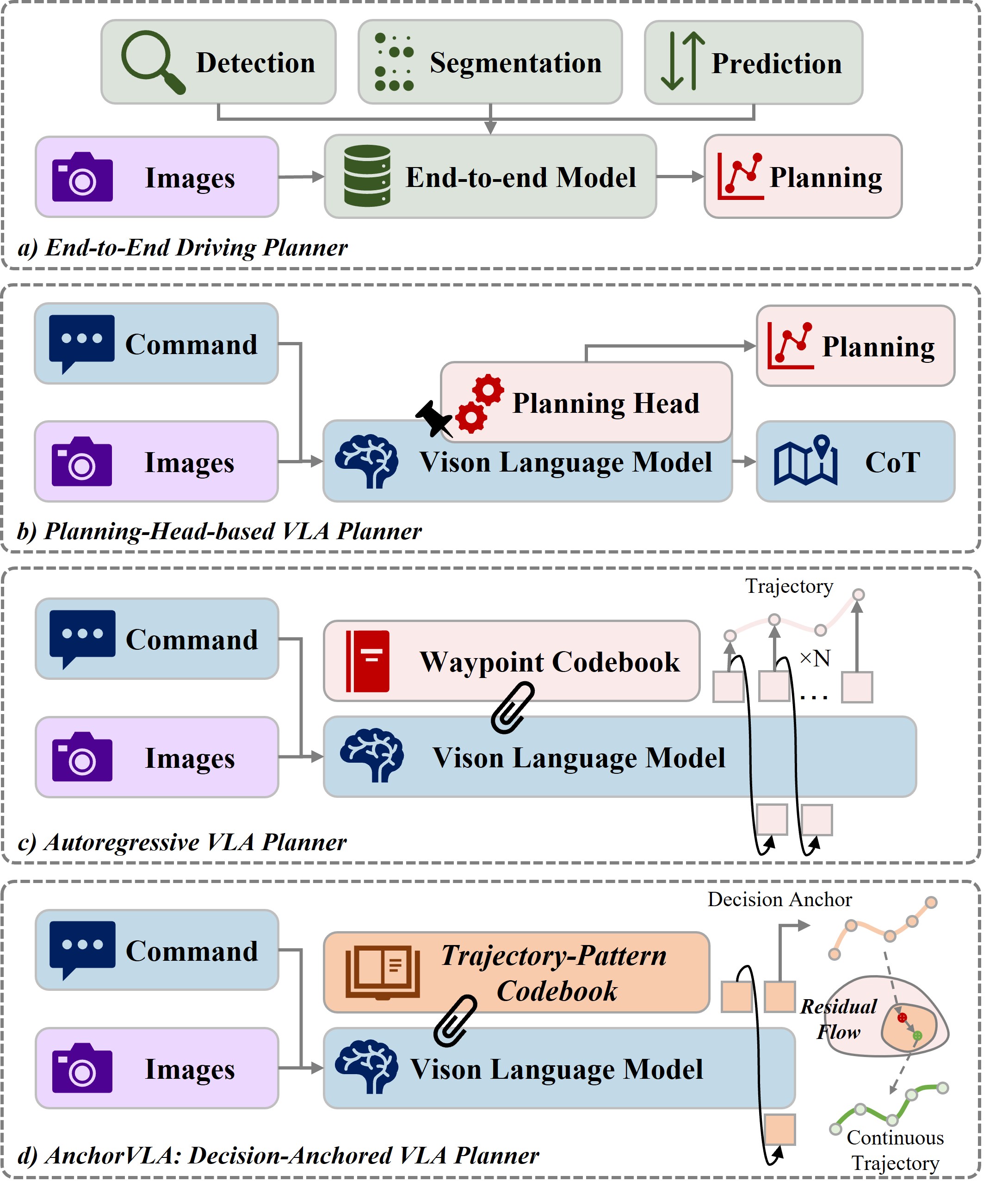}
    \caption{
    Comparison of autonomous driving planning paradigms.
    (a) End-to-end planners unify perception, prediction, mapping, and
    planning within a single network to reduce information loss across modular
    interfaces.
    (b) Planning-head-based VLA planners use VLM/LLM features with an
    independent planning head.
    (c) Autoregressive VLA planners discretize trajectories into waypoint
    tokens and generate them sequentially.
    (d) AnchorVLA represents high-level VLA decisions as trajectory-pattern
    anchors and generates continuous trajectories in the corresponding
    anchor-defined residual space.
    }
    \label{fig:intro}
\end{figure}

\begin{abstract}
Vision-Language-Action (VLA) models have been introduced into driving planning to improve long-tail generalization, commonsense reasoning, high-level semantic understanding, and explainability. However, existing VLA planners mainly follow planning-head-based trajectory prediction or full-trajectory autoregressive generation: the former only weakly constrains continuous trajectory generation with VLA reasoning, while the latter requires long sequences of low-information-density coordinate tokens, resulting in inefficient inference. To address these limitations, we propose \textbf{AnchorVLA}, a hierarchical decision-anchored VLA planning framework that uses trajectory-pattern anchors as an explicit interface between high-level VLA reasoning and continuous trajectory execution. Specifically, \textbf{Decision-as-Anchor Representation} (DAAR) represents behavior-level driving decisions with anchor tokens, each encoding an entire local motion pattern rather than a single coordinate point. \textbf{Decision-Anchored Residual Flow} (DARF) then generates fine-grained continuous trajectories in the selected anchor-defined residual space, capturing multi-modal execution refinements after high-level decision making. By reasoning over compact behavior-level trajectory-pattern anchors instead of autoregressively generating waypoint sequences, AnchorVLA preserves LLM-based decision making while improving inference efficiency, establishing an explicit behavior-level interface between language reasoning and continuous action generation. Experiments on the Bench2Drive closed-loop benchmark show that AnchorVLA achieves a state-of-the-art Success Rate of $77.28$ and a competitive Driving Score of $89.92$.

\end{abstract}


\section{Introduction}

Trajectory planning aims to generate safe and rational future trajectories for the ego vehicle based on the surrounding driving scene, taking into account factors such as feasibility, comfort, and efficiency. End-to-end planning has become a mainstream paradigm in autonomous driving due to its ability to reduce error accumulation from hand-crafted intermediate modules and benefit from data-driven scaling. Existing end-to-end planners can be roughly divided into conventional learning-based planners and Vision-Language-Action (VLA) planners. As shown in Figure~\ref{fig:intro}(a), conventional planners directly regress future trajectories from multimodal sensor inputs~\cite{chen2024vadv2,chitta2022transfuser,hu2022st,hu2023planning,jiang2023vad,li2024hydra,liao2025diffusiondrive,sun2025sparsedrive,weng2024drive,xing2025goalflow,zheng2024genad}, but provide limited semantic reasoning and behavior-level decision-making ability. To address this, recent VLA planners introduce LLMs/VLMs into end-to-end driving planning and mainly follow two paradigms: planning-head-based trajectory prediction and autoregressive trajectory generation. As shown in Figure~\ref{fig:intro}(b), planning-head-based methods~\citep{fu2025orion,hwang2024emma,renz2025simlingo} use LLM/VLM-encoded features to drive an independent planning head, where the LLM/VLM mainly acts as a conditional encoder. As shown in Figure~\ref{fig:intro}(c), autoregressive methods~\citep{zhou2025autovla,wang2026unifying} discretize trajectories into tokens and generate them sequentially, which better involves the LLM in planning.

We argue that these paradigms suffer from a fundamental abstraction gap between high-level language reasoning and low-level continuous trajectory execution. Planning-head-based methods enable fast continuous generation, but their trajectory space is only weakly constrained by VLA reasoning, making local maneuvers difficult to consistently align with high-level decisions. Autoregressive methods involve LLMs more directly by generating trajectory tokens in a unified language-action space. However, when trajectories are tokenized at the coordinate or waypoint level, each token only represents a low-level geometric point with limited semantic meaning. Such low-information-density action tokens lead to long sequences, discretization error accumulation, and weak semantic correspondence with scene context and language instructions. Therefore, coordinate-level autoregressive generation does not fully exploit the LLM's strength in discrete abstraction and semantic decision making.

To address this challenge, we propose \textbf{AnchorVLA}, a hierarchical \textbf{Decision-Anchored Vision-Language-Action Planning} formulation, as illustrated in Figure~\ref{fig:intro}(d). The key idea is to raise the action representation from low-level coordinate tokens to high-level trajectory-pattern anchors. Each anchor represents a complete local motion pattern and can be viewed as a compact behavior-level decision, such as lane keeping, braking, yielding, turning, or overtaking. In this way, the LLM reasons over high-density behavior-level action abstractions rather than long waypoint-token sequences, enabling more efficient language-action modeling through an explicit interface between language reasoning and action selection.

AnchorVLA instantiates this idea with two complementary designs. First, \textbf{Decision-as-Anchor Representation} (DAAR) uses a trajectory-pattern codebook to represent behavior-level driving decisions selected through high-level VLA reasoning. Unlike waypoint-level codebooks that discretize individual coordinates or trajectory points, the trajectory-pattern codebook encodes complete local motion patterns, allowing each selected anchor to serve as a high-density behavior-level token. Second, given the selected anchor, \textbf{Decision-Anchored Residual Flow} (DARF) models continuous trajectory generation in the anchor-defined residual space, where the final trajectory is represented as the selected anchor plus a continuous residual. This follows the decision-then-execution nature of real driving: after determining an overall maneuver, a driver still needs to adapt the detailed trajectory according to surrounding agents, road geometry, and dynamic scene changes. Since multiple feasible refinements may exist under the same high-level decision, DARF uses flow matching to model multi-modal continuous residual refinements within the selected anchor-defined residual space. Together, DAAR and DARF provide a structured interface that directly couples high-level decision selection with continuous trajectory execution while retaining the flexibility of continuous generation.

Our main contributions are summarized as follows:
First, we propose \textbf{AnchorVLA}, a hierarchical VLA planning framework that introduces trajectory-pattern anchors as an explicit interface between high-level reasoning and continuous trajectory execution.
Second, we develop \textbf{Decision-as-Anchor Representation} (DAAR) for compact behavior-level decision modeling and \textbf{Decision-Anchored Residual Flow} (DARF) for multi-modal continuous trajectory refinement in the anchor-defined residual space.
Third, AnchorVLA achieves a state-of-the-art Success Rate of $77.28$ and a competitive Driving Score of $89.92$ on the Bench2Drive closed-loop benchmark.

\section{Related Work}

\subsection{End-to-End Autonomous Driving Planning}

End-to-end autonomous driving planning directly predicts future trajectories
or control signals from sensor inputs. TCP-traj~\cite{wu2022trajectory},
UniAD~\cite{hu2023planning}, VAD~\cite{jiang2023vad},
TransFuser++~\cite{chitta2022transfuser}, and
DriveTransformer~\cite{jia2025drivetransformer} improve closed-loop planning
through joint task modeling, structured scene representations, multimodal
fusion, or temporal reasoning. Recent works further explore probabilistic
trajectory generation: GenAD~\cite{zheng2024genad} adopts generative modeling,
VADv2~\cite{chen2024vadv2} constructs a planning vocabulary, while
DiffusionDrive~\cite{liao2025diffusiondrive},
BridgeDrive~\cite{liu2025bridgedrive}, and
GoalFlow~\cite{xing2025goalflow} employ diffusion, bridge, or flow-based
generation. Trajectory anchors have also been studied in multimodal
forecasting~\cite{chai2019multipathmultipleprobabilisticanchor, varadarajan2021multipathefficientinformationfusion}. In contrast,
AnchorVLA uses trajectory-pattern anchors as an explicit interface between
discrete VLA decisions and continuous executable trajectories.

\subsection{Vision-Language-Action Driving Planning}

Recent VLA driving methods introduce VLMs/LLMs into autonomous driving to enhance scene understanding, reasoning, and language-conditioned planning. Some early works mainly use language as an intermediate representation for driving explanation or decision reasoning~\cite{xu2024drivegpt4,sima2024drivelm,wang2023drivemlm,hwang2024emma}. More recent VLA planners connect visual-language reasoning with action generation. A common paradigm is to use an independent planning head to generate continuous trajectories~\cite{zhou2026opendrivevla,yang2025drivemoe,li2025recogdrive,li2025drivevla}. DiffVLA~\cite{jiang2025diffvla} further combines VLM guidance with a hybrid sparse-dense diffusion policy for multimodal driving behavior generation. ORION~\cite{fu2025orion} integrates a VLM architecture, a generative planner, and planning objectives to bridge reasoning and action spaces. SimLingo~\cite{renz2025simlingo} focuses on aligning linguistic comprehension with driving actions and achieves strong closed-loop performance on Bench2Drive. However, in these methods, LLM/VLM outputs are often used as conditions, guidance signals, or intermediate representations for downstream planners, and the high-level language decision may not explicitly define the continuous trajectory generation space.

Another line of work directly tokenizes trajectories or actions and lets LLMs generate driving actions autoregressively. AutoVLA~\cite{zhou2025autovla} discretizes spatial coordinates into coordinate tokens and autoregressively generates driving actions, while LinkVLA~\cite{wang2026unifying} quantizes waypoints or spatial coordinates into discrete action tokens to unify language and action understanding and generation. These methods bring action generation into the LLM token space, but their action tokens are mainly constructed at the coordinate or waypoint level. Different from these waypoint- or coordinate-level tokenization methods, AnchorVLA tokenizes the trajectory-pattern space, where each anchor token corresponds to a complete local motion pattern. Therefore, the autoregressive component is used for compact behavior-level decision making, while the final trajectory remains continuous through residual generation in the selected anchor-defined space.

\section{Method}

\subsection{AnchorVLA Overview}

Given the current scene observation, navigation information, and language instruction as the context \(x\), the model predicts a future trajectory \(\tau\). Instead of modeling the trajectory distribution \(p(\tau\mid x)\) as a monolithic continuous prediction problem, AnchorVLA decomposes trajectory generation into two stages: behavior-level decision making over trajectory-pattern anchors and fine-grained continuous execution within the selected anchor-defined residual space.

Specifically, we construct a trajectory-pattern codebook
\(\mathcal{A}=\{a_k\}_{k=1}^{K}\), where each anchor \(a_k\) represents a complete local motion pattern. Under \textbf{Decision-as-Anchor Representation} (DAAR), the anchor-decision model predicts an anchor distribution \(p(a_k\mid x)\) from the context \(x\). Each selected anchor serves as a compact behavior-level decision, such as lane keeping, braking, yielding, turning, or overtaking, with subsequent continuous generation performed within its anchor-defined residual space. Given an anchor \(a_k\), the final trajectory is represented as \(\tau=a_k+r_k\), where \(r_k\) denotes the continuous residual in the coordinate system centered at \(a_k\). Therefore, the global trajectory distribution can be decomposed as
\begin{equation}
p(\tau \mid x)
=
\sum_{k=1}^{K}
p(a_k\mid x)\,
p_r(\tau-a_k\mid x,a_k).
\label{eq:anchor_decomposition}
\end{equation}

High-level VLA reasoning first selects behavior-level anchors, and low-level continuous generation is then performed within the corresponding anchor-defined residual spaces. The following sections introduce how trajectory-pattern anchors are predicted under DAAR and how executable continuous trajectories are generated using \textbf{Decision-Anchored Residual Flow} (DARF).

\subsection{VLA Backbone}

\begin{figure*}[t]
    \centering
    \includegraphics[width=0.85\textwidth]{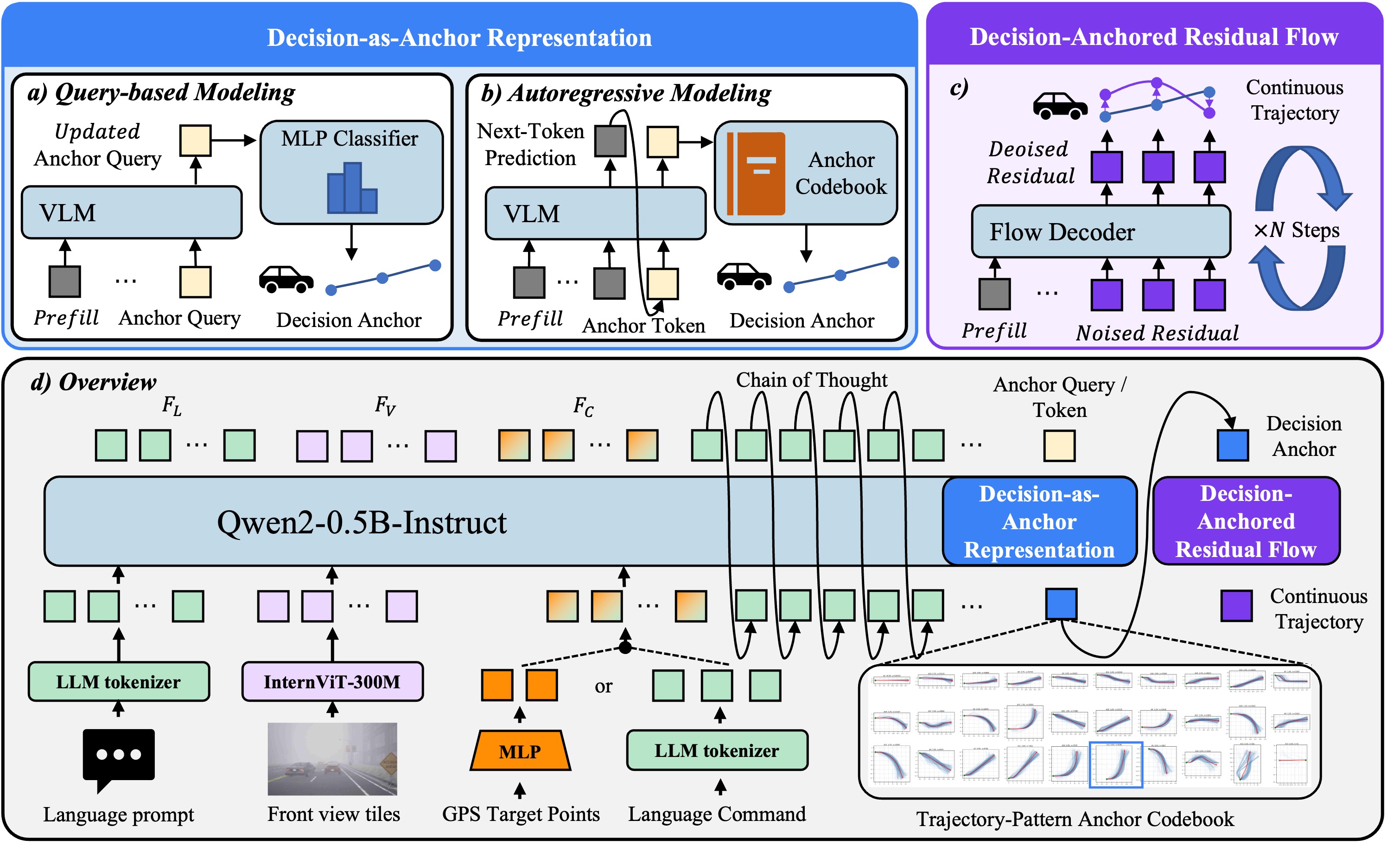}
    \caption{
    Overview of AnchorVLA. Query-based modeling (a) and autoregressive modeling (b) are two instantiations of Decision-as-Anchor Representation (DAAR) for selecting trajectory-pattern anchors. Given the selected anchors, Decision-Anchored Residual Flow (DARF) performs anchor-defined residual generation and confidence-based trajectory selection (c). The complete decision-to-execution pipeline is shown in (d).
    }
    \label{fig:overview}
\end{figure*}

As shown in Figure~\ref{fig:overview}(d), AnchorVLA takes language prompts, front-view image tiles, and navigation inputs as inputs. The multimodal context is encoded by an InternVL2-1B VLM~\cite{gao2024mini}, where image inputs are processed by InternViT-300M, and text inputs are tokenized and modeled by Qwen2-0.5B-Instruct~\cite{yang2024qwen2technicalreport}. GPS target points are projected into token embeddings through an MLP and integrated with the language tokens. The resulting multimodal tokens provide context features and chain-of-thought reasoning tokens for downstream planning. 

\subsection{Decision-as-Anchor Representation}

Decision-as-Anchor Representation (DAAR) maps high-level VLA reasoning
into a compact trajectory-pattern anchor space. Instead of requiring the
LLM to autoregressively generate low-level coordinate or waypoint tokens,
DAAR represents each behavior-level driving decision with a
trajectory-pattern anchor, where each anchor corresponds to a complete
local motion pattern. Given a trajectory-pattern codebook
\(\mathcal{A}=\{a_k\}_{k=1}^{K}\), constructed by clustering trajectories
from the training data, anchor prediction is formulated as estimating a
distribution over the shared anchor space.

We instantiate DAAR with two decision-modeling approaches: query-based
modeling and autoregressive modeling. Both approaches predict a
distribution over the same trajectory-pattern anchors, but differ in how
the anchor distribution is parameterized.

\paragraph{Query-Based Modeling.}
In query-based modeling, a learnable query is used as a feature-based
anchor classifier over the LLM/VLM context features. Specifically, the
learnable query is appended to the input sequence. After language-model
encoding, the output feature \(h_{\mathrm{qry}}\) corresponding to the
query is passed through an MLP head to produce anchor logits. The resulting
anchor distribution is
\begin{equation}
p_{\theta}^{\mathrm{qry}}(a_k\mid x)
=
\mathrm{softmax}(s^{\mathrm{qry}})_k,
\label{eq:query_anchor_prob}
\end{equation}
where \(s^{\mathrm{qry}}\in\mathbb{R}^{K}\) denotes the logits produced by
the query-based anchor classifier. This formulation predicts trajectory
anchors directly from the encoded context features without using the LLM
next-token prediction interface.

\paragraph{Autoregressive Modeling.}
Existing trajectory-decision modeling approaches exhibit two complementary
limitations. Full-trajectory autoregressive methods preserve the native
next-token prediction interface of the LLM, but represent a trajectory as a
long sequence of coordinate tokens $\mathbf{c}_{1:L}$:
\begin{equation}
p(\tau\mid x)
\approx
p(\mathbf{c}_{1:L}\mid x)
=
\prod_{\ell=1}^{L}
p(c_{\ell}\mid x,\mathbf{c}_{<\ell}),
\label{eq:coordinate_ar_factorization}
\end{equation}
where each decoding step predicts only one discretized coordinate token.
Consequently, generating a complete trajectory requires $L$ sequential
decoding steps. Query-based modeling avoids coordinate-token autoregression by
predicting anchor scores from learnable query features, but relies on an
additional classification head outside the native language-token output space.

Autoregressive DAAR combines compact anchor prediction with the native
next-token interface of the LLM. We introduce $K$ special anchor tokens
$\{y_k\}_{k=1}^{K}$ into the LLM vocabulary and establish a fixed one-to-one
mapping $y_k\leftrightarrow a_k$, where each anchor $a_k$ represents a complete
trajectory pattern. Given the multimodal input $x$ and preceding language
reasoning tokens $\mathbf{w}_{1:R}$, the anchor decision is formulated as
\begin{equation}
p_{\theta}^{\mathrm{ar}}
(a_k\mid x,\mathbf{w}_{1:R})
=
p_{\theta}
(y_k\mid x,\mathbf{w}_{1:R})
=
\mathrm{softmax}(s^{\mathrm{ar}})_k,
\label{eq:ar_anchor_prob}
\end{equation}
where $s^{\mathrm{ar}}\in\mathbb{R}^{K}$ denotes the logits over the
anchor-token vocabulary.

Compared with full-trajectory autoregression, DAAR replaces the $L$-step
coordinate-token sequence with a single anchor-token prediction, thereby
increasing the trajectory-level information represented by each decoding step
and reducing sequential decoding depth. Compared with query-based modeling, language reasoning and anchor decisions are
represented in the same native LLM token space, and the anchor probability is
directly conditioned on the preceding reasoning context through the same
next-token prediction objective, thereby strengthening the association between
VLM reasoning and the corresponding trajectory-pattern anchors.

\paragraph{Loss Design.}
For both query-based and autoregressive modeling, the predicted anchor
logits are supervised using the same soft target. Given the codebook
\(\mathcal{A}=\{a_k\}_{k=1}^{K}\), we compute the average distance
between the ground-truth trajectory \(\tau_{\mathrm{gt}}\) and each
codebook anchor \(a_k\):
\begin{equation}
d(\tau_{\mathrm{gt}},a_k)
=
\frac{1}{T}
\sum_{t=1}^{T}
\left\|
\tau_{\mathrm{gt}}^{t}-a_k^{t}
\right\|_2,
\qquad k=1,\ldots,K.
\label{eq:anchor_distance}
\end{equation}

We then select the \(N\) anchors closest to
\(\tau_{\mathrm{gt}}\) and denote them, in ascending order of distance,
as \(\{a_{(n)}\}_{n=1}^{N}\). The soft target assigned to the \(n\)-th
nearest anchor is defined as
\begin{equation}
q_n
=
\frac{
\exp\left(
-d(\tau_{\mathrm{gt}},a_{(n)})/\gamma
\right)
}{
\sum_{j=1}^{N}
\exp\left(
-d(\tau_{\mathrm{gt}},a_{(j)})/\gamma
\right)
},
\qquad n=1,\ldots,N,
\label{eq:soft_anchor_target}
\end{equation}
where \(\gamma\) is the temperature coefficient.

The logits from both modeling approaches are optimized using the same
soft-label cross-entropy loss:
\begin{equation}
\mathcal{L}_{\mathrm{DAAR}}^{m}
=
-\sum_{n=1}^{N}
q_n
\log p_{\theta}^{m}(a_{(n)}\mid x),
\qquad
m\in\{\mathrm{qry},\mathrm{ar}\}.
\label{eq:daar_anchor_loss}
\end{equation}

\subsection{Decision-Anchored Residual Flow}

While DAAR selects a behavior-level anchor as the high-level driving decision, the final executable trajectory still requires fine-grained continuous execution. This follows the decision-then-execution nature of real driving: after deciding an overall maneuver such as lane changing, yielding, or lane keeping, a driver still adapts the detailed trajectory according to nearby agents, road geometry, and dynamic scene changes. Therefore, the anchor-conditioned residual is naturally uncertain and multi-modal. Compared with deterministic residual regression, which tends to average multiple feasible corrections into a single prediction, generative flow matching models the residual distribution and enables diverse fine-grained trajectory refinement.

\begin{figure}[t]
    \centering
    \includegraphics[width=\columnwidth]{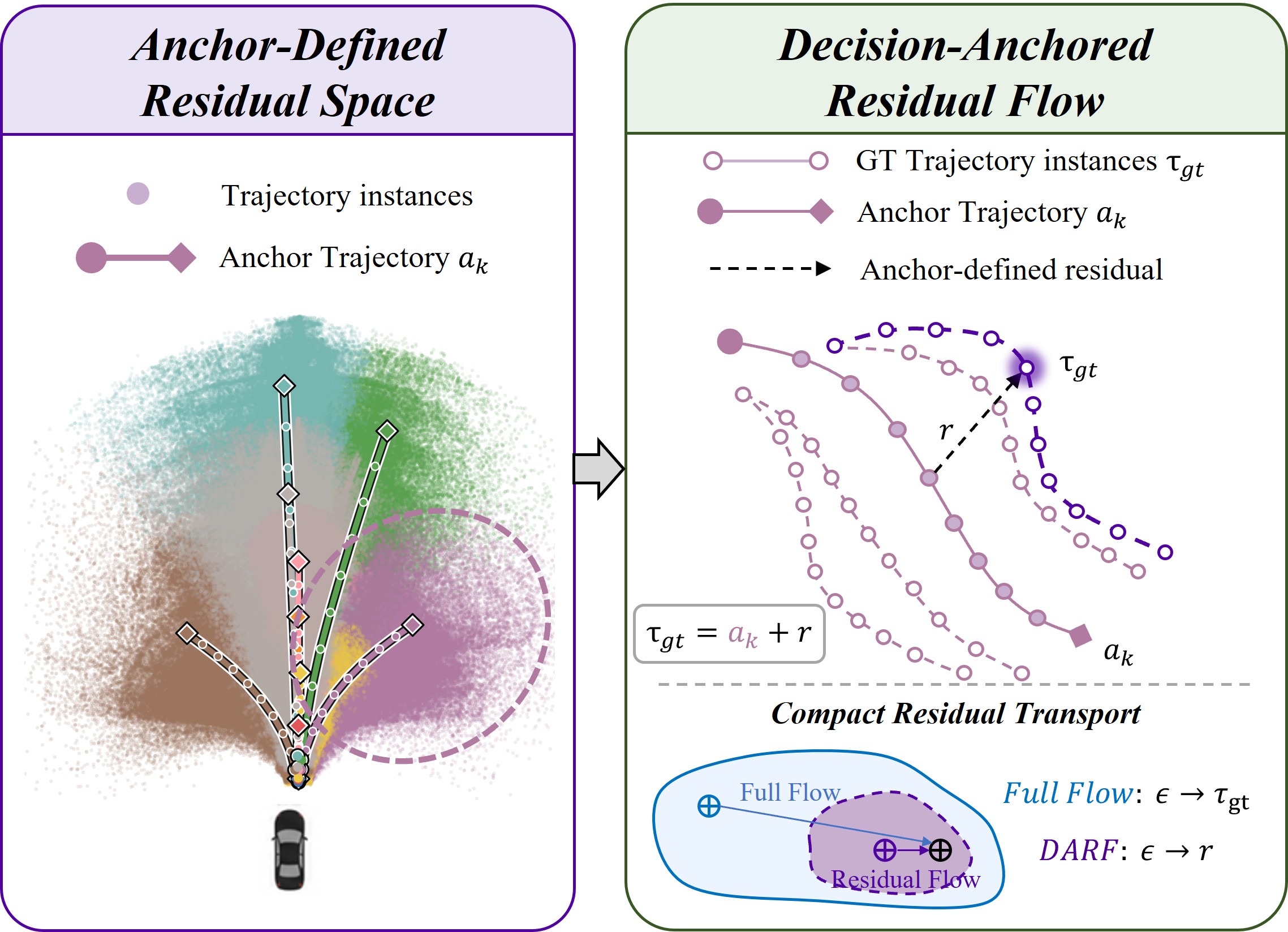}
    \caption{
    Illustration of Decision-Anchored Residual Flow (DARF), which generates a continuous residual in the space defined by the selected trajectory-pattern anchor.
    }
    \label{fig:resiflow}
\end{figure}

Figure~\ref{fig:resiflow} illustrates the proposed \textbf{Decision-Anchored Residual Flow} (DARF). Given the top-\(M\) candidate trajectory-pattern anchors predicted by DAAR, denoted by the index set \(\mathcal{I}_M\), DARF generates continuous trajectories in the corresponding anchor-defined residual spaces. For each candidate \(a_k\), \(k\in\mathcal{I}_M\), the candidate trajectory is represented as \(\tau_k=a_k+r_k\), where \(r_k\) is the continuous residual defined by anchor \(a_k\).

Given the ground-truth trajectory \(\tau_{\mathrm{gt}}\), we construct the target residual for each candidate anchor as \(r_{\mathrm{gt}}^k=\tau_{\mathrm{gt}}-a_k\). Full-trajectory flow directly transports noise toward the complete trajectory target according to \(z_t^{\mathrm{full}}=(1-t)\epsilon+t\tau_{\mathrm{gt}}\), whereas DARF transports noise toward the anchor-defined residual target. When the selected anchor captures the coarse motion pattern of the target trajectory, the residual satisfies \(\|r_{\mathrm{gt}}^k\|_2<\|\tau_{\mathrm{gt}}\|_2\). Under isotropic Gaussian noise, this yields a shorter expected transport path:
\begin{equation}
\mathbb{E}\!\left[
\|r_{\mathrm{gt}}^k-\epsilon\|_2^2
\right]
<
\mathbb{E}\!\left[
\|\tau_{\mathrm{gt}}-\epsilon\|_2^2
\right].
\label{eq:transport_distance}
\end{equation}
Therefore, under this condition and the same number of flow-integration steps, DARF requires a smaller overall correction from noise to the target. This reduces the integration difficulty and accumulated numerical integration error under a limited step budget, leading to more accurate continuous trajectory generation.

We sample initial noise \(\epsilon\sim\mathcal{N}(0,I)\) and a timestep \(t\in[0,1]\), and construct the residual flow path for each candidate anchor as
\begin{equation}
z_t^k=(1-t)\epsilon+t r_{\mathrm{gt}}^k,
\qquad
v_k^\ast=r_{\mathrm{gt}}^k-\epsilon.
\label{eq:tarflow_path}
\end{equation}
DARF predicts the corresponding velocity as
\(\hat{v}_k=v_\theta(z_t^k,t,x,a_k)\), where \(v_k^\ast\) is the target velocity and \(\hat{v}_k\) is the predicted velocity.

Among the top-\(M\) candidates, we select the anchor closest to the ground-truth trajectory, \(k^\dagger=\arg\min_{k\in\mathcal{I}_M}d(\tau_{\mathrm{gt}},a_k)\), and compute the flow-matching loss on the matched candidate:
\begin{equation}
\mathcal{L}_{\mathrm{flow}}
=
\left\|
\hat{v}_{k^\dagger}
-
v_{k^\dagger}^{\ast}
\right\|_1.
\label{eq:tarflow_loss}
\end{equation}
This matched-anchor supervision encourages the decoder to refine the most relevant candidate among the predicted high-level decisions while preserving multiple behavior hypotheses before final selection.

DARF also predicts a confidence score \(s_k\) for each candidate trajectory induced by anchor \(a_k\). The confidence score evaluates the compatibility between each candidate anchor and the driving context and is used for final trajectory selection during inference. During training, the confidence is supervised by the matched anchor:
\begin{equation}
\mathcal{L}_{\mathrm{conf}}
=
-\log
\frac{\exp(s_{k^\dagger})}
{\sum_{k\in\mathcal{I}_M}\exp(s_k)}.
\label{eq:tarflow_conf_loss}
\end{equation}

During inference, starting from the initial noisy residual \(z_0^k=\epsilon\), we integrate the velocity field with a few Euler steps:
\begin{equation}
z_{t+\Delta t}^k
=
z_t^k
+
\Delta t\,
v_\theta(z_t^k,t,x,a_k).
\label{eq:tarflow_inference}
\end{equation}
The candidate trajectory is reconstructed as
\(\hat{\tau}_k=a_k+z_1^k\), and the final trajectory is selected according to the confidence scores as
\(\hat{\tau}=\hat{\tau}_{\arg\max_{k\in\mathcal{I}_M}s_k}\).
Through this design, DARF generates continuous trajectories for multiple candidate trajectory-pattern anchors and maintains multiple high-level behavior hypotheses before final trajectory selection.

\section{Experiment}

\begin{table}[t]
\centering

\renewcommand{\arraystretch}{1.08}
\setlength{\tabcolsep}{3pt}
\scriptsize

\resizebox{\columnwidth}{!}{%
\begin{tabular}{l|c|c|c|rrrr}
\hline
\multirow{2}{*}{\textbf{Method}}
& \multirow{2}{*}{\textbf{Expert}}
& \multirow{2}{*}{\textbf{VLA}}
& \multirow{2}{*}{\shortstack{\textbf{Generative}\\\textbf{Planning}}}
& \multicolumn{4}{c}{\textbf{Closed-loop metrics} \(\uparrow\)} \\
\cline{5-8}
& & & &
\textbf{DS} &
\textbf{SR (\%)} &
\textbf{Eff.} &
\textbf{Comf.} \\
\hline

TCP-traj~\cite{wu2022trajectory}
& Think2Drive & \(\times\) & \(\times\)
& 59.90 & 30.00 & 76.54 & 18.08 \\

ThinkTwice~\cite{jia2023think}
& Think2Drive & \(\times\) & \(\times\)
& 62.44 & 31.23 & 69.33 & 16.22 \\

DriveAdapter~\cite{jia2023driveadapter}
& Think2Drive & \(\times\) & \(\times\)
& 64.22 & 33.08 & 70.22 & 16.01 \\

AD-MLP~\cite{zhai2023rethinking}
& Think2Drive & \(\times\) & \(\times\)
& 18.05 & 0.00 & 48.45 & 22.63 \\

UniAD-Base~\cite{hu2023planning}
& Think2Drive & \(\times\) & \(\times\)
& 45.81 & 16.36 & 129.21 & 43.58 \\

VAD~\cite{jiang2023vad}
& Think2Drive & \(\times\) & \(\times\)
& 42.35 & 15.00 & 157.94 & \textbf{46.01} \\

DriveTransformer~\cite{jia2025drivetransformer}
& Think2Drive & \(\times\) & \(\times\)
& 63.46 & 35.01 & 100.64 & 20.78 \\

Orion-VAE~\cite{fu2025orion}
& Think2Drive & \(\checkmark\) & \(\times\)
& 77.74 & 54.62 & 151.48 & 17.38 \\

Orion-Diffusion~\cite{fu2025orion}
& Think2Drive & \(\checkmark\) & \(\checkmark\)
& 71.97 & 46.54 & -- & -- \\

AutoVLA~\cite{zhou2025autovla}
& PDM-Lite & \(\checkmark\) & \(\times\)
& 78.84 & 57.73 & 146.93 & 39.33 \\

DiffusionDrive~\cite{liao2025diffusiondrive}
& PDM-Lite & \(\times\) & \(\checkmark\)
& 80.79 & 58.18 & 248.18 & 24.56 \\

SimLingo~\cite{renz2025simlingo}
& PDM-Lite & \(\checkmark\) & \(\times\)
& 85.07 & 67.27 & \textbf{259.23} & 33.67 \\

TransFuser++~\cite{chitta2022transfuser}
& PDM-Lite & \(\times\) & \(\times\)
& 84.21 & 67.27 & -- & -- \\

LinkVLA~\cite{wang2026unifying}
& PDM-Lite & \(\checkmark\) & \(\times\)
& \textbf{91.01} & 74.55 & 255.84 & 34.62 \\

BridgeDrive~\cite{liu2025bridgedrive}
& PDM-Lite & \(\times\) & \(\checkmark\)
& 87.99 & 74.99 & 236.49 & 20.98 \\

\hline
\textbf{AnchorVLA}
& PDM-Lite & \(\checkmark\) & \(\checkmark\)
& 89.92 & \textbf{77.28} & 251.14 & 28.94 \\
\hline
\end{tabular}%
}

\caption{
Comparison with existing methods on the Bench2Drive closed-loop benchmark.
``VLA'' indicates the use of vision-language-action models or
language-conditioned reasoning, while ``Generative Planning'' denotes
diffusion- or flow-based continuous trajectory generation.
}
\label{tab:closed_loop_results}

\end{table}
\subsection{Settings}

\textbf{Training data and benchmark.}
We follow the training-data setting of SimLingo~\citep{renz2025simlingo}, which includes expert trajectories, instruction-conditioned dream trajectories, and mixed language supervision consisting of VQA, driving commentary, and no-language samples. We evaluate AnchorVLA on the Bench2Drive~\citep{jia2024bench2drive} closed-loop benchmark built on CARLA and report Driving Score (DS), Success Rate (SR), Efficiency, Comfort, and Multi-Ability scores. SR measures whether the agent completes the driving task and reflects closed-loop reliability.

\textbf{VLA backbone.}
We build AnchorVLA on the SimLingo~\citep{renz2025simlingo} VLA backbone, which provides multimodal scene encoding and language-conditioned reasoning. We keep the backbone unchanged and modify only the planning interface.

\textbf{Trajectory-pattern anchors.}
We construct the trajectory-pattern codebook by applying \(k\)-means clustering to the training trajectories. Unless otherwise specified, we use \(K=100\) anchors, each corresponding to a cluster center in the trajectory space. During inference, the top-\(M\) predicted anchors are retained as candidate high-level decisions for DARF, with \(M=6\).

\textbf{Training.}
We train AnchorVLA in two stages. First, we train the anchor-decision model under DAAR for \(15\) epochs on \(8\) NVIDIA A100 GPUs with a batch size of \(16\). We then freeze the VLA backbone and train DARF for residual-based continuous trajectory generation for another \(15\) epochs on \(4\) NVIDIA A100 GPUs with a batch size of \(32\). Additional optimization and implementation details are provided in the supplementary material.

\begin{figure*}[t]
    \centering
    \includegraphics[
        width=0.85\textwidth
    ]{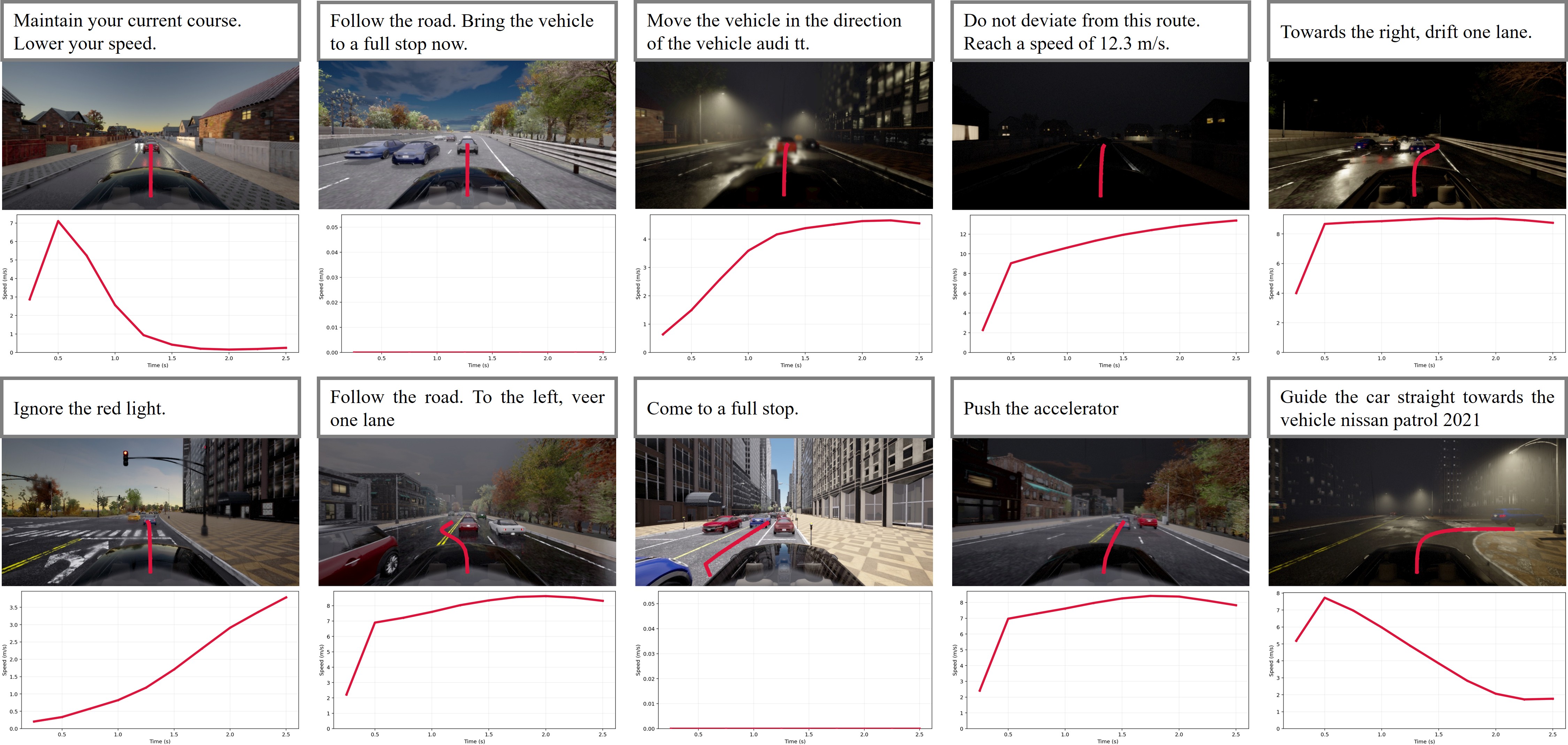}
    \caption{
    Qualitative visualization of the top-1 trajectory-pattern anchors
    selected by DAAR on the SimLingo Action Dreaming data.
    }
    \label{fig:anchor_visualization}
\end{figure*}

\subsection{Main Results}

Table~\ref{tab:closed_loop_results} compares AnchorVLA with representative
end-to-end and VLA planning methods on the Bench2Drive closed-loop
benchmark. AnchorVLA achieves the highest SR among the compared methods
while maintaining a competitive DS. Compared with LinkVLA, it obtains a
higher SR with a slightly lower DS, and it outperforms BridgeDrive on both
metrics. More importantly, AnchorVLA and SimLingo use the same VLA backbone and
training-data setting, providing a controlled comparison of the planning
architecture. Replacing the query-based planning head with DAAR and DARF
substantially improves both DS and SR. These results support the
effectiveness of representing high-level decisions with trajectory-pattern
anchors and refining continuous trajectories in anchor-defined residual
coordinates. AnchorVLA also achieves competitive efficiency and the highest Comfort
score among the compared generative planners, indicating that its improved
closed-loop performance does not come at the expense of trajectory comfort.

\subsection{Multi-Ability Evaluation}

Table~\ref{tab:multi_ability} reports the Bench2Drive multi-ability results
across Merging, Overtake, Brake, Give-Way, and Traffic-Sign scenarios.
AnchorVLA achieves the highest mean score among the compared methods,
showing balanced performance across diverse interactive driving abilities. In particular, AnchorVLA obtains the best result in Overtake, indicating
strong performance in scenarios that require active interaction and
behavior-level decision making. Overall, the multi-ability results further
support the effectiveness of the proposed hierarchical decision and
continuous refinement framework.

\begin{table}[t]
\centering
\renewcommand{\arraystretch}{1.10}
\setlength{\tabcolsep}{3pt}
\scriptsize

\begin{adjustbox}{width=\columnwidth}
\begin{tabular}{l|cccccc}
\hline
\multirow{2}{*}{\textbf{Method}}
& \multicolumn{6}{c}{\textbf{Multi-Ability (\%)} \(\uparrow\)} \\
\cline{2-7}
& \textbf{Merging}
& \textbf{Overtake}
& \textbf{Brake}
& \textbf{Give-Way}
& \textbf{Traffic-Sign}
& \textbf{Mean} \\
\hline
TCP-traj~\cite{wu2022trajectory} & 8.89 & 24.29 & 51.67 & 40.00 & 46.28 & 34.22 \\
ThinkTwice~\cite{jia2023think} & 27.38 & 18.42 & 35.82 & 50.00 & 54.23 & 37.17 \\
DriveAdapter~\cite{jia2023driveadapter} & 28.82 & 26.38 & 48.76 & 50.00 & 56.43 & 42.08 \\
AD-MLP~\cite{zhai2023rethinking} & 0.00 & 0.00 & 0.00 & 0.00 & 4.35 & 0.87 \\
UniAD-Base~\cite{hu2023planning} & 14.10 & 17.78 & 21.67 & 10.00 & 14.21 & 15.55 \\
VAD~\cite{jiang2023vad} & 8.11 & 24.44 & 18.64 & 20.00 & 19.15 & 18.07 \\
DriveTransformer~\cite{jia2025drivetransformer} & 17.57 & 35.00 & 48.36 & 40.00 & 52.10 & 38.60 \\
Orion-VAE~\cite{fu2025orion} & 25.00 & 71.11 & 78.33 & 30.00 & 69.15 & 54.72 \\
DiffusionDrive~\cite{liao2025diffusiondrive} & 50.63 & 26.67 & 68.33 & 50.00 & 76.32 & 54.38 \\
SimLingo~\cite{renz2025simlingo} & 53.75 & 68.89 & 81.67 & 50.00 & 82.11 & 67.28 \\
TransFuser++~\cite{chitta2022transfuser} & 58.75 & 57.77 & 83.33 & 40.00 & 82.11 & 64.39 \\
LinkVLA~\cite{wang2026unifying} & 60.00 & 80.00 & \textbf{93.33} & 50.00 & 83.68 & 73.40 \\
BridgeDrive~\cite{liu2025bridgedrive} & \textbf{69.92} & 66.67 & 90.00 & 50.00 & \textbf{89.47} & 73.15 \\
\hline
\textbf{AnchorVLA}
& 65.00
& \textbf{81.11}
& 90.00
& 50.00
& 85.00
& \textbf{74.22} \\
\hline
\end{tabular}
\end{adjustbox}
\caption{
Comparison with existing methods on the Bench2Drive multi-ability evaluation.
}
\label{tab:multi_ability}
\end{table}

\subsection{Qualitative Analysis}
We conduct a qualitative analysis on the SimLingo Action Dreaming
data, which contains rich and diverse language
instructions and therefore provides a suitable setting for examining how
trajectory-pattern anchors connect discrete language-conditioned decisions
with continuous trajectories. Figure~\ref{fig:anchor_visualization}
presents ten representative examples. For each example, the first row shows
the language command, the second row overlays the top-1 future-path anchor
predicted by DAAR on the front-view camera image, and the third row shows
the corresponding top-1 speed anchor over the planning horizon. The path
anchor describes the spatial motion pattern, while the speed anchor
specifies the associated temporal velocity profile. As shown in the figure, the predicted trajectory-pattern anchors provide an
explicit interface between language-conditioned driving decisions and
continuous path and speed representations.

\begin{table}[t]
\centering
\renewcommand{\arraystretch}{1.10}
\setlength{\tabcolsep}{2.5pt}
\scriptsize

\begin{adjustbox}{width=\columnwidth}
\begin{tabular}{l|c|c|c|cccc}
\hline
\multirow{2}{*}{\textbf{Method}}
&
\multirow{2}{*}{
  \shortstack{\textbf{Decision}\\\textbf{Latency}}
}
&
\multirow{2}{*}{
  \shortstack{\textbf{Flow}\\\textbf{Method}}
}
&
\multirow{2}{*}{
  \shortstack{\textbf{Flow}\\\textbf{Steps}}
}
&
\multicolumn{4}{c}{\textbf{Closed-loop metrics} \(\uparrow\)}
\\
\cline{5-8}
& & & &
\textbf{DS}
&
\textbf{SR (\%)}
&
\textbf{Eff.}
&
\textbf{Comf.}
\\
\hline


None
& --
& DARF
& 2
& 87.49
& 70.91
& \textbf{260.97}
& 28.96
\\

Query Based
& 33 ms
& DARF
& 2
& 88.67
& 73.81
& 256.31
& \textbf{31.39}
\\

\textbf{Autoregressive}
& 64 ms
& DARF
& 2
& \textbf{89.92}
& \textbf{77.28}
& 251.14
& 28.94
\\
\hline
\end{tabular}
\end{adjustbox}
\caption{
Ablation of Decision-as-Anchor Modeling.
}
\label{tab:ablation_decision_method}
\end{table}

\begin{table}[t]
\centering
\renewcommand{\arraystretch}{1.10}
\setlength{\tabcolsep}{2.5pt}
\scriptsize

\begin{adjustbox}{width=\columnwidth}
\begin{tabular}{l|c|c|c|cccc}
\hline
\multirow{2}{*}{
  \shortstack{\textbf{Flow Matching}\\\textbf{Method}}
}
&
\multirow{2}{*}{
  \shortstack{\textbf{Decoder}\\\textbf{Latency}}
}
&
\multirow{2}{*}{
  \shortstack{\textbf{Decision}\\\textbf{Method}}
}
&
\multirow{2}{*}{
  \shortstack{\textbf{Flow}\\\textbf{Steps}}
}
&
\multicolumn{4}{c}{\textbf{Closed-loop metrics} \(\uparrow\)}
\\
\cline{5-8}
& & & &
\textbf{DS}
&
\textbf{SR (\%)}
&
\textbf{Eff.}
&
\textbf{Comf.}
\\
\hline

Deterministic
& 8 ms
& Autoregressive
& 1
& 86.74
& 70.45
& \textbf{254.52}
& 28.11
\\

Full Flow
& 10 ms
& Autoregressive
& 2
& 88.34
& 72.73
& 250.30
& 26.13
\\

\textbf{DARF}
& 10 ms
& Autoregressive
& 2
& \textbf{89.92}
& \textbf{77.28}
& 251.14
& \textbf{28.94}
\\
\hline
\end{tabular}
\end{adjustbox}
\caption{
Ablation of Flow-Matching Formulations.
}
\label{tab:ablation_flow_matching_method}
\end{table}

\subsection{Ablation Study}

\paragraph{Ablation on Decision-as-Anchor Modeling.}
Table~\ref{tab:ablation_decision_method} compares three high-level decision
modeling choices: no explicit anchor decision, query-based anchor prediction,
and autoregressive anchor prediction. All variants use the same
trajectory-pattern codebook and DARF decoder, differing only in how the
anchors are selected. Directly passing all codebook anchors to DARF yields the weakest
closed-loop performance, while explicitly predicting the relevant anchors
leads to clear improvements. Autoregressive anchor prediction further
outperforms its query-based counterpart. By representing trajectory-pattern
anchors as LLM action tokens, the autoregressive formulation performs
behavior-level trajectory selection through the native next-token
prediction interface, providing a direct connection between VLA reasoning
and the subsequent continuous trajectory refinement.

\paragraph{Ablation on Flow-Matching Formulations.}
Table~\ref{tab:ablation_flow_matching_method} compares deterministic
residual prediction, full-trajectory flow matching, and our DARF. All variants share the same autoregressive anchor-decision model and decoder architecture and differ only in the continuous trajectory-generation formulation.Full-trajectory flow matching improves over deterministic residual regression, while DARF achieves the best closed-loop performance. Notably, Full Flow and DARF use the same two flow-integration steps and have comparable decoder latency, indicating that the gain does not arise from additional sampling steps or model capacity. When the selected anchor provides a suitable coarse trajectory reference, the corresponding residual target is more compact than the complete trajectory target. The controlled comparison therefore supports modeling continuous refinement in anchor-defined residual coordinates under a limited integration-step budget.

\begin{table}[t]
\centering
\setlength{\tabcolsep}{5.5pt}
\renewcommand{\arraystretch}{1.08}

\begin{adjustbox}{width=\columnwidth}
\begin{tabular}{l|cc|cccccc}
\toprule
\multirow{2}{*}{\textbf{Method}}
& \multicolumn{2}{c|}{\textbf{Closed-loop metrics} \(\uparrow\)}
& \multicolumn{6}{c}{\textbf{Multi-Ability (\%)} \(\uparrow\)} \\
\cmidrule(lr){2-3}
\cmidrule(lr){4-9}
& \textbf{DS}
& \textbf{SR (\%)}
& \textbf{Merging}
& \textbf{Overtake}
& \textbf{Brake}
& \textbf{Give-Way}
& \textbf{Traffic-Sign}
& \textbf{Mean} \\
\midrule

GPS target
& 89.92
& \textbf{77.28}
& \textbf{65.00}
& \textbf{81.11}
& 90.00
& \textbf{50.00}
& \textbf{85.00}
& \textbf{74.22} \\

Nav. command
& \textbf{90.26}
& 76.82
& 61.25
& 80.00
& \textbf{95.00}
& 50.00
& 84.74
& 74.20 \\
\bottomrule
\end{tabular}
\end{adjustbox}
\caption{
Ablation of navigation modalities.
}
\label{tab:ablation_navigation_input}
\end{table}

\paragraph{Navigation Modalities.}
Table~\ref{tab:ablation_navigation_input} shows that GPS target points and navigation commands achieve comparable closed-loop performance, indicating that AnchorVLA is robust to different forms of navigation inputs.

\begin{table}[t]
    \centering
    \resizebox{\columnwidth}{!}{
    \begin{tabular}{lcccc}
        \toprule
        Method
        & Planning Type
        & Latency (ms) $\downarrow$
        & DS $\uparrow$
        & SR (\%) $\uparrow$ \\
        \midrule

        SimLingo~\cite{renz2025simlingo}
        & MLP
        & \textbf{38}
        & 85.07
        & 67.27 \\

        SimLingo-AR
        & Coordinate AR
        & 361
        & 89.57
        & 73.18 \\

        AnchorVLA
        & DAAR + DARF
        & 74
        & \textbf{89.92}
        & \textbf{77.28} \\
        \bottomrule
    \end{tabular}
    }
    \caption{
        Comparison of action-planning approaches based on the same SimLingo VLA backbone. 
        }
    \label{tab:action_planning_comparison}
\end{table}

\paragraph{Inference Latency.}
Table~\ref{tab:action_planning_comparison} compares SimLingo, SimLingo-AR, and AnchorVLA under the same VLA backbone and trajectory horizon. SimLingo-AR applies LinkVLA-style coordinate-token discretization and autoregressive decoding to SimLingo. The reported latency covers complete trajectory generation without the shared CoT stage. Compared with coordinate-token autoregressive decoding, AnchorVLA substantially reduces inference latency while achieving higher DS and SR, demonstrating a better trade-off between planning efficiency and closed-loop performance.

\section{Conclusion and Discussion}

We present \textbf{AnchorVLA}, a hierarchical Decision-Anchored VLA planning framework that bridges high-level VLA reasoning and continuous trajectory generation through trajectory-pattern anchors. AnchorVLA represents behavior-level decisions as trajectory-pattern anchors, which serve as explicit decision interfaces and guide subsequent continuous generation through the corresponding anchor-defined residual spaces. Based on the selected anchors, Decision-Anchored Residual Flow models multi-modal continuous residual refinements for executable trajectory generation. Experiments on Bench2Drive show that AnchorVLA achieves a state-of-the-art closed-loop Success Rate while maintaining a competitive Driving Score. These results demonstrate the effectiveness of trajectory-pattern anchors as an explicit intermediate interface between high-level VLA reasoning and continuous trajectory execution.

\bibliography{aaai2027}

\clearpage
\appendix

\section{Dual-Trajectory Representation in the SimLingo Instantiation}
\label{sec:dual_trajectory_representation}

The AnchorVLA formulation presented in the main paper operates on a generic
continuous trajectory $\tau$ and is independent of its specific
parameterization. Our implementation follows the dual-trajectory action
representation of SimLingo~\cite{renz2025simlingo}. A complete planning output
is represented as
\begin{equation}
    \mathbf{u}
    =
    \left(
        \tau^{\mathrm{dir}},
        \tau^{\mathrm{vel}}
    \right),
\end{equation}
where $\tau^{\mathrm{dir}}$ denotes a
\textcolor{red}{Direction trajectory} and $\tau^{\mathrm{vel}}$ denotes a
\textcolor{green}{Velocity trajectory}. The two trajectories jointly describe
the spatial and temporal evolution of the future ego motion.

The \textcolor{red}{Direction trajectory}
$\tau^{\mathrm{dir}}$ is a navigation-conditioned difference trajectory that
describes the directional evolution of future motion relative to the
navigation reference. It comprises 20 spatial waypoints, each corresponding
to a $1\,\mathrm{m}$ increment along the future path, and therefore covers a
total path length of $20\,\mathrm{m}$. This trajectory primarily represents
path geometry, lateral motion, and directional changes.

The \textcolor{green}{Velocity trajectory}
$\tau^{\mathrm{vel}}$ consists of 10 future ego waypoints sampled at intervals
of $0.25\,\mathrm{s}$, covering a prediction horizon of $2.5\,\mathrm{s}$.
The spatial displacements between consecutive waypoints represent the
longitudinal velocity of the vehicle and its temporal evolution.

For each trajectory type
$q\in\{\mathrm{dir},\mathrm{vel}\}$, we construct a trajectory-pattern
codebook
\begin{equation}
    \mathcal{A}^{q}
    =
    \{a_k^{q}\}_{k=1}^{K_q},
\end{equation}
and instantiate the DAAR-based anchor prediction and DARF-based continuous
residual refinement. The trajectory produced for each type is represented as
\begin{equation}
    \tau^{q}
    =
    a_k^{q}+r_k^{q},
    \qquad
    q\in\{\mathrm{dir},\mathrm{vel}\},
\end{equation}
where $a_k^{q}$ is the selected trajectory-pattern anchor and $r_k^{q}$ is the
continuous residual generated by DARF. This gives two trajectory-specific
instances of the generic decision-anchored formulation, whose outputs jointly
form the SimLingo planning representation.

In the autoregressive DAAR model, the Direction- and Velocity-anchor tokens
are predicted sequentially, with the Direction token generated first and the
Velocity token generated second.

DARF subsequently generates continuous residuals for the selected Direction
and Velocity anchors. The refined trajectories
$(\hat{\tau}^{\mathrm{dir}},\hat{\tau}^{\mathrm{vel}})$ constitute the complete
planning output used by the SimLingo execution interface. The complete
dual-trajectory generation process, including the decoding of both anchor
tokens and the DARF refinement of both trajectory streams, is used as the
evaluation unit for planning latency.


\newcommand{\anchorfigdir}{Figures}

\newlength{\anchorpanelheight}
\setlength{\anchorpanelheight}{0.155\textwidth}

\newcommand{\anchorpanel}[1]{%
    \parbox[b][\anchorpanelheight][c]{0.218\textwidth}{%
        \centering
        \includegraphics[
            width=0.218\textwidth,
            height=\anchorpanelheight,
            keepaspectratio
        ]{#1}%
    }%
}

\newcommand{\anchorrowlabel}[1]{%
    \parbox[b][\anchorpanelheight][c]{0.028\textwidth}{%
        \centering
        \rotatebox[origin=c]{90}{\small\bfseries #1}%
    }%
}

\newcommand{\anchorcolumnheader}[1]{%
    \parbox[c][2.5em][c]{0.218\textwidth}{%
        \centering
        \scriptsize #1%
    }%
}

\begin{figure*}[t]
    \centering

    \begingroup
    \setlength{\tabcolsep}{1.5pt}
    \renewcommand{\arraystretch}{0.92}

    \begin{tabular}{@{}c@{\hspace{1mm}}cccc@{}}
        &
        \multicolumn{2}{c}{\large\textbf{Velocity Anchors}}
        &
        \multicolumn{2}{c}{\large\textbf{Direction Anchors}}
        \\[-0.5mm]

        &
        \anchorcolumnheader{Top-1 Selection Distribution}
        &
        \anchorcolumnheader{Top-6 Speed Profiles}
        &
        \anchorcolumnheader{Top-1 Selection Distribution}
        &
        \anchorcolumnheader{Top-6 Spatial Paths}
        \\[0.8mm]

        \anchorrowlabel{Stop}
        &
        \anchorpanel{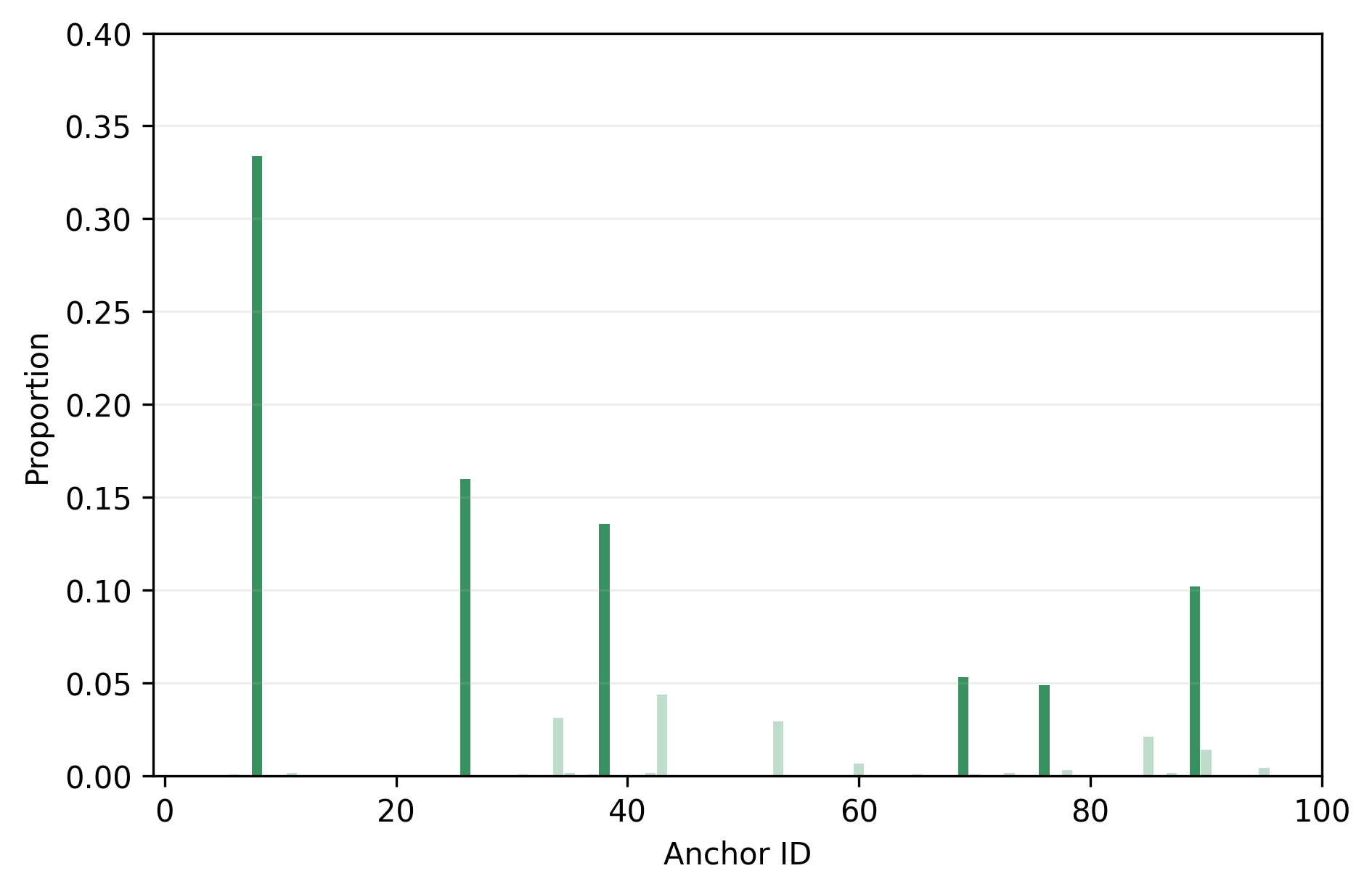}
        &
        \anchorpanel{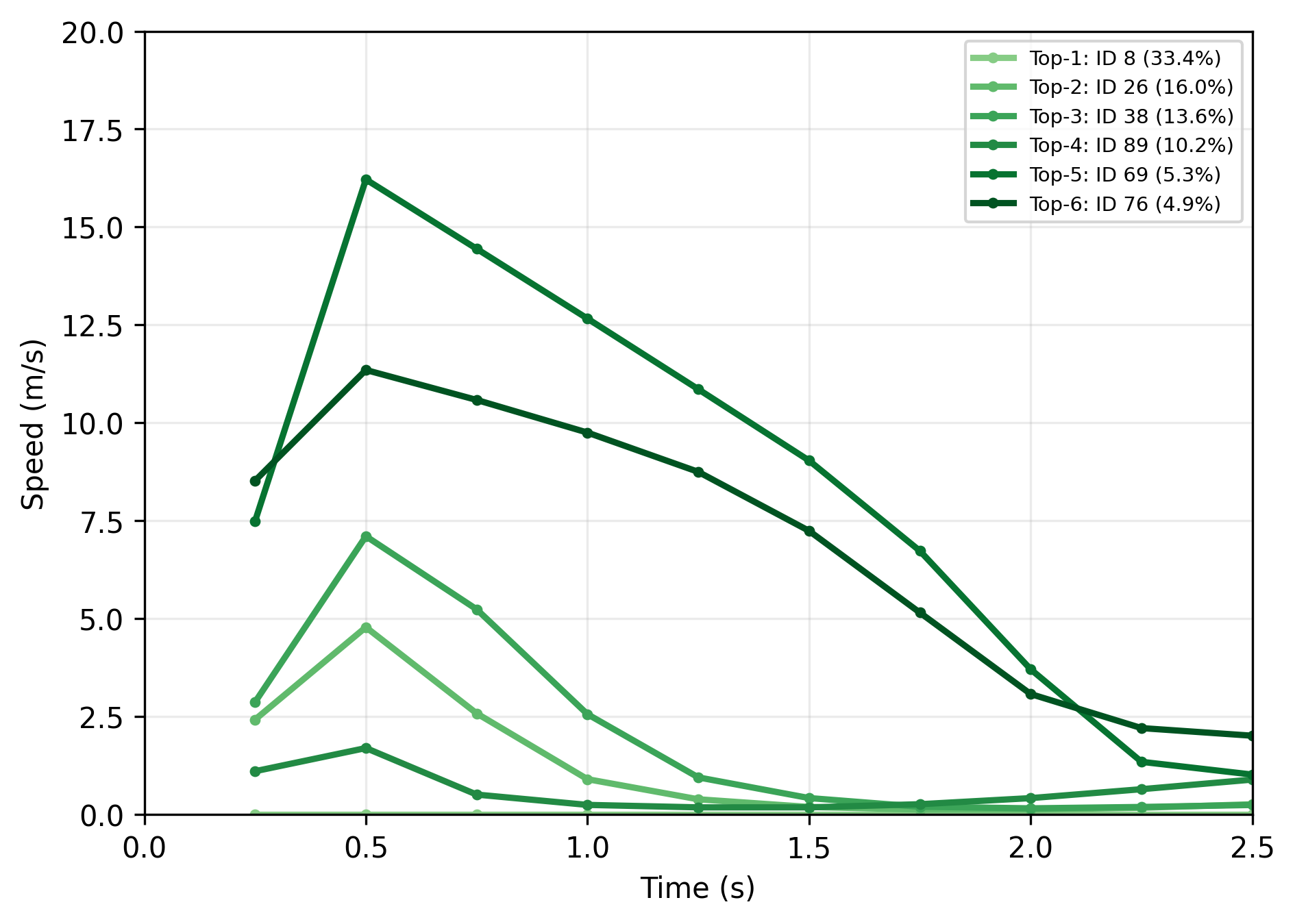}
        &
        \anchorpanel{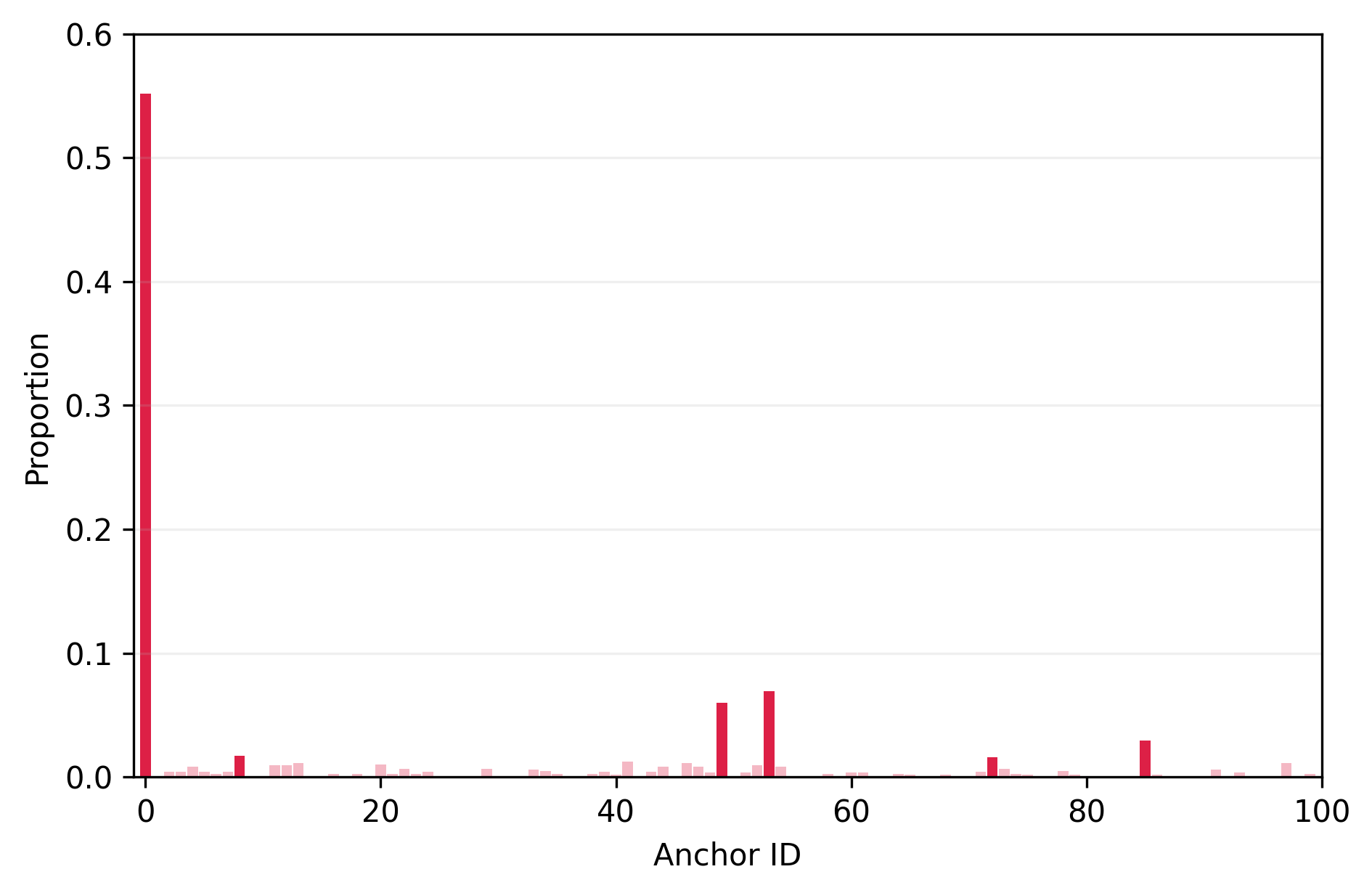}
        &
        \anchorpanel{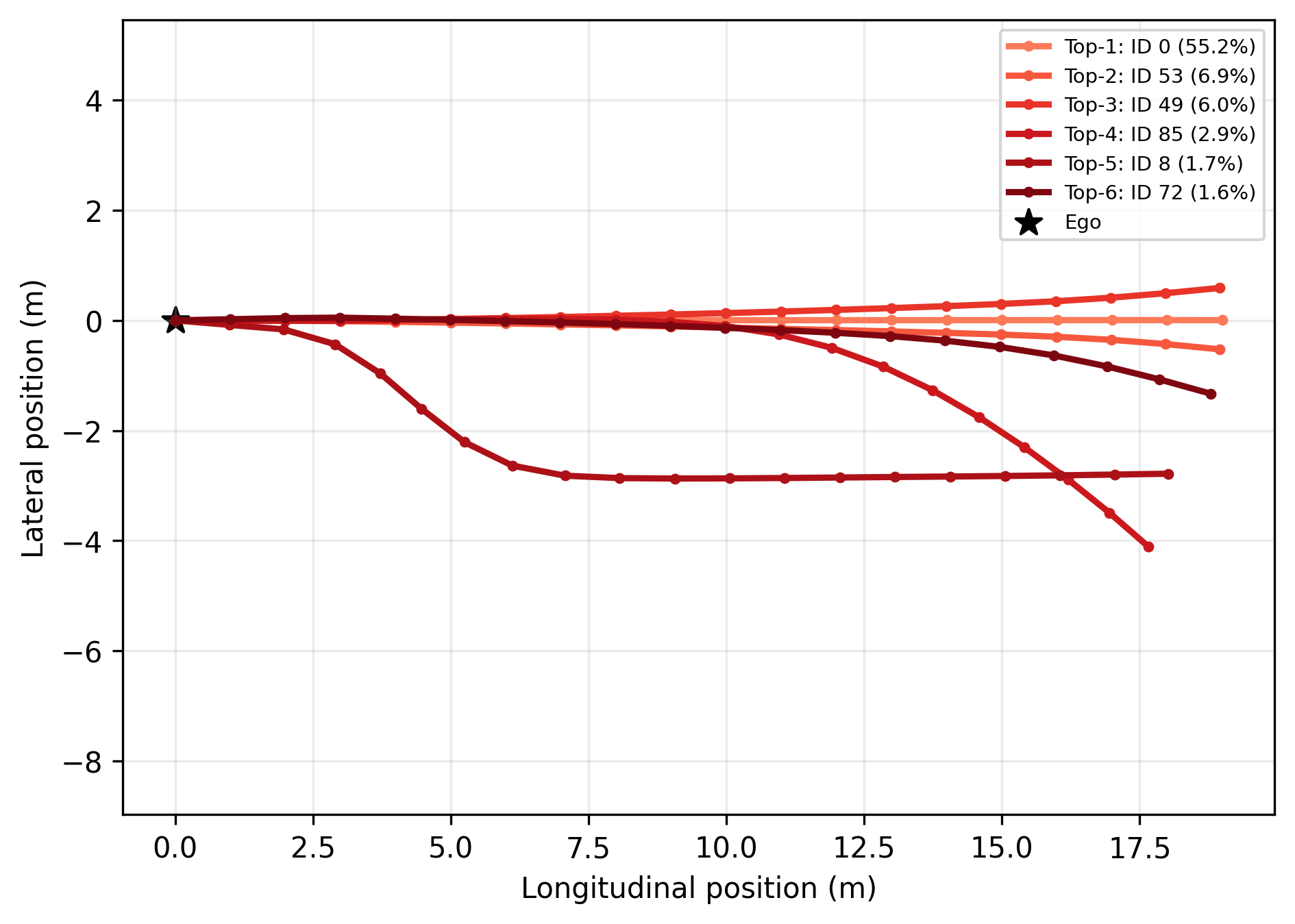}
        \\[0.5mm]

        \anchorrowlabel{Slower}
        &
        \anchorpanel{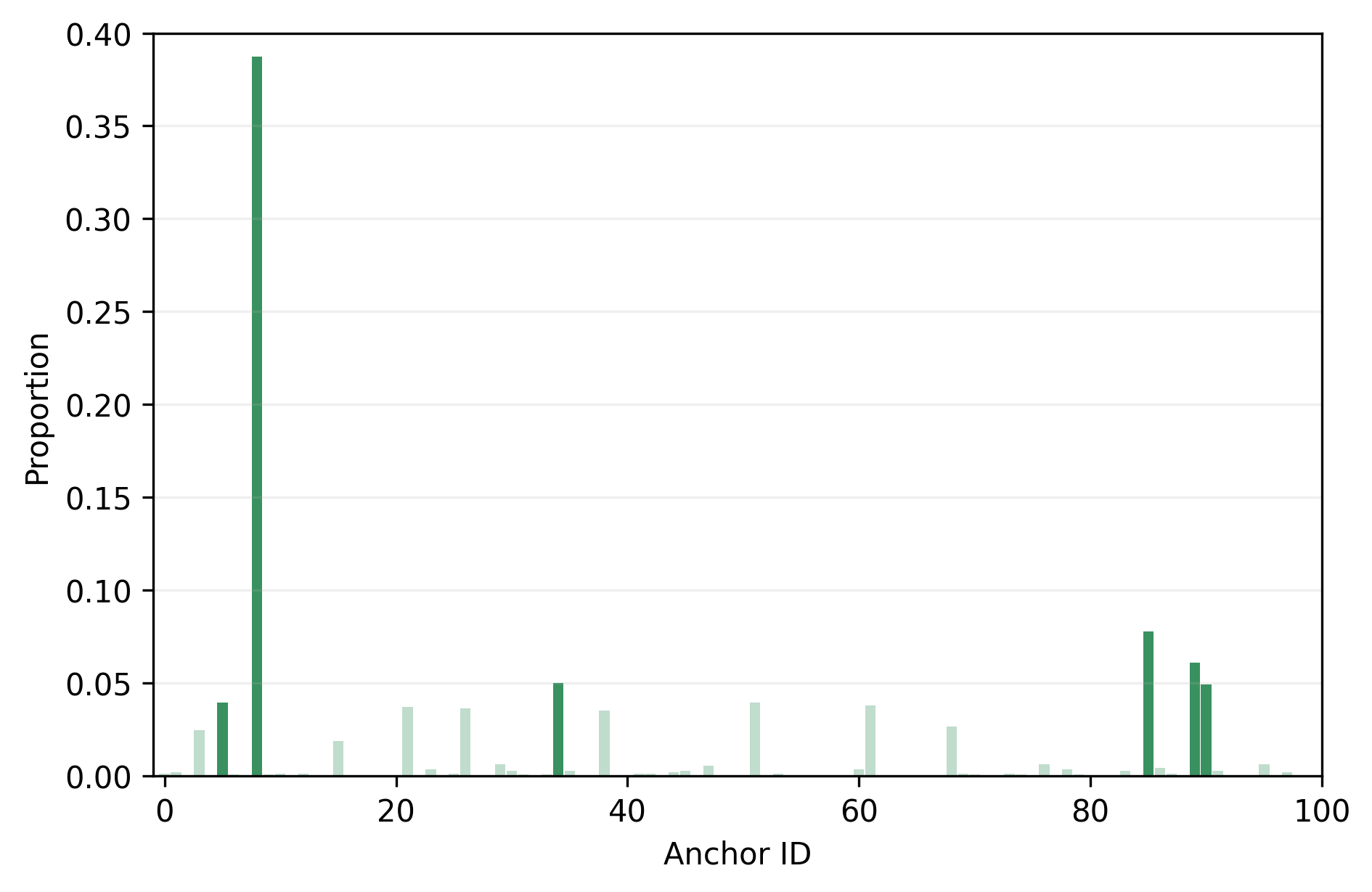}
        &
        \anchorpanel{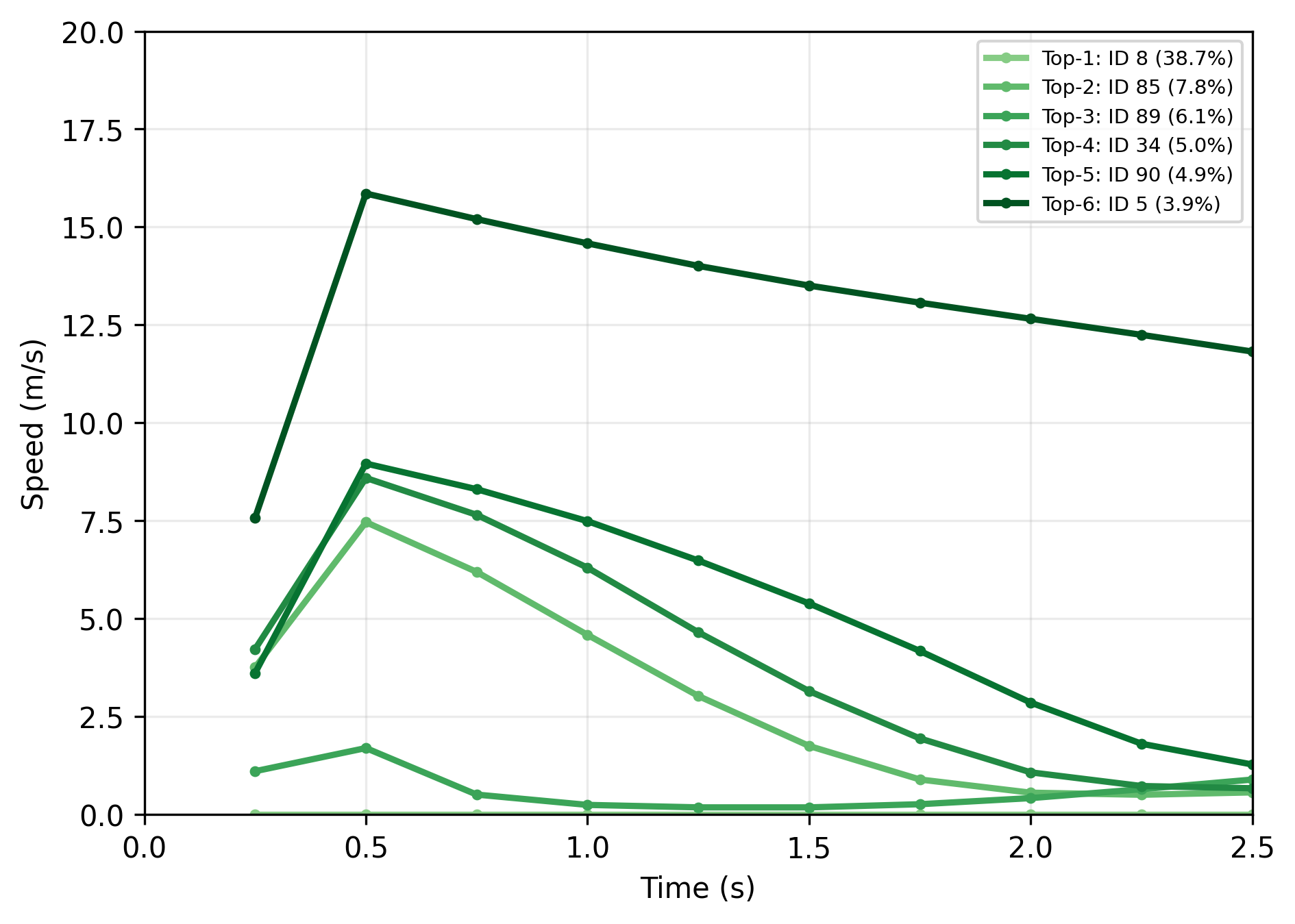}
        &
        \anchorpanel{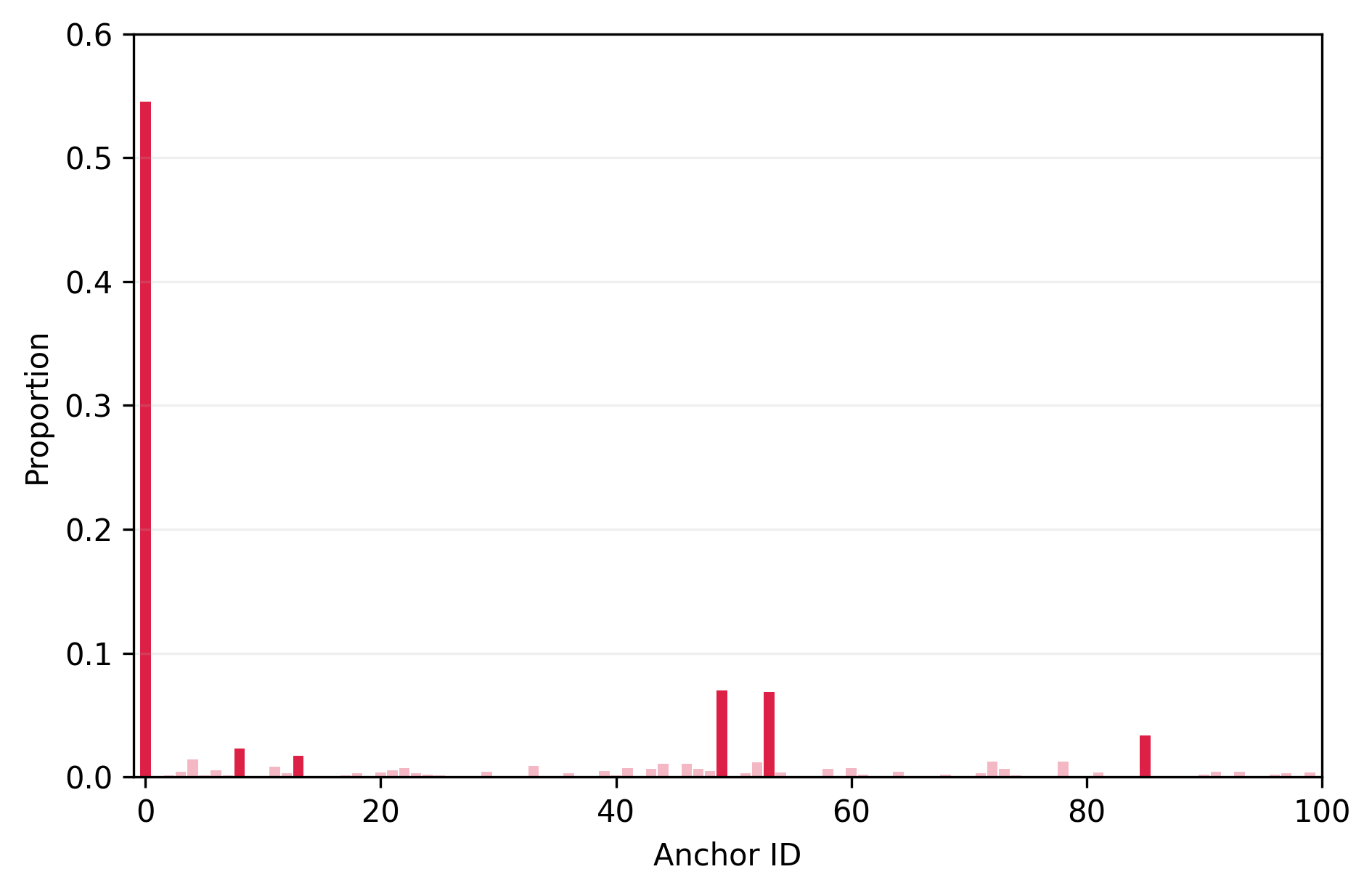}
        &
        \anchorpanel{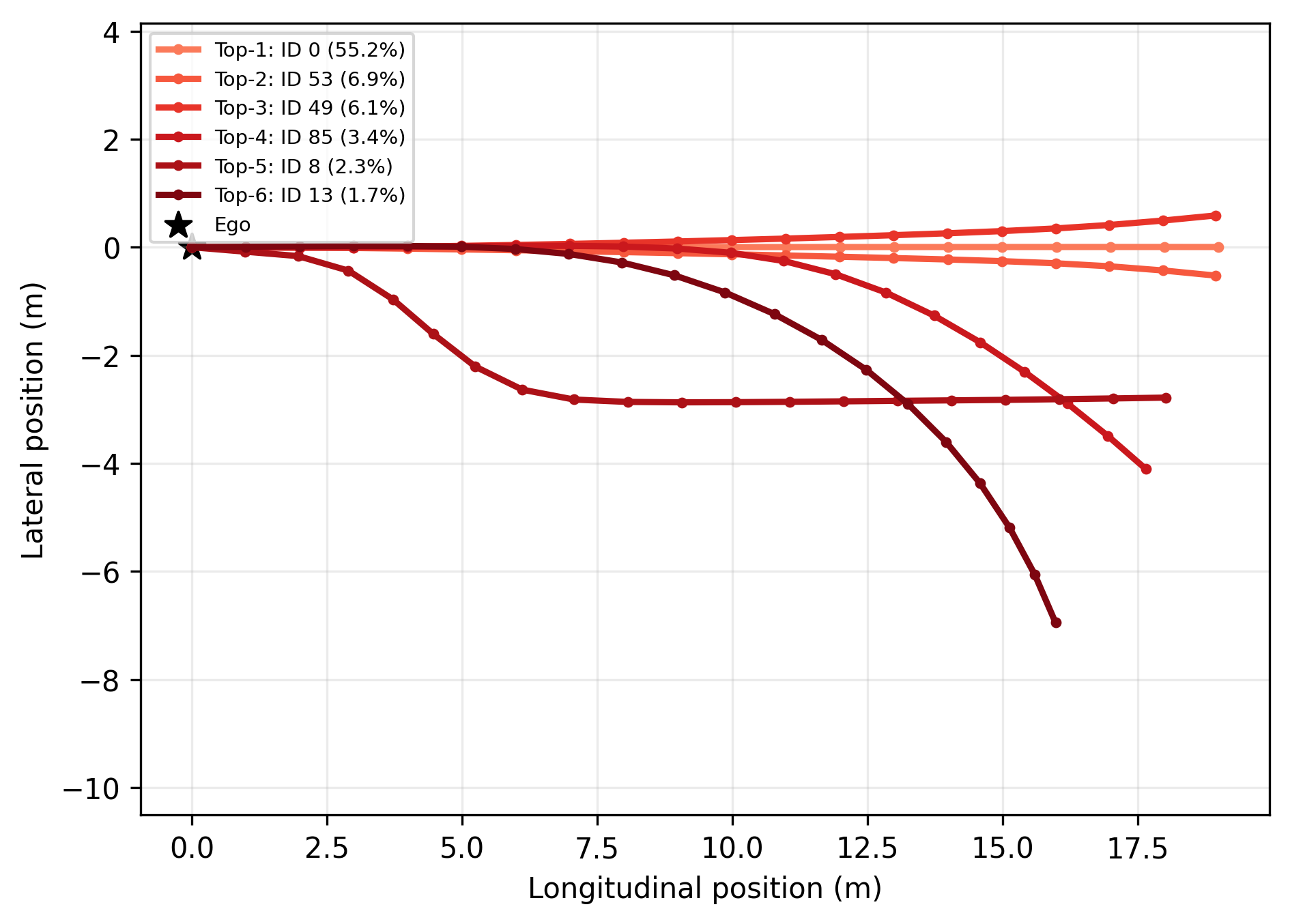}
        \\[0.5mm]

        \anchorrowlabel{Faster}
        &
        \anchorpanel{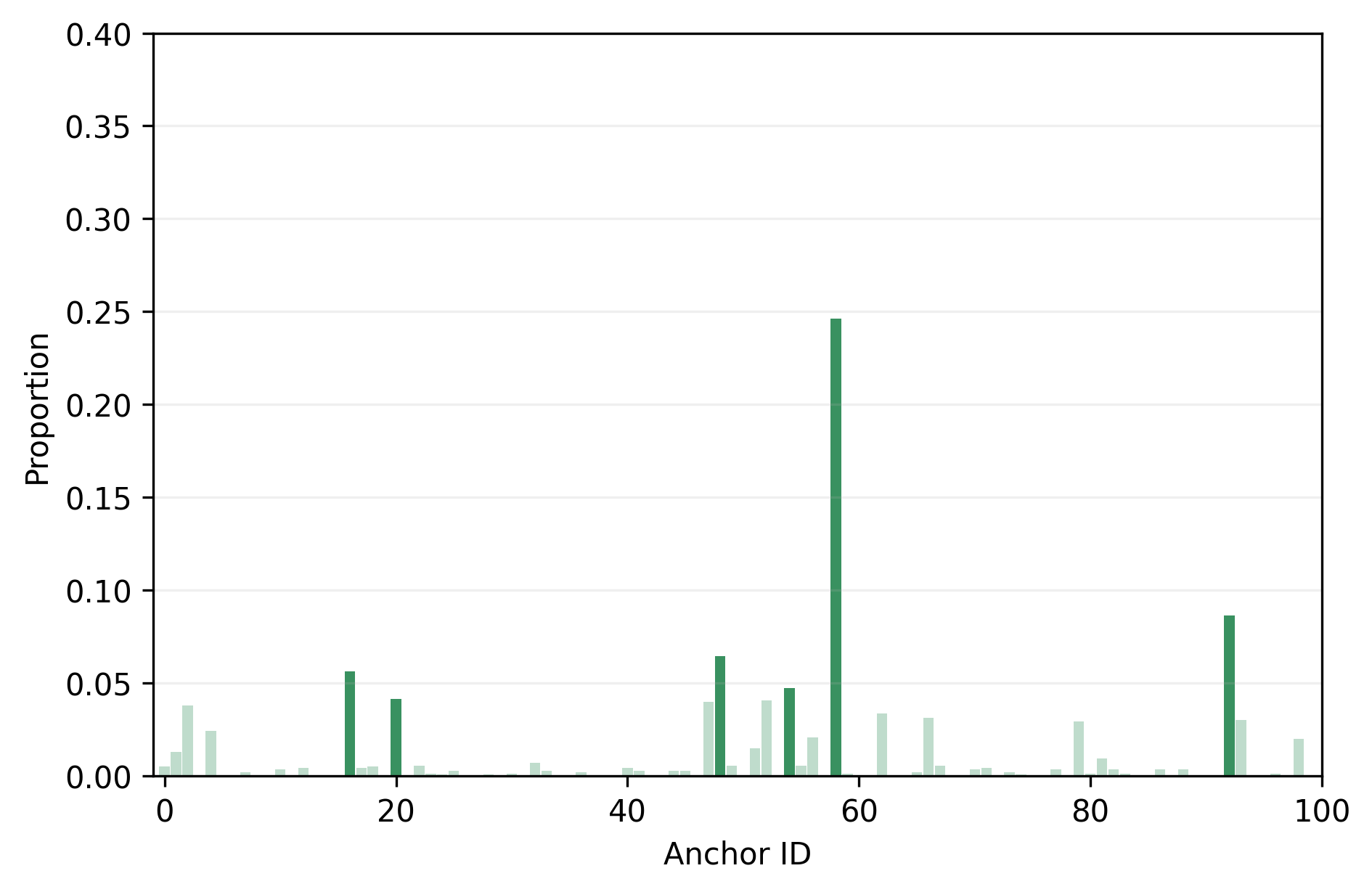}
        &
        \anchorpanel{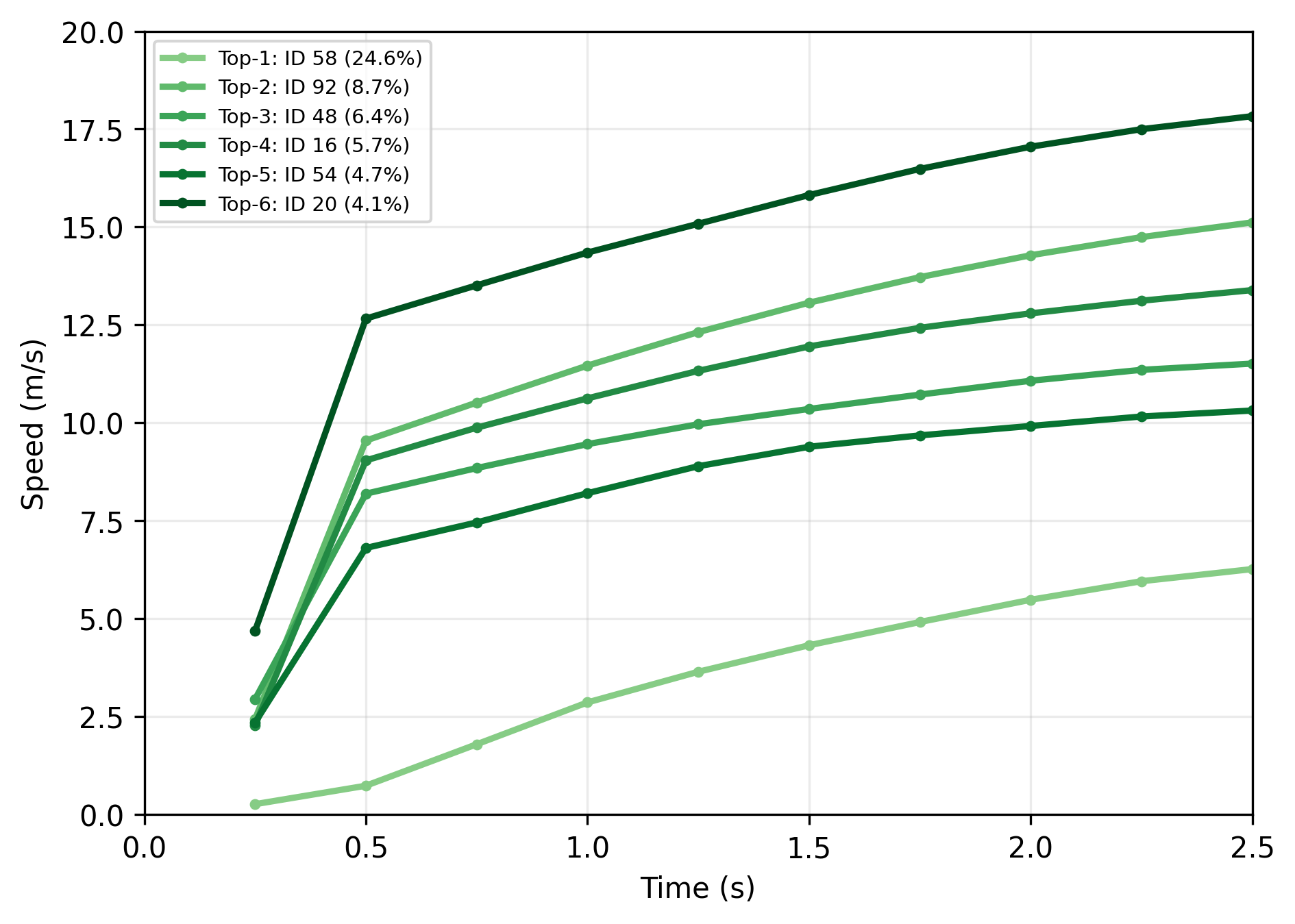}
        &
        \anchorpanel{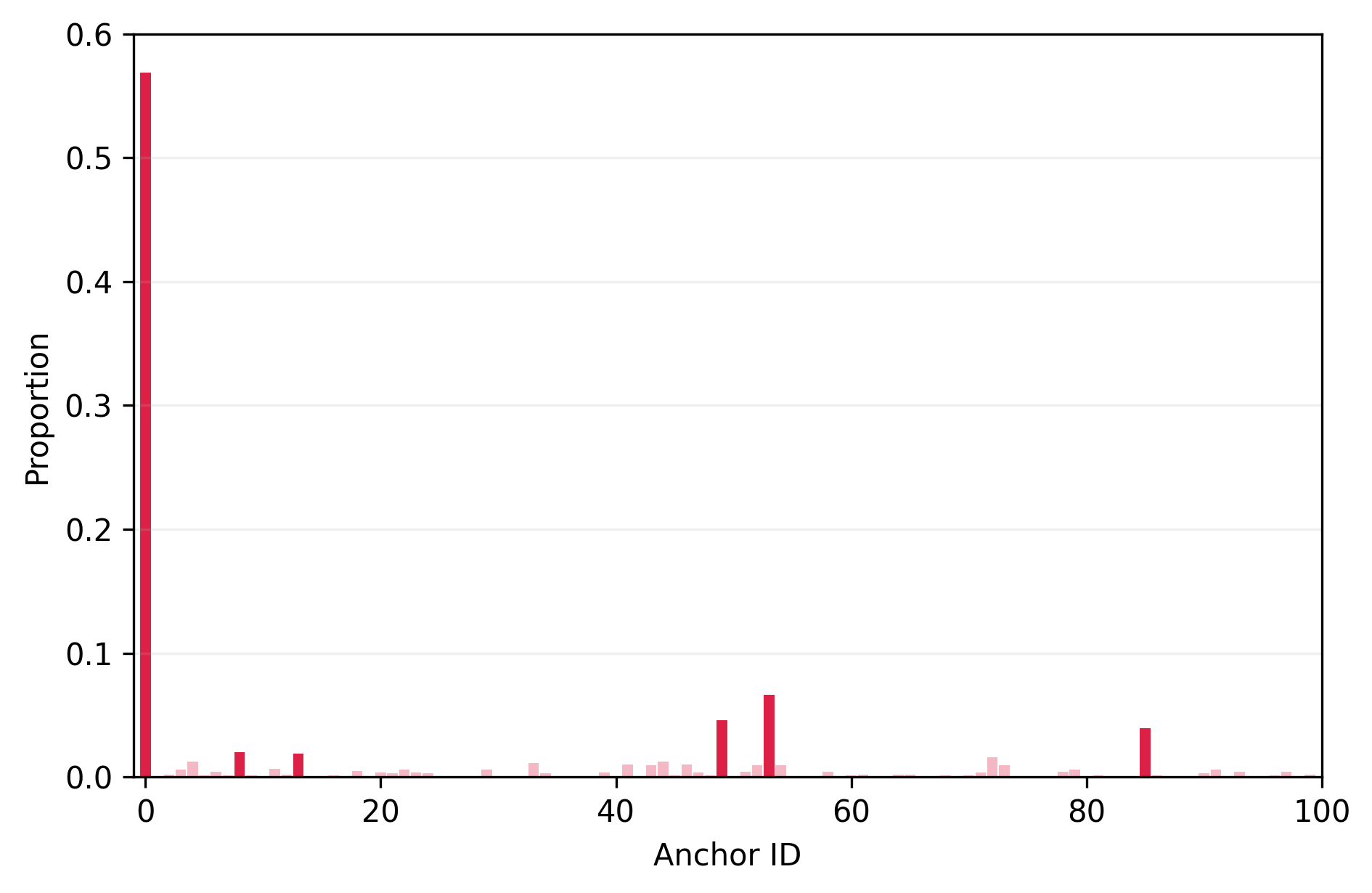}
        &
        \anchorpanel{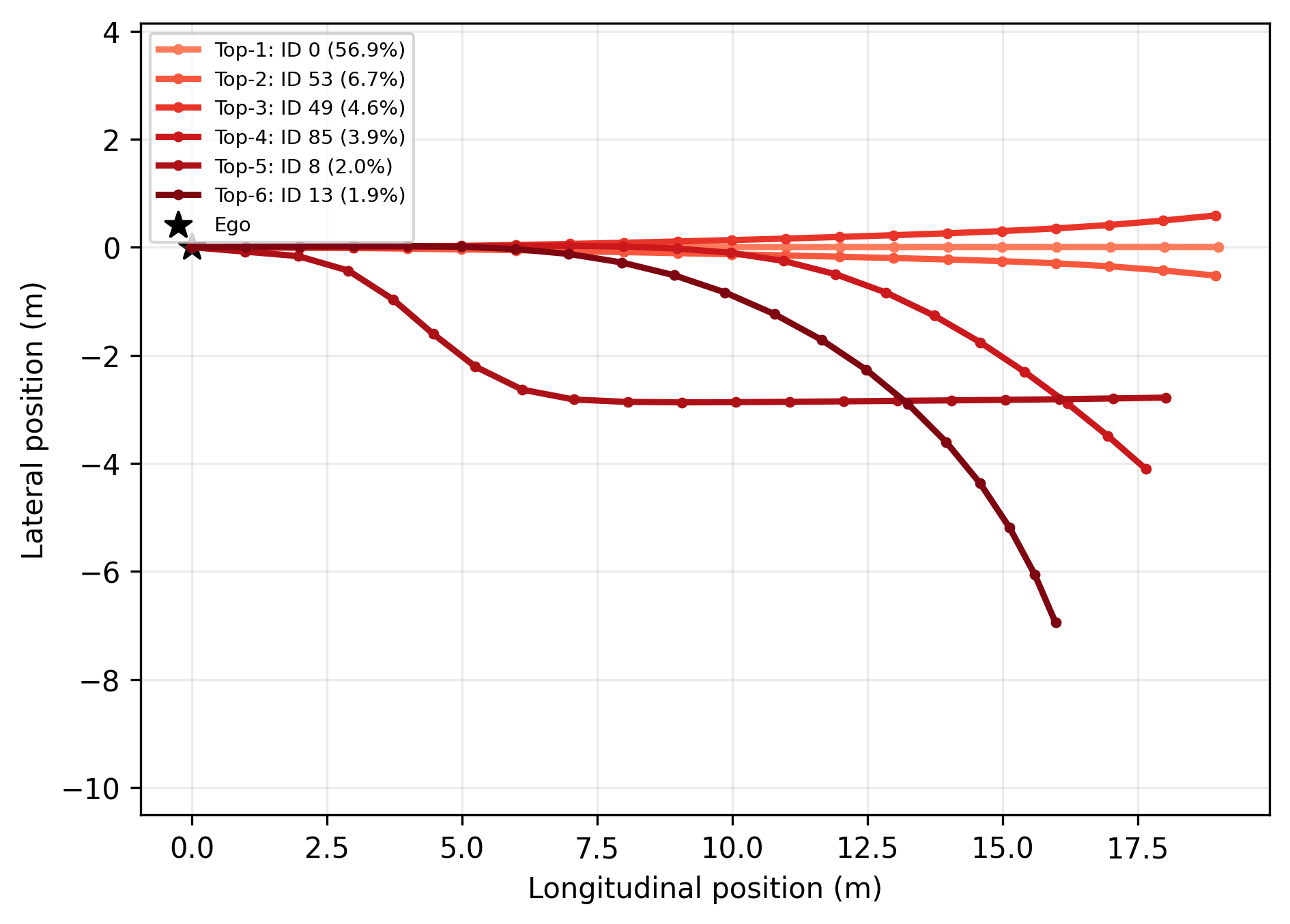}
        \\[0.5mm]

        \anchorrowlabel{Lane Change}
        &
        \anchorpanel{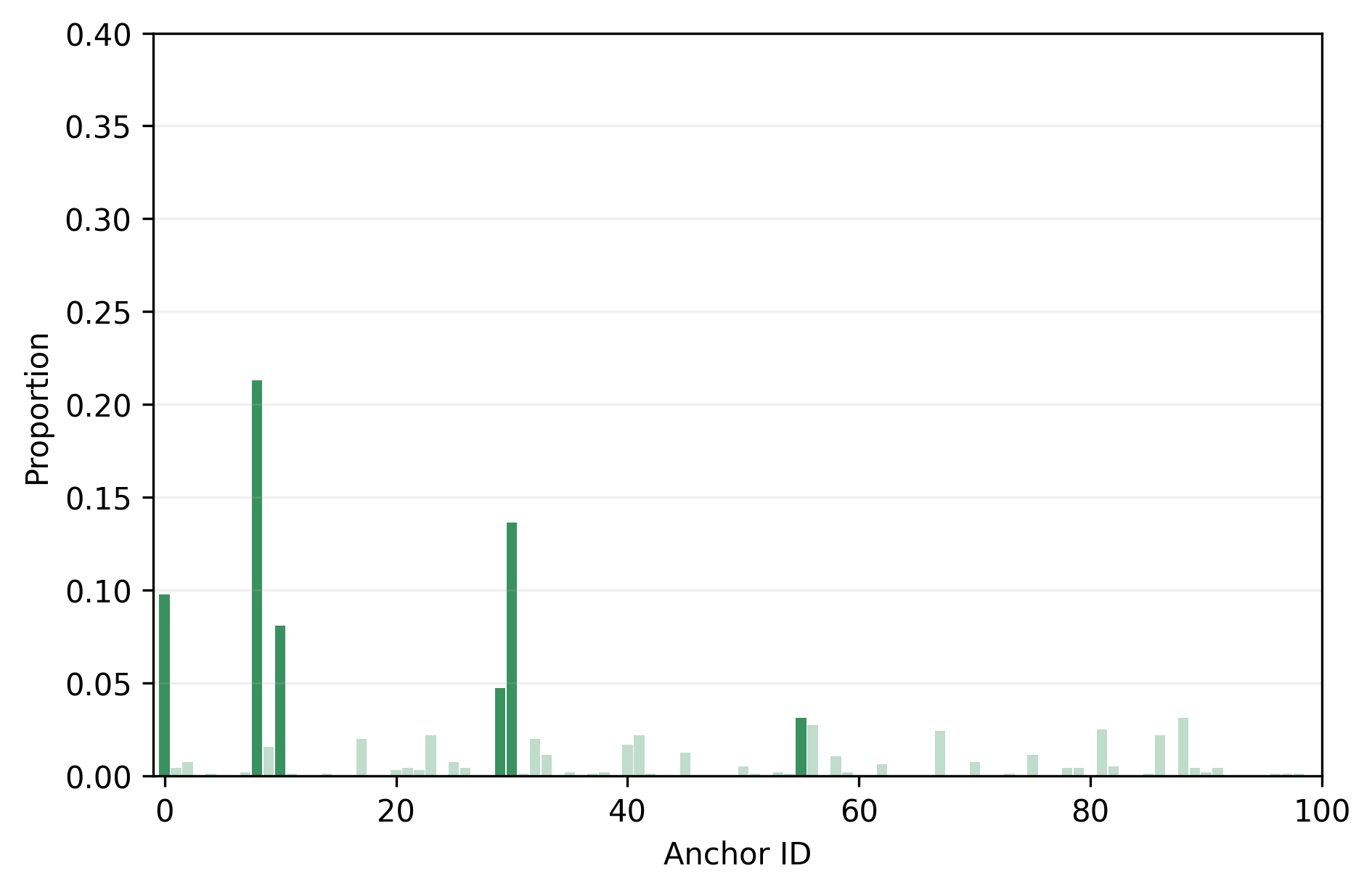}
        &
        \anchorpanel{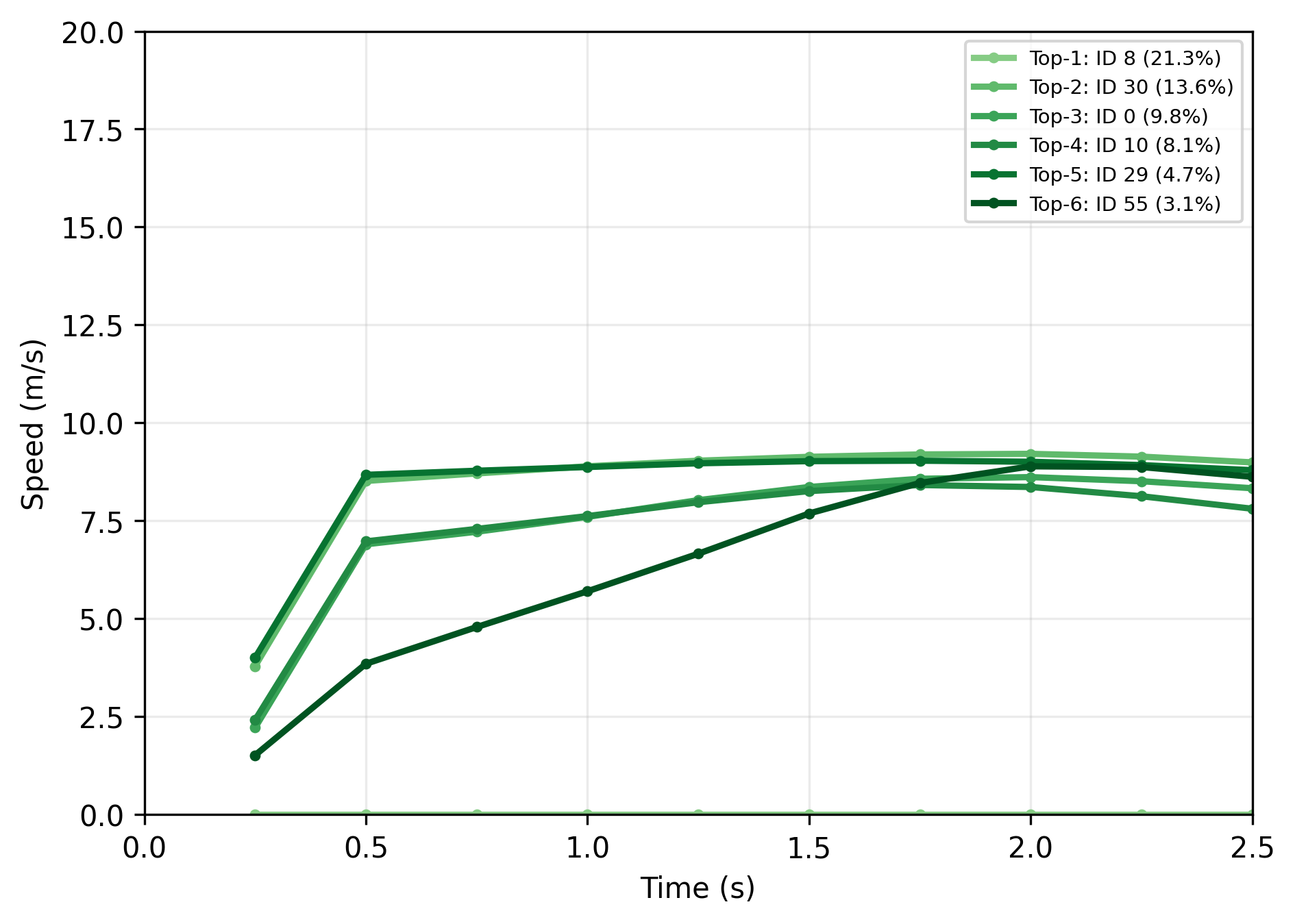}
        &
        \anchorpanel{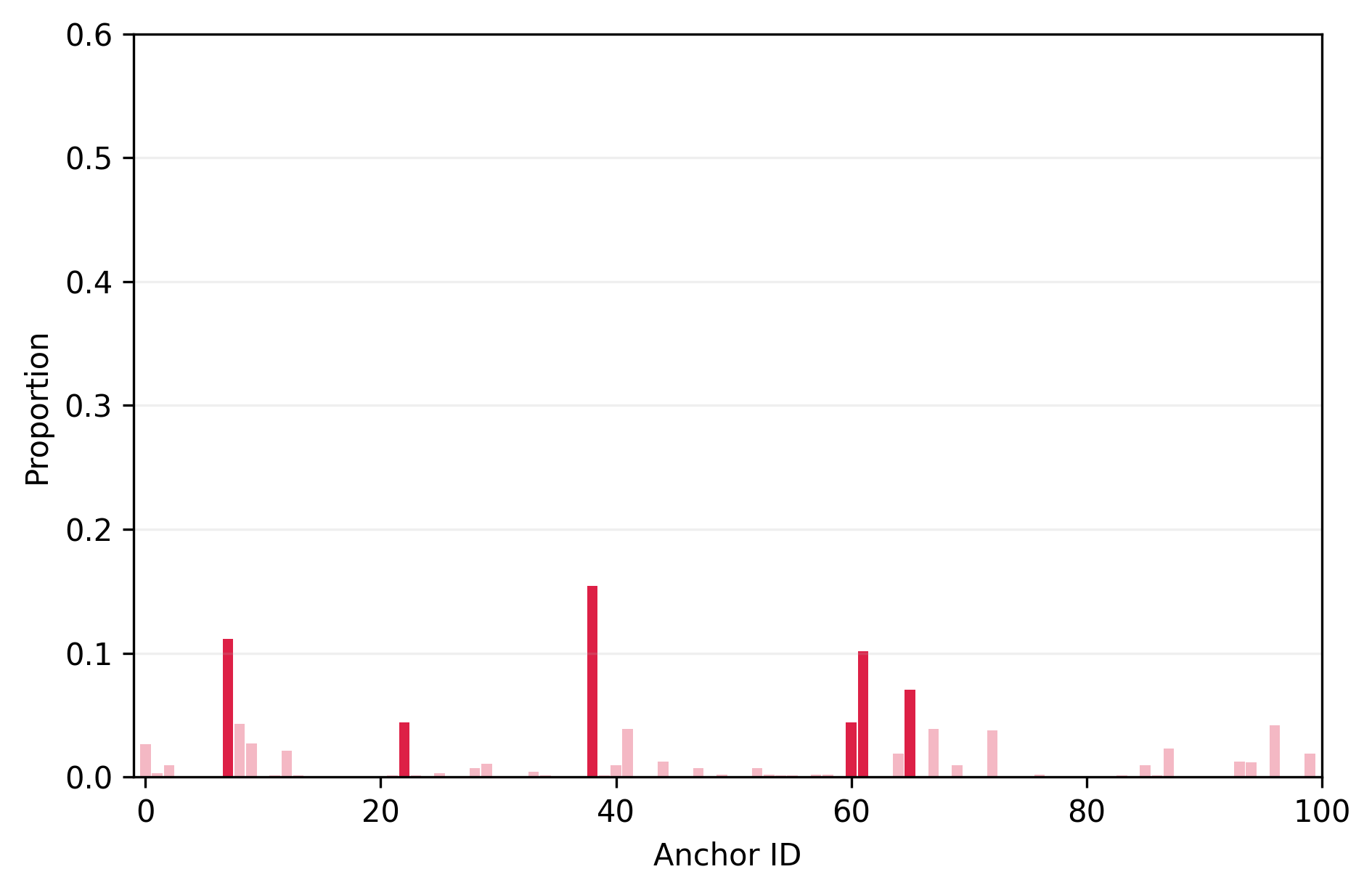}
        &
        \anchorpanel{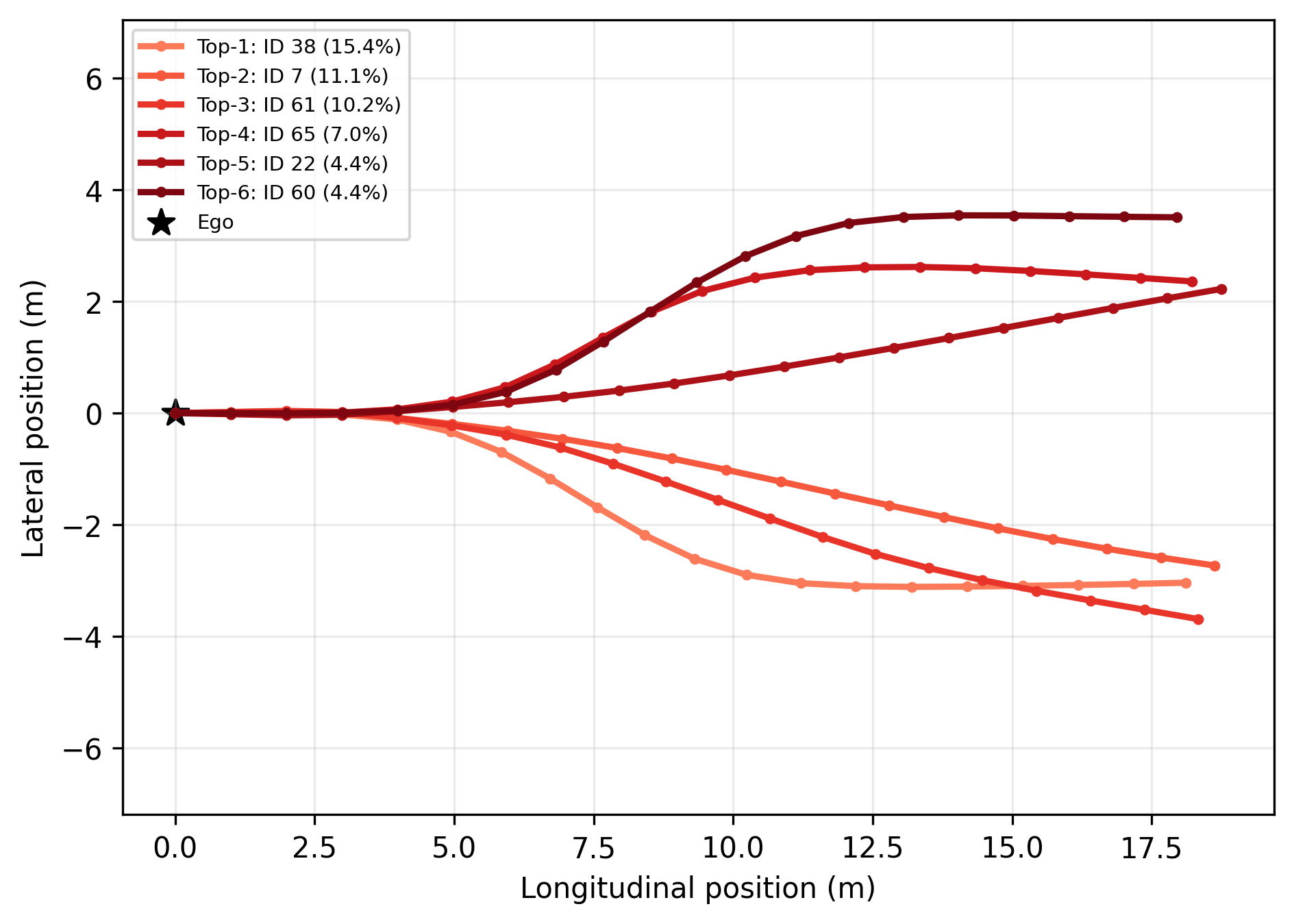}
        \\
    \end{tabular}

    \endgroup

    \caption{
        Instruction-conditioned distributions and representative patterns of
        predicted trajectory anchors on the Action Dreaming dataset.
        The histograms show normalized top-1 selection frequencies, while the
        adjacent panels visualize the six most frequently selected Velocity and
        Direction anchors as speed profiles and spatial paths, respectively.
    }
    \label{fig:instruction_conditioned_anchor_analysis}
\end{figure*}

\section{Instruction-Conditioned Anchor Analysis}
\label{subsec:instruction_conditioned_anchors}

We further analyze whether the predicted trajectory anchors carry the driving
decisions produced by the VLA. We consider four representative commands from
the Action Dreaming dataset: \emph{stop}, \emph{slower}, \emph{faster}, and
\emph{lane change}. Only samples labeled as
\texttt{<INSTRUCTION\_FOLLOWING>} are included, thereby excluding cases in
which safety constraints override the original instruction. After filtering,
the \emph{stop}, \emph{slower}, \emph{faster}, and \emph{lane change} subsets
contain 1,187, 1,423, 1,398, and 953 samples, respectively, resulting in a
total of 4,961 samples.

For each command, we compute the normalized distribution of the top-1 anchor
selected by DAAR. We additionally visualize the six anchors that are most
frequently selected as top-1 predictions under each command, together with
their selection proportions. Velocity anchors are converted into speed
profiles over the $2.5\,\mathrm{s}$ prediction horizon, whereas Direction
anchors are visualized as spatial paths.

As shown in
Figure~\ref{fig:instruction_conditioned_anchor_analysis}, the predicted anchor
distributions exhibit clear instruction-conditioned patterns. The
\emph{stop}, \emph{slower}, and \emph{faster} commands primarily alter the
Velocity-anchor distribution. Their dominant anchors correspond to different
speed-evolution patterns, including stopping, deceleration, and acceleration.
Meanwhile, the Direction-anchor distributions remain comparatively stable
across these longitudinal commands, since they do not directly modify the
desired path geometry.

In contrast, the \emph{lane change} command activates a substantially different
subset of Direction anchors. Its most frequently selected anchors exhibit
pronounced lateral motion patterns, while the corresponding Velocity anchors
primarily describe the continuous speed evolution during the maneuver.
Together, these results show that the discrete anchors respond systematically
to different driving commands and provide an explicit interface for carrying
VLA decisions into continuous trajectory generation.

\section{Quantitative Analysis of the Trajectory-Pattern Codebook and DARF}
\label{sec:quantitative_analysis}

Based on the dual-trajectory representation described above, we provide two
complementary quantitative analyses. First, we study how the
trajectory-pattern codebook size affects anchor coverage, geometric
representation, and closed-loop driving performance. Second, we analyze how
DARF refines the predicted decision anchors and compare the transport lengths
of full-trajectory and residual-flow generation. All open-loop metrics are
computed over 36,720 evaluation samples from the SimLingo evaluation dataset.

\begin{figure*}[t]
    \centering
    \includegraphics[width=\textwidth]{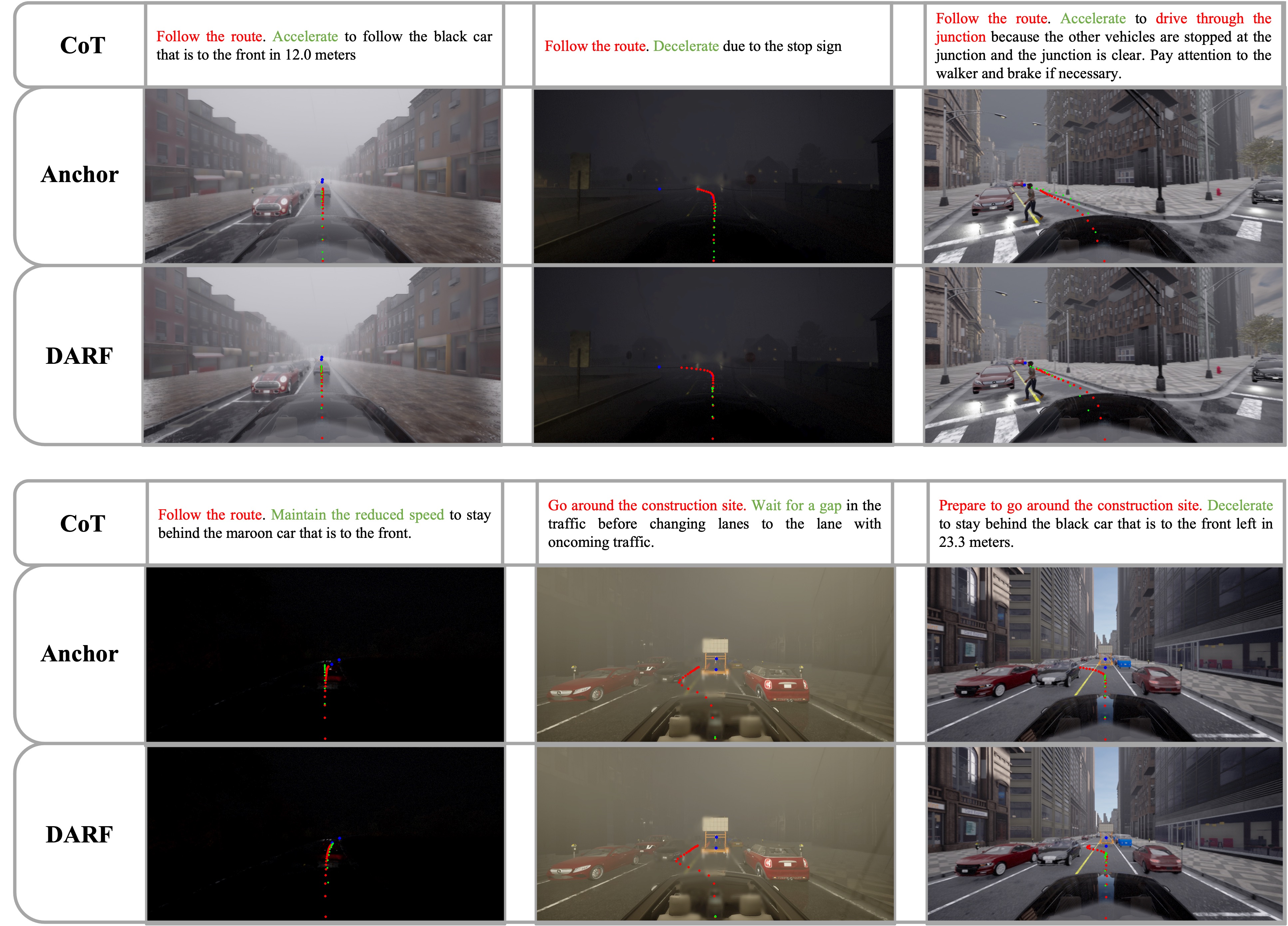}
    \caption{
    Qualitative visualization of AnchorVLA from the Bench2Drive closed-loop evaluation.
    Each case presents the CoT-based driving decision, the trajectory-pattern anchors predicted by the anchor-decision model, and the final trajectories generated by \textbf{Decision-Anchored Residual Flow} (DARF).
    The predicted anchors follow the behavior-level decisions produced by the VLA and serve as explicit decision interfaces, while DARF generates continuous and executable trajectories in the corresponding anchor-defined residual spaces.
    }
    \label{fig:qualitative_visualization}
\end{figure*}

\begin{figure*}[t]
    \centering
    \includegraphics[width=\textwidth]{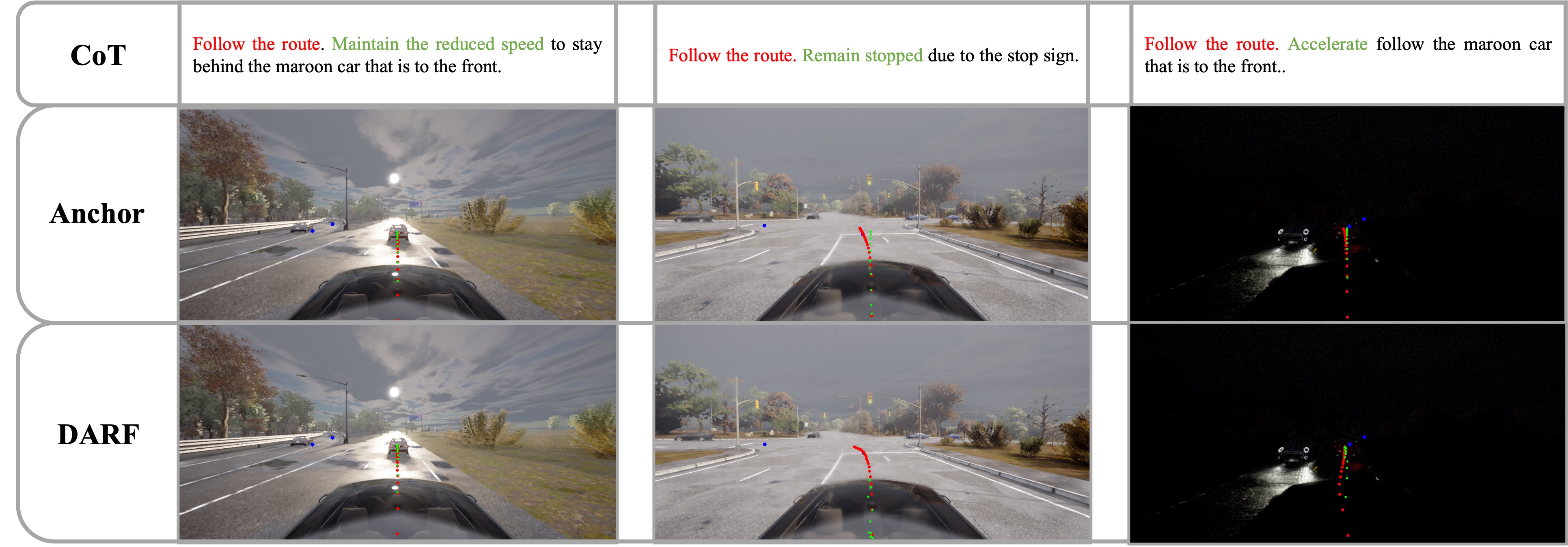}
    \caption{
    Representative failure cases observed during the Bench2Drive closed-loop evaluation.
    The examples include off-route anchor prediction, inconsistency between the predicted anchor and the language instruction, and inaccurate residual refinement.
    An incorrect anchor may propagate an undesired behavior pattern to the final trajectory, whereas imperfect residual prediction may introduce local geometric deviations despite a reasonable anchor-level decision.
    }
    \label{fig:failure_cases}
\end{figure*}

\subsection{Trajectory-Pattern Codebook Size}
\label{subsec:codebook_size}

We evaluate trajectory-pattern codebooks with
$K\in\{50,100,200\}$ for both the Direction and Velocity trajectories while
keeping the number of predicted candidate anchors fixed at $M=6$.
Recall@$m$ measures whether the globally nearest codebook anchor is included
in the predicted top-$m$ candidates. minADE@$m$ measures the minimum trajectory
displacement error between the ground-truth trajectory and the predicted
top-$m$ anchors.

\begin{table*}[t]
    \centering
    \renewcommand{\arraystretch}{1.15}
    \setlength{\tabcolsep}{7pt}

    \resizebox{\textwidth}{!}{
    \begin{tabular}{cc|ccc|ccc|cc}
        \toprule
        \makecell[c]{Codebook\\Size}
        & \makecell[c]{Trajectory\\Type}
        & \makecell[c]{Recall@1\\(\%)}
        & \makecell[c]{Recall@3\\(\%)}
        & \makecell[c]{Recall@6\\(\%)}
        & \makecell[c]{minADE\\@1}
        & \makecell[c]{minADE\\@3}
        & \makecell[c]{minADE\\@6}
        & \makecell[c]{Driving\\Score}
        & \makecell[c]{Success\\Rate}
        \\
        \midrule

        50
        & Direction
        & \textbf{92.80}
        & \textbf{98.32}
        & \textbf{99.35}
        & 0.2042
        & 0.1824
        & 0.1758
        & \multirow{2}{*}{\textbf{90.43}}
        & \multirow{2}{*}{74.43}
        \\
        50
        & Velocity
        & \textbf{80.92}
        & \textbf{96.70}
        & \textbf{99.33}
        & 0.9142
        & 0.7654
        & 0.7418
        &
        &
        \\
        \midrule

        100
        & Direction
        & 90.96
        & 97.47
        & 98.78
        & 0.1600
        & 0.1407
        & 0.1335
        & \multirow{2}{*}{89.92}
        & \multirow{2}{*}{\textbf{77.28}}
        \\
        100
        & Velocity
        & 79.04
        & 95.35
        & 98.92
        & 0.7548
        & 0.6250
        & 0.5956
        &
        &
        \\
        \midrule

        200
        & Direction
        & 88.06
        & 96.49
        & 98.09
        & \textbf{0.1365}
        & \textbf{0.1177}
        & \textbf{0.1096}
        & \multirow{2}{*}{89.51}
        & \multirow{2}{*}{73.64}
        \\
        200
        & Velocity
        & 73.35
        & 91.29
        & 96.96
        & \textbf{0.6928}
        & \textbf{0.5484}
        & \textbf{0.5035}
        &
        &
        \\
        \bottomrule
    \end{tabular}
    }
    \caption{
        Effect of the trajectory-pattern codebook size on open-loop anchor
        coverage and Bench2Drive closed-loop performance. Each trajectory type
        uses an independent codebook of the indicated size. Recall and minADE
        are diagnostic open-loop metrics computed using logged ground-truth
        trajectories, while Driving Score and Success Rate are measured for
        the complete dual-trajectory model through closed-loop evaluation.
    }
    \label{tab:codebook_size_analysis}
\end{table*}

As shown in Table~\ref{tab:codebook_size_analysis}, increasing the codebook
size improves its geometric representation resolution. From $K=50$ to
$K=200$, minADE@$6$ decreases from 0.1758 to 0.1096 for the Direction
trajectory and from 0.7418 to 0.5035 for the Velocity trajectory. However,
finer discretization also makes anchor prediction more challenging.
Recall@$6$ decreases from 99.35\% to 98.09\% for Direction and from 99.33\%
to 96.96\% for Velocity.

The closed-loop results exhibit a non-monotonic relationship with codebook
size. The coarse $K=50$ codebook facilitates anchor prediction and achieves
the highest Driving Score of 90.43, but its limited motion-pattern resolution
results in a lower Success Rate of 74.43. Increasing the codebook size to
$K=100$ improves motion-pattern resolution while maintaining reliable
top-$M$ candidate coverage, yielding the highest Success Rate of 77.28.

When the codebook is further enlarged to $K=200$, its open-loop geometric
accuracy continues to improve, but its closed-loop Success Rate decreases to
73.64. The finer partition distributes the prediction probability over a
larger number of nearby and geometrically similar anchors, potentially
reducing the diversity of the top-$M$ candidate set and making closed-loop
anchor selection more difficult. Overall, these results support $K=100$, as
used in our main experiments, as a balanced setting between motion-pattern
resolution and reliable decision-anchor prediction.

\subsection{DARF Refinement and Transport Analysis}
\label{subsec:darf_transport}

We next analyze whether DARF performs continuous refinement around the
selected decision anchors and whether residual-flow generation reduces the
empirical transport length. For the predicted top-1 anchor
$a_{\hat{k}}$ and its DARF-refined trajectory $\hat{\tau}$, we define the
relative anchor-to-output deviation as
\begin{equation}
    \delta
    =
    \frac{\|\hat{\tau}-a_{\hat{k}}\|_2}
    {\|a_{\hat{k}}\|_2+\epsilon},
    \label{eq:anchor_output_deviation}
\end{equation}
where $\epsilon$ is a small constant for numerical stability. We report the
median relative deviation because near-stationary Velocity anchors may have
norms close to zero and produce disproportionately large relative values. We
further compare the RMS transport lengths of full-trajectory flow and
residual flow under the same noise distribution.

\begin{table*}[t]
    \centering

    \renewcommand{\arraystretch}{1.15}
    \setlength{\tabcolsep}{14pt}

    \begin{tabular}{cc|ccc}
        \toprule
        \makecell[c]{Codebook\\Size}
        & \makecell[c]{Trajectory\\Type}
        & \makecell[c]{Median Relative\\
          Anchor-to-Output Deviation (\%)}
        & \makecell[c]{Full-Trajectory\\RMS Transport}
        & \makecell[c]{Residual-Flow\\RMS Transport}
        \\
        \midrule
        50
        & Direction
        & 1.41
        & 48.83
        & 1.88
        \\
        50
        & Velocity
        & 15.77
        & 30.16
        & 5.14
        \\
        \midrule
        100
        & Direction
        & 0.81
        & 48.82
        & 1.46
        \\
        100
        & Velocity
        & 12.58
        & 29.92
        & 4.08
        \\
        \midrule
        200
        & Direction
        & 0.62
        & 48.81
        & 1.19
        \\
        200
        & Velocity
        & 11.81
        & 29.83
        & 3.60
        \\
        \bottomrule
    \end{tabular}
    \caption{
        Quantitative analysis of DARF refinement and transport length using
        the predicted top-1 anchors. Relative Anchor-to-Output Deviation
        measures the magnitude of DARF refinement relative to the selected
        anchor and is reported as the median percentage. Full-Trajectory and
        Residual-Flow RMS Transport measure the empirical transport lengths
        under the same noise distribution.
    }
    \label{tab:darf_quantitative_analysis}
\end{table*}

As shown in Table~\ref{tab:darf_quantitative_analysis}, DARF performs moderate
continuous refinement around the predicted decision anchors. For the Direction
trajectory, the median relative anchor-to-output deviation ranges from 0.62\%
to 1.41\%. For the Velocity trajectory, the median relative deviation ranges
from 11.81\% to 15.77\%. These results show that the refined trajectories
preserve the coarse motion patterns represented by the anchors while DARF
adjusts their continuous execution according to the current driving context.

The transport results further support the residual-flow formulation. For the
Direction trajectory, the RMS transport length is reduced from approximately
48.8 in full-trajectory space to 1.19--1.88 in the corresponding
anchor-defined residual spaces, representing a reduction of approximately
96.1--97.6\%. For the Velocity trajectory, the RMS transport length is reduced
from approximately 30.0 to 3.60--5.14, representing a reduction of
approximately 83.0--87.9\%. These results demonstrate that trajectory-pattern
anchors provide compact references for residual generation and substantially
shorten the empirical transport path.

\section{Visualization Analysis}
\label{sec:visualization_analysis}

We present representative qualitative results and failure cases collected
from actual closed-loop rollouts on the Bench2Drive benchmark.

\subsection{Qualitative Results}
\label{subsec:qualitative_results}

Figure~\ref{fig:qualitative_visualization} presents representative closed-loop
examples of the VLA decisions, the predicted Direction and Velocity anchors,
and the trajectories refined by DARF. The selected anchors capture the coarse
motion patterns associated with the VLA decisions, while DARF adjusts their
continuous geometry according to the current driving context. These examples
qualitatively illustrate the decision--execution decomposition of AnchorVLA.

\subsection{Failure Case Analysis}
\label{subsec:failure_cases}

Figure~\ref{fig:failure_cases} presents representative failure cases from the
Bench2Drive closed-loop evaluation. The first two cases correspond to
anchor-level errors, including off-route and language-inconsistent anchor
predictions, which propagate undesired motion patterns to the final
trajectory. The third case corresponds to a residual-level error, where
inaccurate DARF refinement introduces local geometric deviations despite a
reasonable anchor. These examples illustrate the two principal failure
sources of AnchorVLA: anchor selection and continuous residual refinement.

\section{Decision-Anchored Residual Flow Decoder}
\label{sec:darf_decoder}

Given the top-$M$ trajectory-pattern anchors predicted by DAAR, the
\textbf{Decision-Anchored Residual Flow} (DARF) decoder generates a continuous
residual for each candidate anchor and predicts an anchor--context compatibility
score for candidate selection.

As introduced above, our SimLingo-based implementation contains Direction
and Velocity trajectories with $T=20$ and $T=10$ future waypoints,
respectively. For clarity, we describe the DARF
decoder below for a generic trajectory type and omit the trajectory-type
superscript. The same formulation is instantiated for both trajectory types.

For each candidate anchor $a_k$, $k\in\mathcal{I}_M$, the decoder receives the
noisy residual state $z_t^k\in\mathbb{R}^{T\times2}$, the flow timestep $t$,
the candidate anchor $a_k\in\mathbb{R}^{T\times2}$, and the multimodal context
features $H_x$ extracted from the visual, language, and navigation inputs.
All top-$M$ candidates within the same trajectory type share the decoder
parameters. The Direction and Velocity trajectories are processed using the
same decoder design, while their input and output projection dimensions are
configured separately to accommodate their different prediction horizons.

\paragraph{Input encoding.}
The noisy residual trajectory and candidate anchor are encoded independently.
Each 2D waypoint coordinate is mapped to a sine--cosine embedding. The
embeddings over the $T$ future waypoints are then flattened and projected into
the decoder hidden dimension using MLPs. This produces the noisy-residual
feature $h_z^k$ and two anchor features,
$h_{a,k}^{\mathrm{pos}}$ and $h_{a,k}^{\mathrm{conf}}$, which are used for
residual-flow prediction and confidence prediction, respectively. The flow
timestep is encoded using a sinusoidal time embedding followed by an MLP,
producing the time feature $h_t$.

\paragraph{Residual-flow branch.}
The residual-flow branch initializes its hidden state with $h_z^k$. At each
decoder layer, the hidden state is conditioned on the flow timestep through
feature-wise affine modulation:
\begin{equation}
    \mathrm{Mod}(h,h_t)
    =
    h\odot\left(1+s(h_t)\right)+b(h_t),
    \label{eq:feature_modulation}
\end{equation}
where $s(\cdot)$ and $b(\cdot)$ are learned MLP projections. The anchor feature
is subsequently added as a query bias:
\begin{equation}
    q_l^k
    =
    \mathrm{Mod}(h_l^k,h_t)+h_{a,k}^{\mathrm{pos}}.
    \label{eq:anchor_query_bias}
\end{equation}
The query $q_l^k$ attends to the multimodal context $H_x$ through
cross-attention, followed by a residual connection, LayerNorm, and an FFN.
After the final decoder layer, the prediction head outputs the flow velocity
field:
\begin{equation}
    \hat{v}_{\theta}^{k}
    =
    \hat{v}_{\theta}(z_t^k,t,a_k,H_x)
    \in\mathbb{R}^{T\times2}.
    \label{eq:decoder_velocity_output}
\end{equation}
During training, the predicted flow velocity field is supervised by the flow
matching objective defined in the main paper. During inference, the velocity
field is integrated to generate the continuous residual in the corresponding
anchor-defined space, which is then added to the candidate anchor to reconstruct
the refined trajectory.

\paragraph{Confidence prediction branch.}
The confidence branch evaluates the compatibility between each candidate
anchor and the multimodal driving context. It adopts the WTA supervision
defined in the main paper, following the candidate-selection strategy used by
DiffusionDrive~\cite{liao2025diffusiondrive}.

Because the confidence score evaluates anchor--context compatibility, this
branch does not use the noisy residual state $z_t^k$ or the flow timestep $t$.
It initializes its hidden state with $h_{a,k}^{\mathrm{conf}}$ and attends to
the same multimodal context $H_x$. The resulting feature is updated through
residual connections, LayerNorm, and FFNs. Finally, the confidence head
predicts a scalar score for each candidate:
\begin{equation}
    s_\phi^k
    =
    s_\phi(a_k,H_x)
    \in\mathbb{R}.
    \label{eq:decoder_confidence_output}
\end{equation}

At inference time, candidate selection is performed within each trajectory
type according to the predicted confidence scores. The highest-confidence
Direction trajectory and the highest-confidence Velocity trajectory jointly
form the final dual-trajectory planning output used by the original SimLingo
execution interface.

The two decoder branches serve complementary roles: the residual-flow branch
generates continuous trajectory refinements in the anchor-defined residual
spaces, while the confidence branch provides the candidate scores used for
final trajectory selection.

\subsection{Additional Hyperparameter Settings}

Table~\ref{tab:additional_hyperparameters} presents additional hyperparameters
used in our experiments.

\begin{table}[t]
\centering
\small
\setlength{\tabcolsep}{4pt}
\caption{Additional hyperparameter settings used in experiments.}
\label{tab:additional_hyperparameters}
\begin{tabular}{p{0.49\columnwidth}p{0.41\columnwidth}}
\hline
\textbf{Hyperparameter} & \textbf{Setting} \\
\hline
Optimizer & AdamW \\
Learning rate & $3\times10^{-5}$ \\
Adam coefficients & $\beta_1=0.9,\ \beta_2=0.999$ \\
Weight decay & 0.1 \\
Learning-rate schedule & OneCycleLR \\
Warm-up fraction & 0.05 \\
Gradient clipping & Global norm of 0.3 \\
Numerical precision & FP16 mixed precision \\
Distributed optimization & DeepSpeed ZeRO Stage 2 \\
FP16 loss scale & 32 \\
LoRA rank & 32 \\
LoRA $\alpha$ & 64 \\
LoRA dropout & 0.1 \\
Image augmentation probability & 0.5 \\
Image-shift augmentation probability & 0.5 \\
Direction soft-label temperature & 0.2 \\
Velocity soft-label temperature & 0.5 \\
Noise schedule & Cosine \\
Training noise timesteps & 1,000 \\
\hline
\end{tabular}
\end{table}

\end{document}


\maketitle

\section{Disentangled Path-and-Speed Representation in the SimLingo
Instantiation}
\label{sec:dual_trajectory_representation}

The AnchorVLA formulation in the main paper is defined for a generic
continuous trajectory representation $\tau$. Our implementation follows the
disentangled planning output of SimLingo~\cite{renz2025simlingo}, which
consists of geometric path waypoints and temporal speed waypoints. We refer
to these two output components as Direction and Velocity, respectively.

The Direction component corresponds to the geometric path waypoints in
SimLingo. Its target is constructed from the remaining route expressed in
the current ego coordinate frame and resampled according to accumulated path
length, yielding 20 waypoints at spatial intervals of $1\,\mathrm{m}$. It
therefore represents path geometry and directional evolution as a function
of traveled distance.

The Velocity component corresponds to the temporal speed waypoints in
SimLingo. It consists of 10 future ego positions sampled at intervals of
$0.25\,\mathrm{s}$ over a prediction horizon of $2.5\,\mathrm{s}$. The
temporal displacements between these waypoints encode the evolution of
vehicle speed.

Accordingly, the waypoint indices of the two components are defined over different domains: traveled distance for Direction and elapsed time for Velocity. We therefore employ component-specific anchor spaces and instantiate DAAR-based anchor prediction and DARF-based continuous refinement for each component. Under the autoregressive DAAR instantiation, the model sequentially predicts two component-specific anchor tokens, one for Direction and one for Velocity, from their respective anchor vocabularies. Although their candidates are ranked within their respective output spaces, both components are conditioned on the same multimodal VLA context and high-level reasoning information. During closed-loop execution, the Direction waypoints are used by the lateral controller to determine steering, whereas the temporal displacements of the Velocity waypoints are converted into the desired speed for throttle and braking control, following the original SimLingo planning and control interface.


\newcommand{\anchorfigdir}{Figures}

\newlength{\anchorpanelheight}
\setlength{\anchorpanelheight}{0.155\textwidth}

\newcommand{\anchorpanel}[1]{%
    \parbox[b][\anchorpanelheight][c]{0.218\textwidth}{%
        \centering
        \includegraphics[
            width=0.218\textwidth,
            height=\anchorpanelheight,
            keepaspectratio
        ]{#1}%
    }%
}

\newcommand{\anchorrowlabel}[1]{%
    \parbox[b][\anchorpanelheight][c]{0.028\textwidth}{%
        \centering
        \rotatebox[origin=c]{90}{\small\bfseries #1}%
    }%
}

\newcommand{\anchorcolumnheader}[1]{%
    \parbox[c][2.5em][c]{0.218\textwidth}{%
        \centering
        \scriptsize #1%
    }%
}

\begin{figure*}[t]
    \centering

    \begingroup
    \setlength{\tabcolsep}{1.5pt}
    \renewcommand{\arraystretch}{0.92}

    \begin{tabular}{@{}c@{\hspace{1mm}}cccc@{}}
        &
        \multicolumn{2}{c}{\large\textbf{Velocity Anchors}}
        &
        \multicolumn{2}{c}{\large\textbf{Direction Anchors}}
        \\[-0.5mm]

        &
        \anchorcolumnheader{Top-1 Selection Distribution}
        &
        \anchorcolumnheader{Top-6 Speed Profiles}
        &
        \anchorcolumnheader{Top-1 Selection Distribution}
        &
        \anchorcolumnheader{Top-6 Spatial Paths}
        \\[0.8mm]

        \anchorrowlabel{Stop}
        &
        \anchorpanel{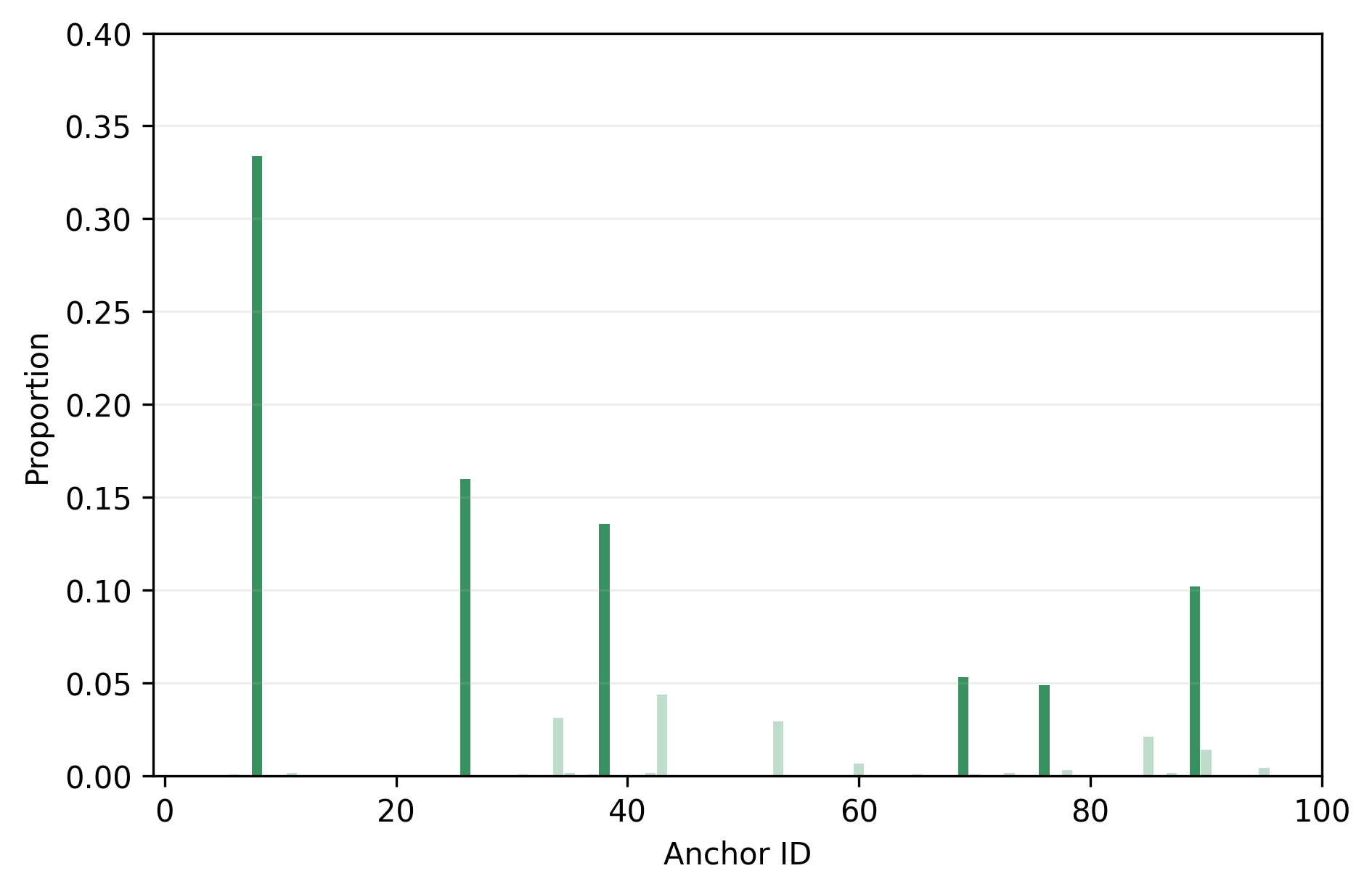}
        &
        \anchorpanel{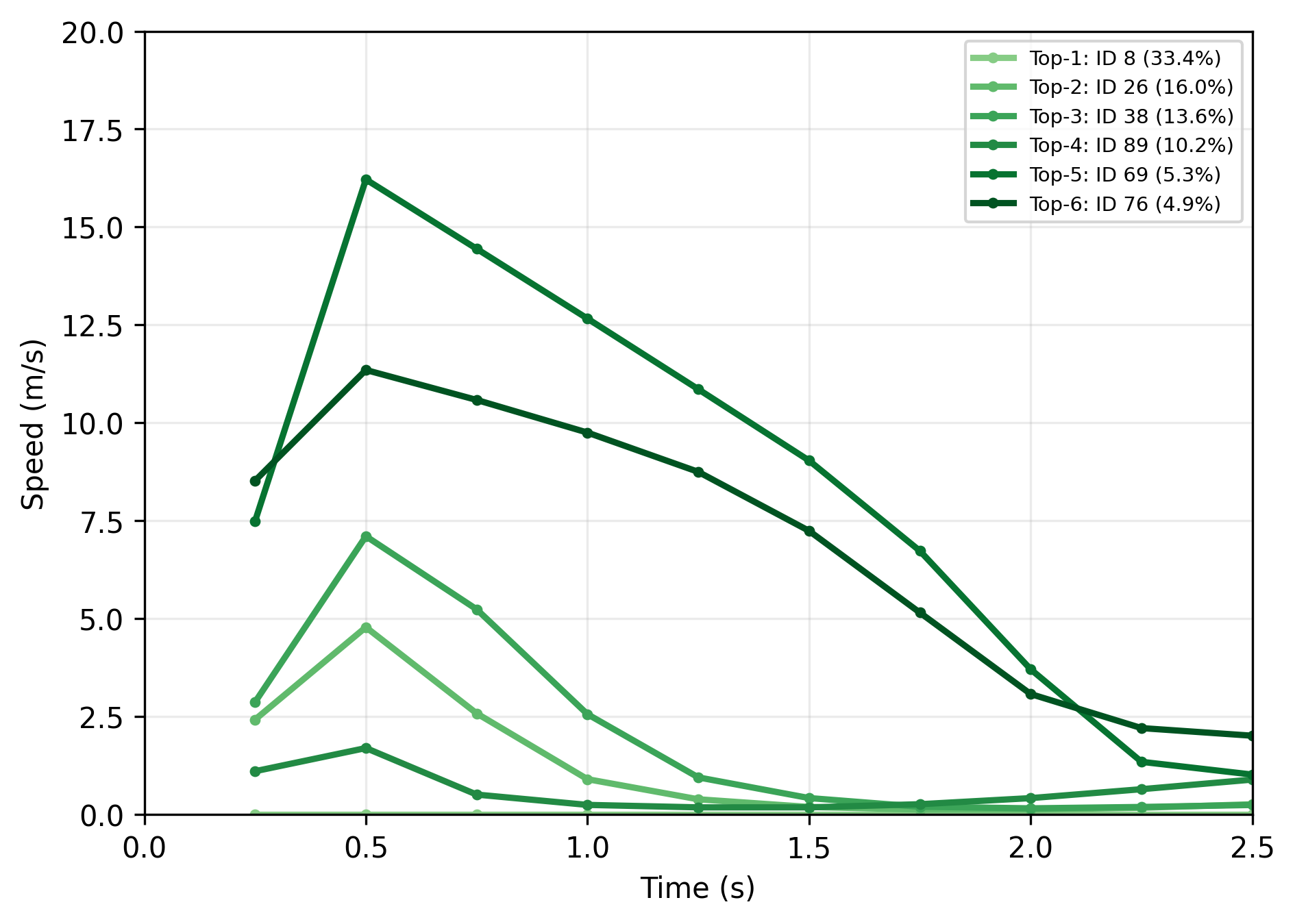}
        &
        \anchorpanel{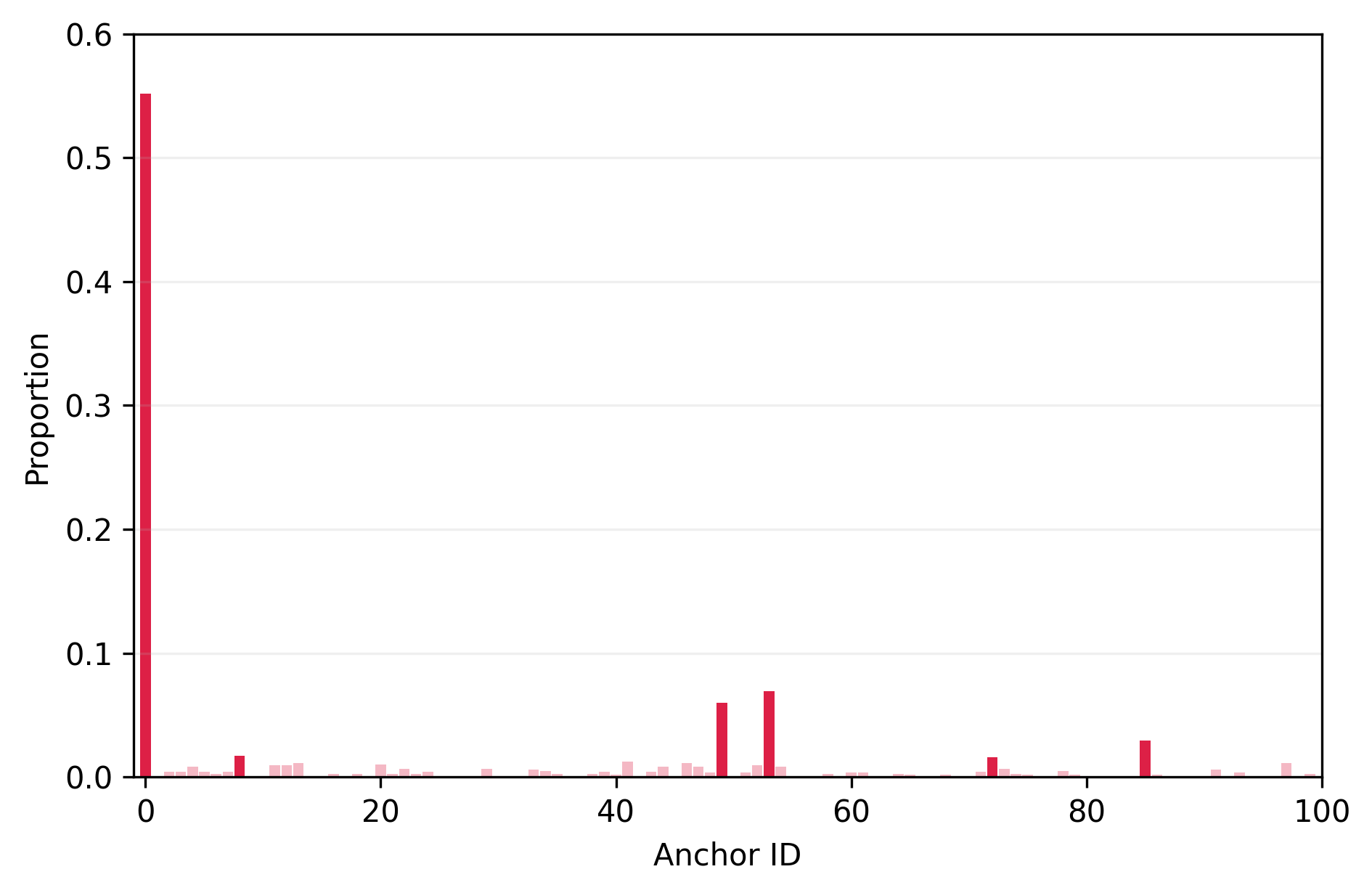}
        &
        \anchorpanel{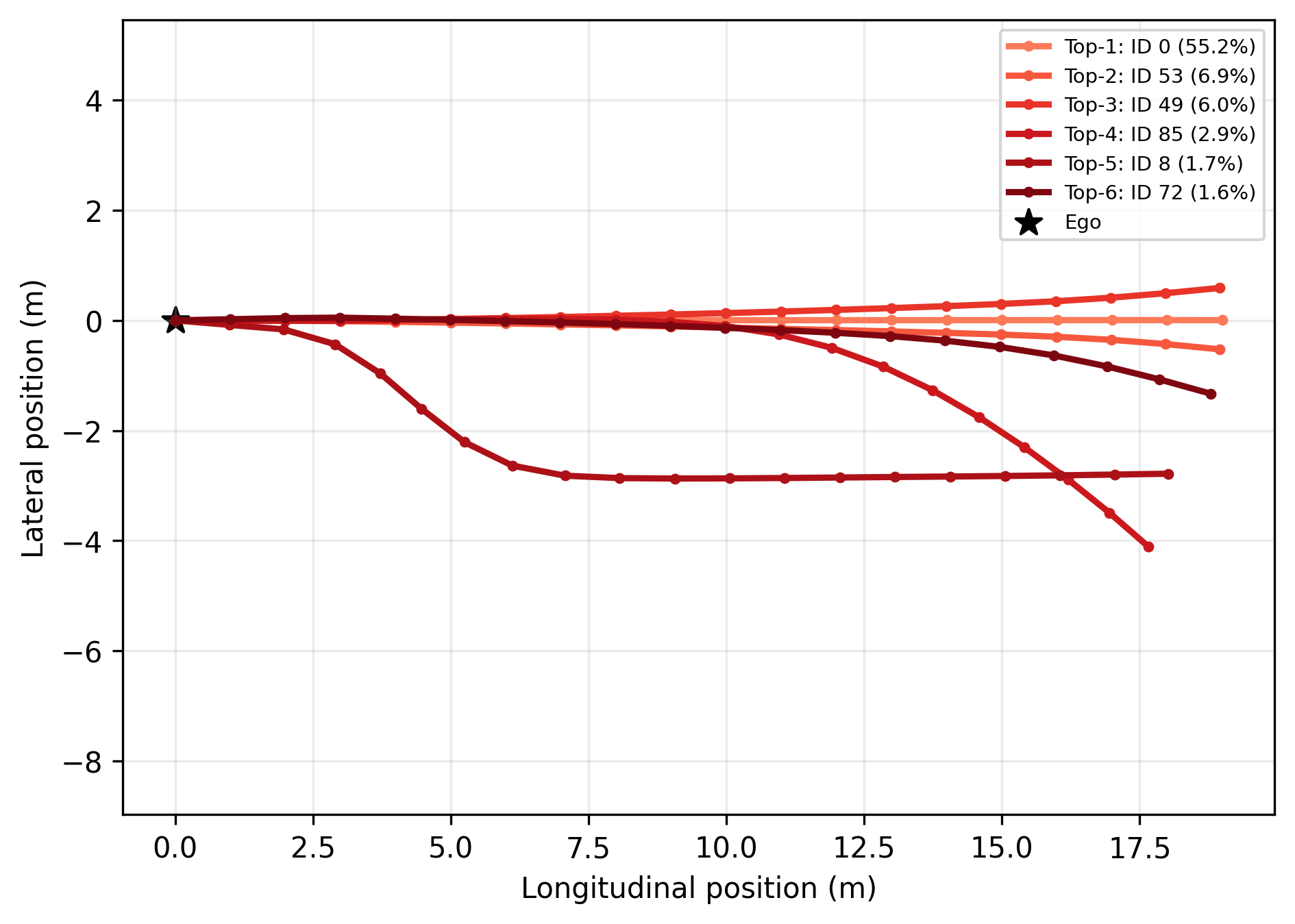}
        \\[0.5mm]

        \anchorrowlabel{Slower}
        &
        \anchorpanel{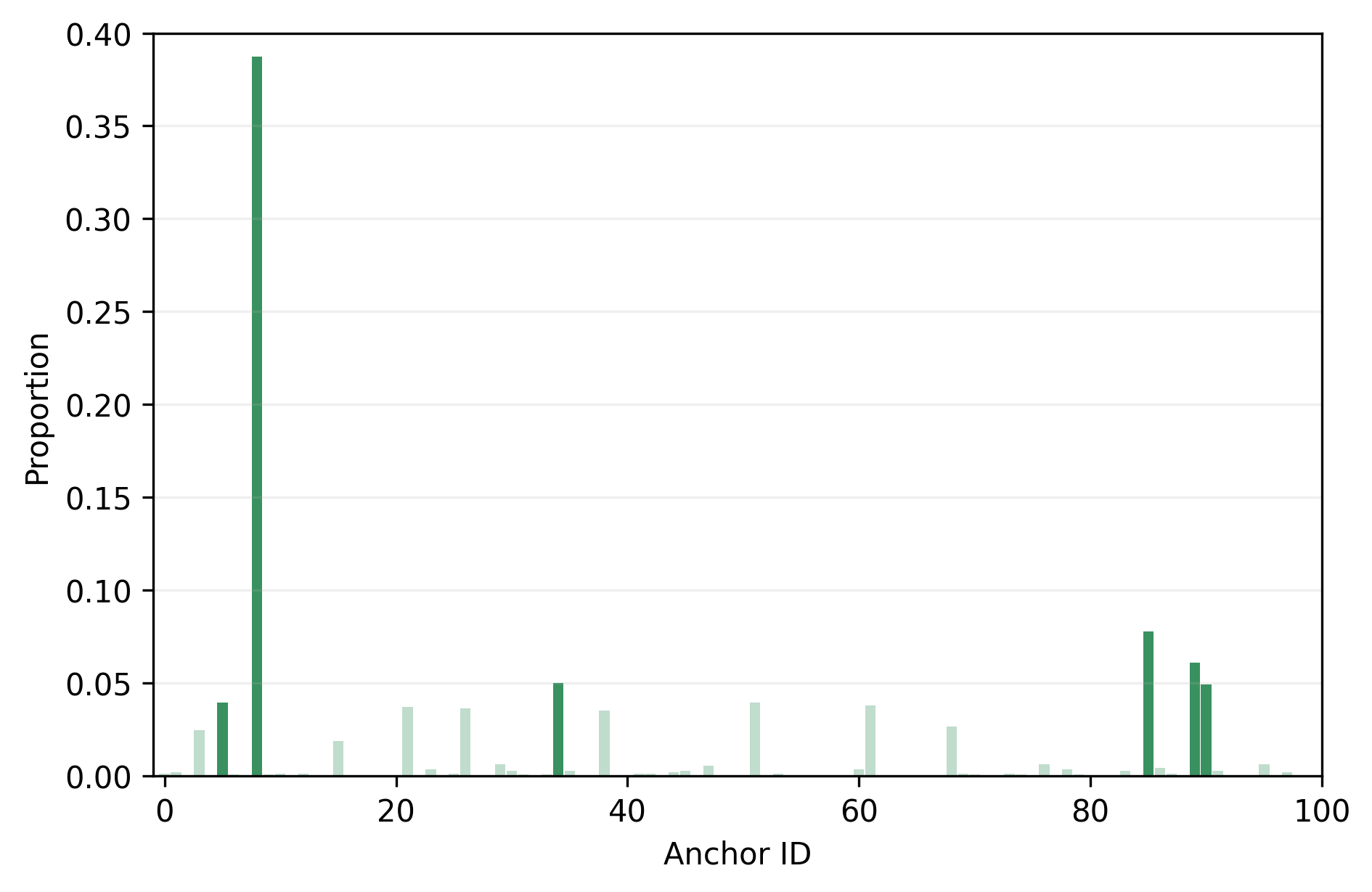}
        &
        \anchorpanel{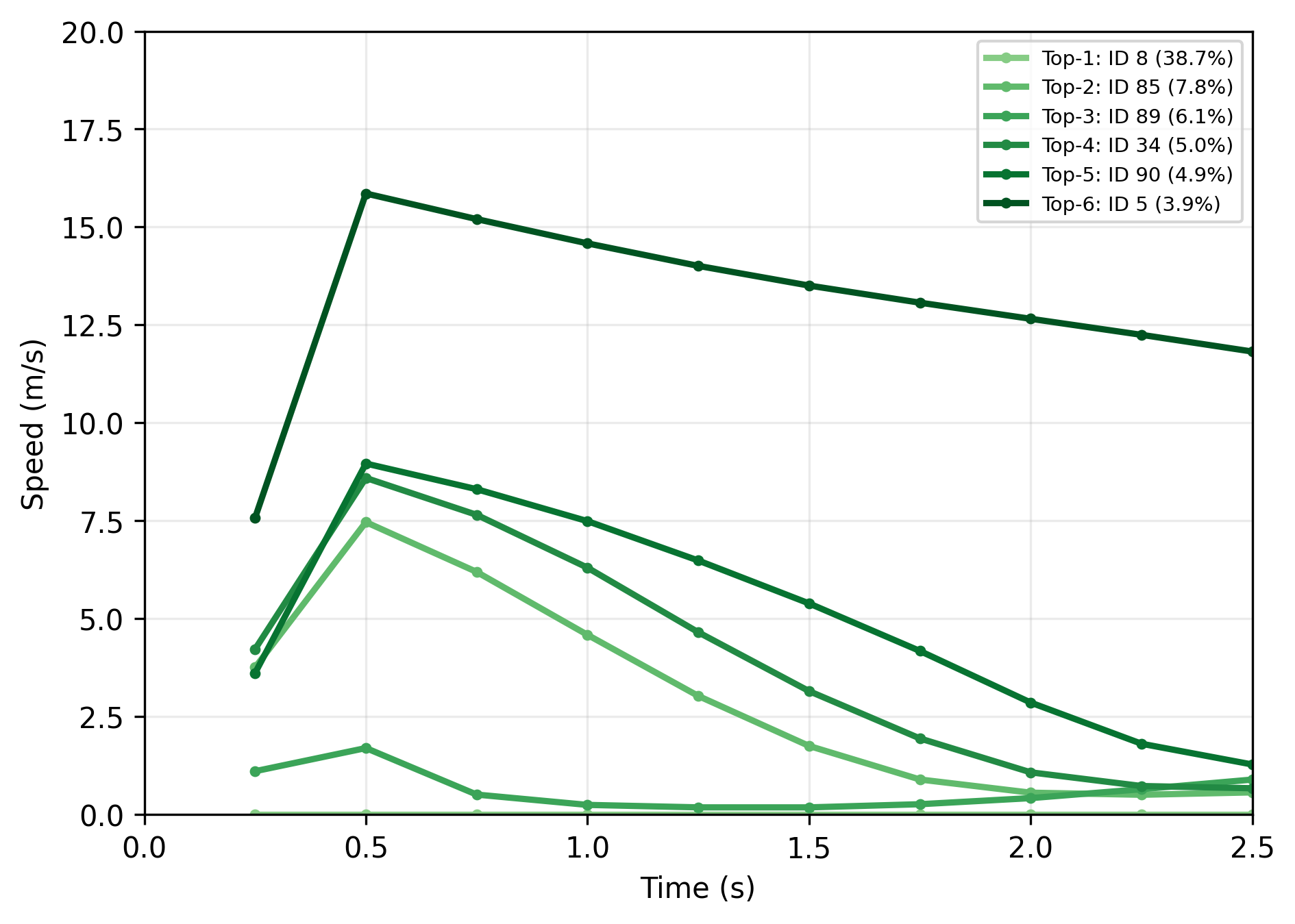}
        &
        \anchorpanel{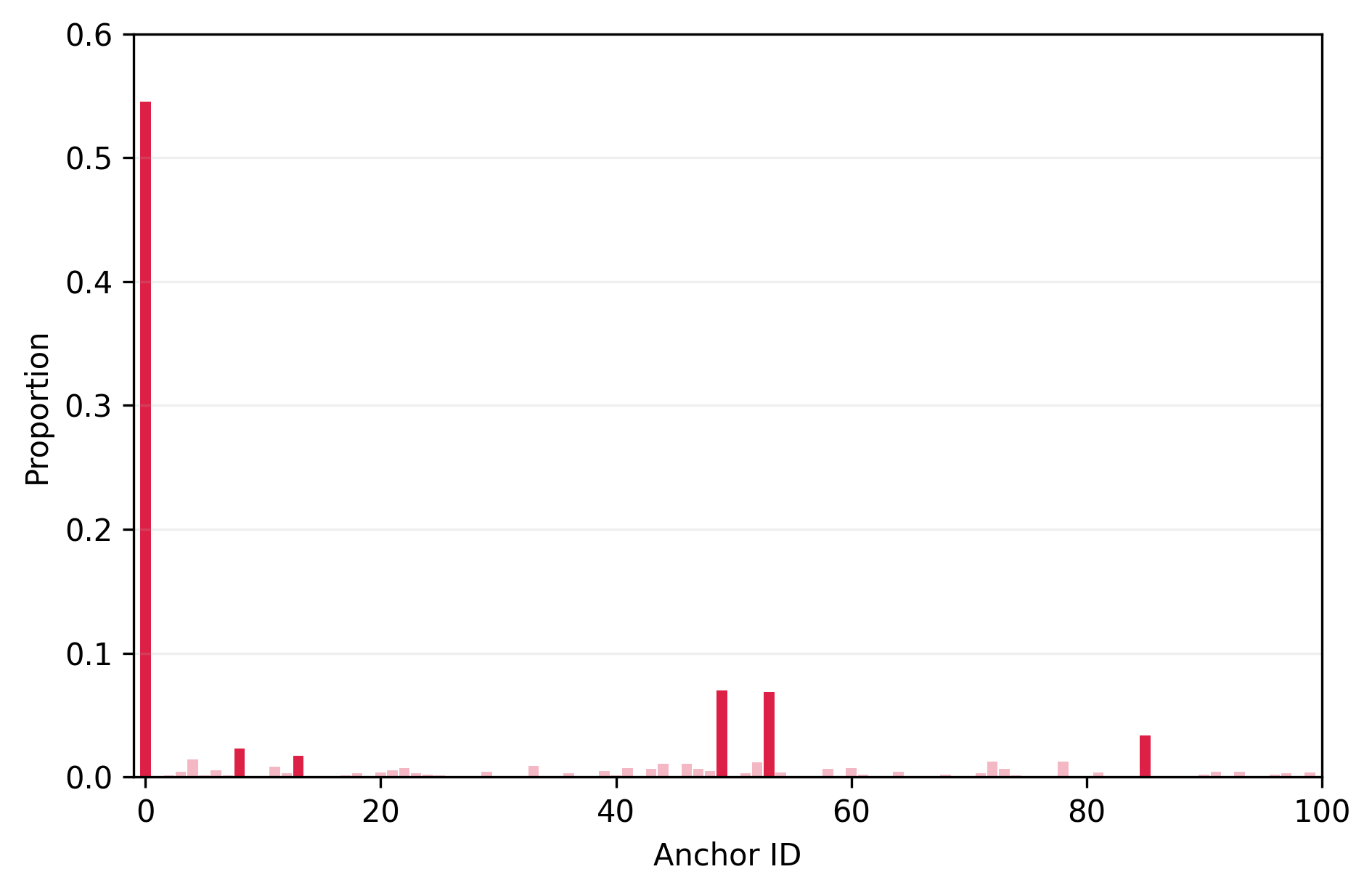}
        &
        \anchorpanel{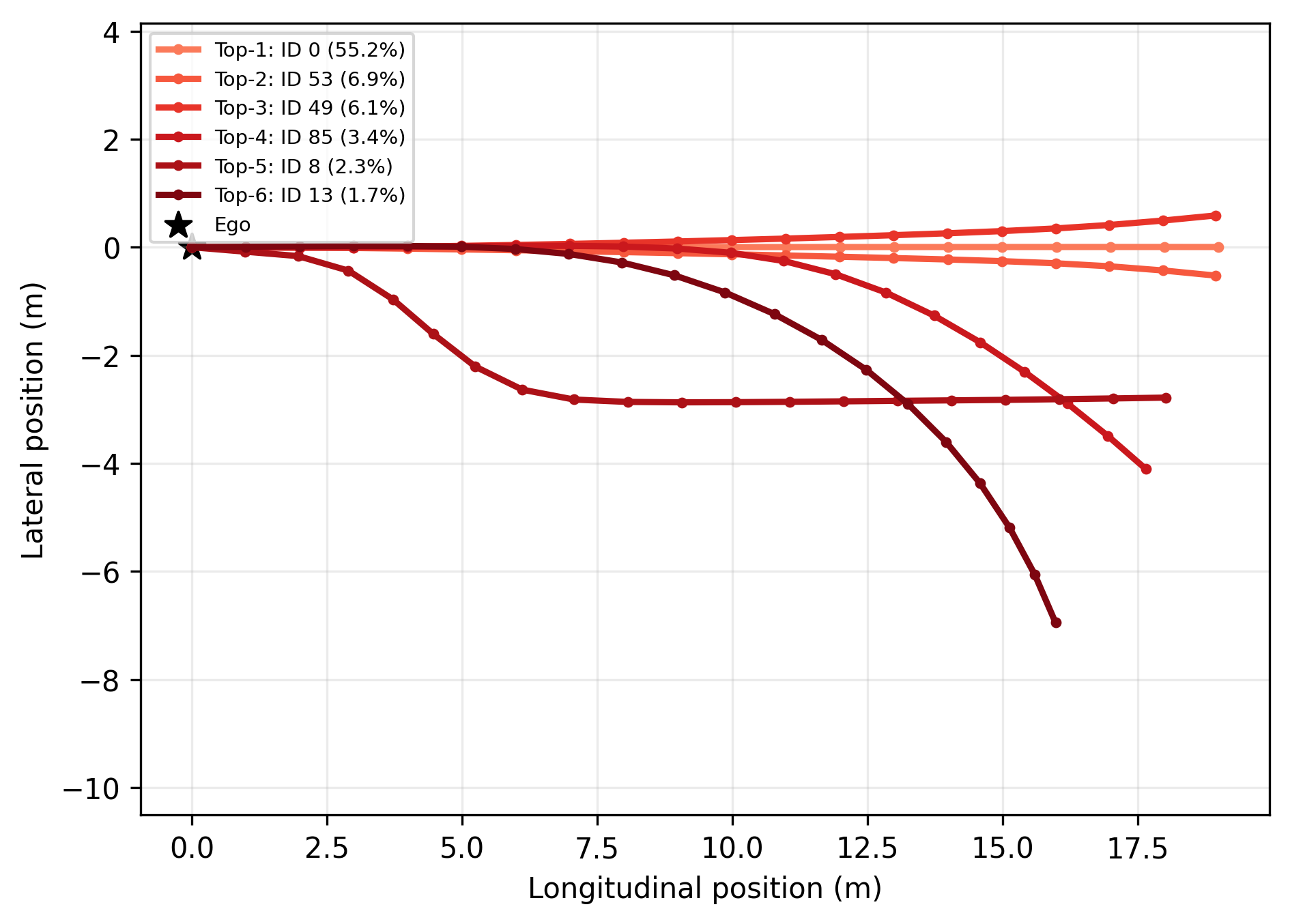}
        \\[0.5mm]

        \anchorrowlabel{Faster}
        &
        \anchorpanel{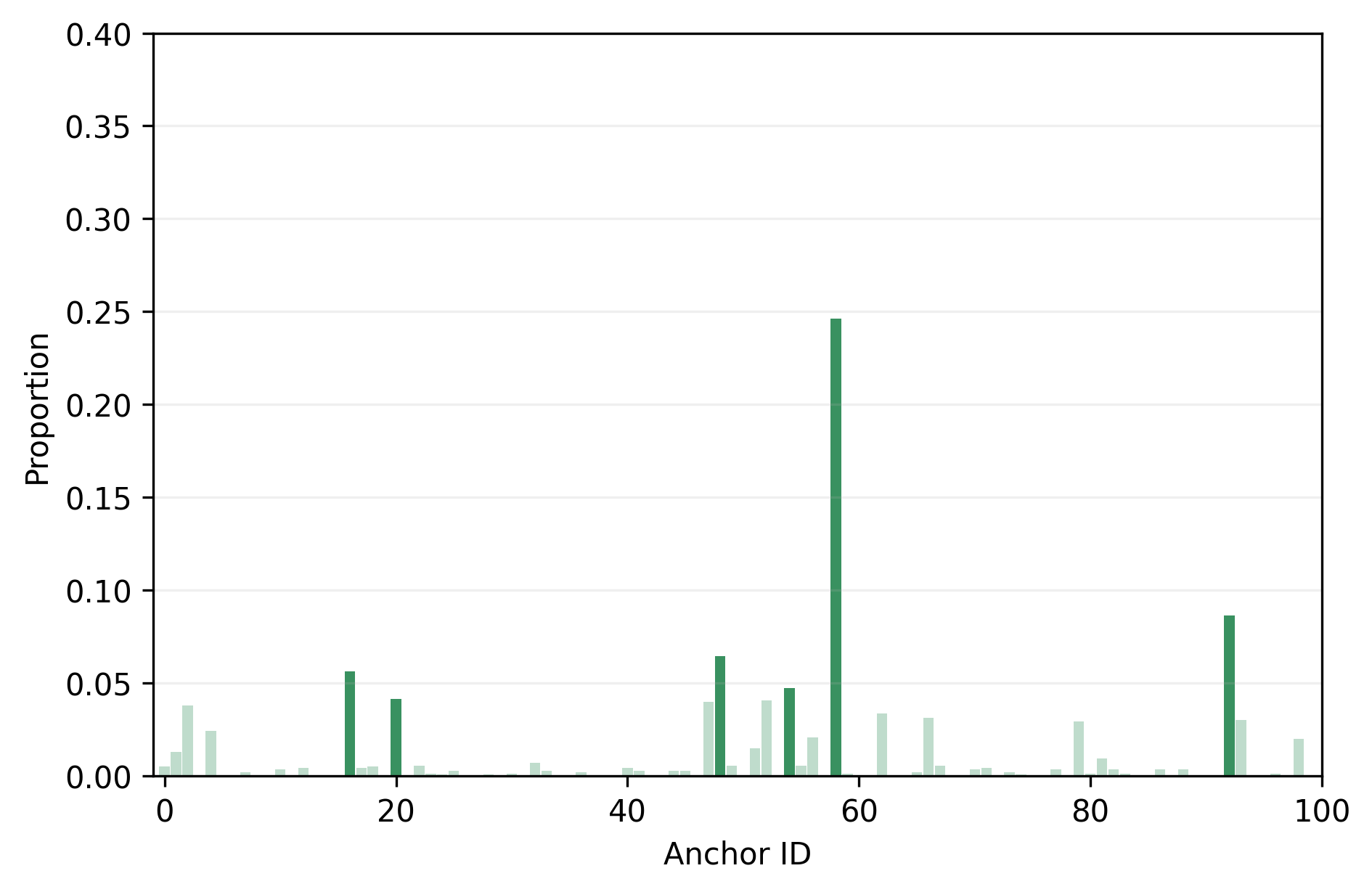}
        &
        \anchorpanel{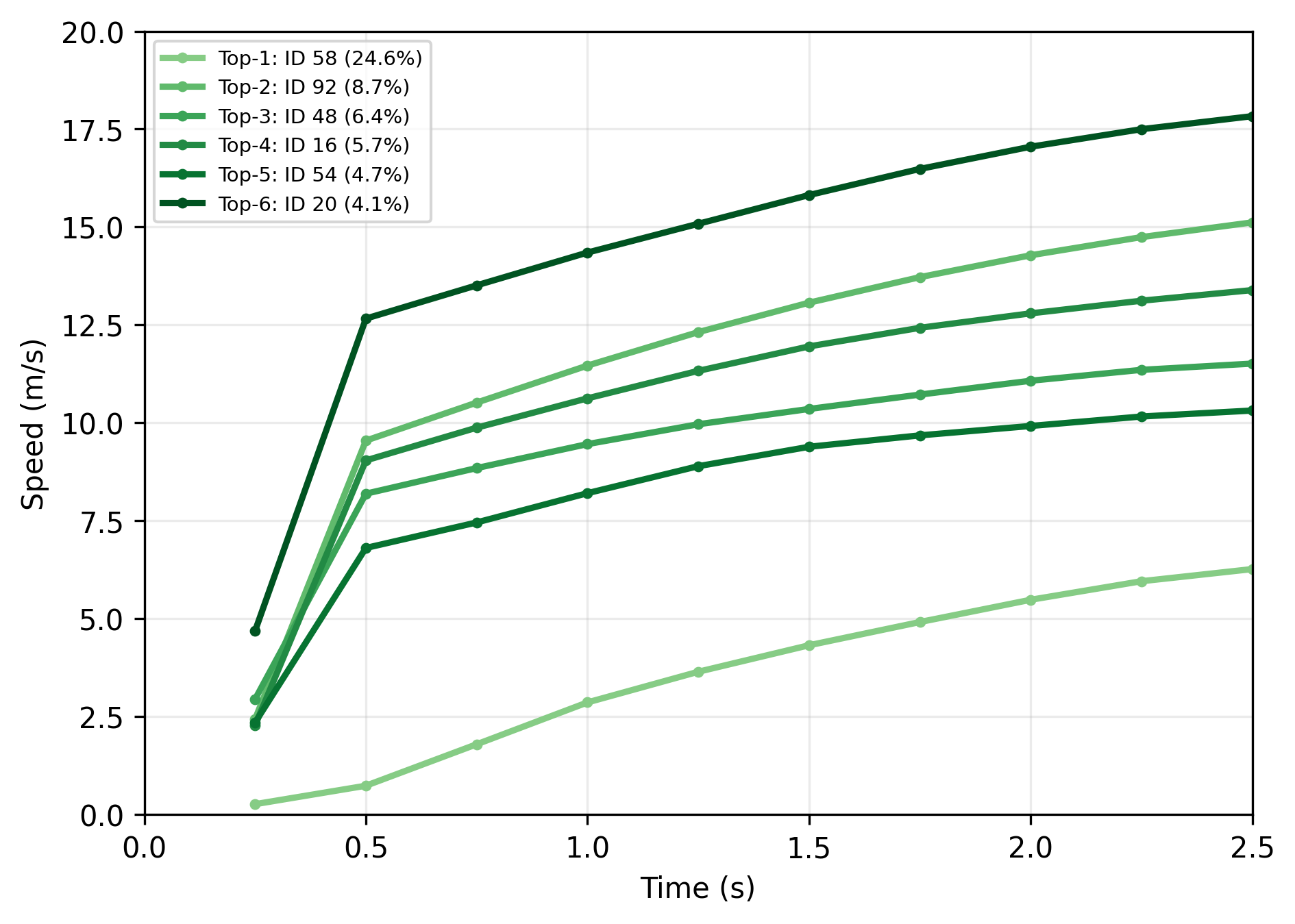}
        &
        \anchorpanel{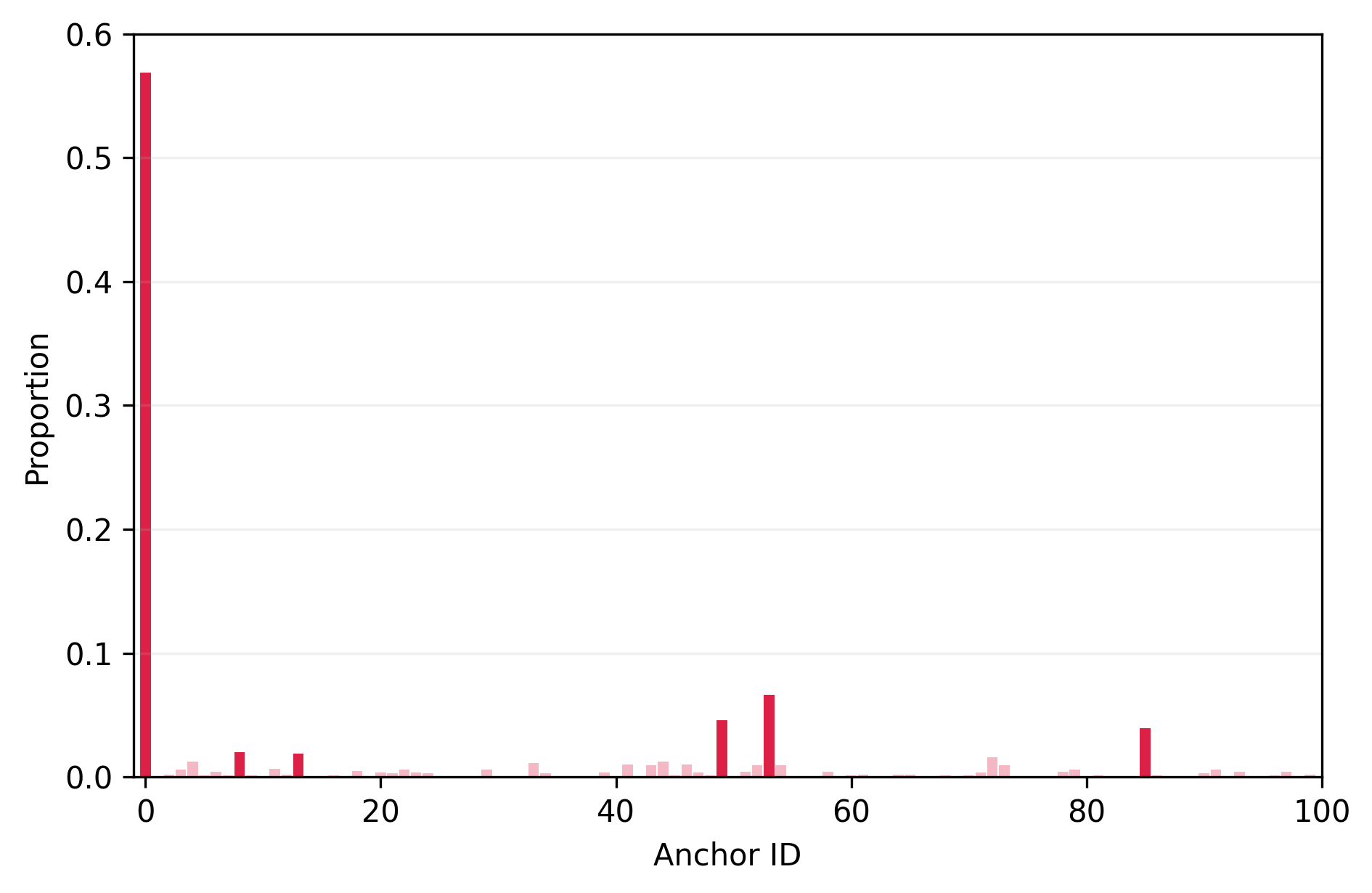}
        &
        \anchorpanel{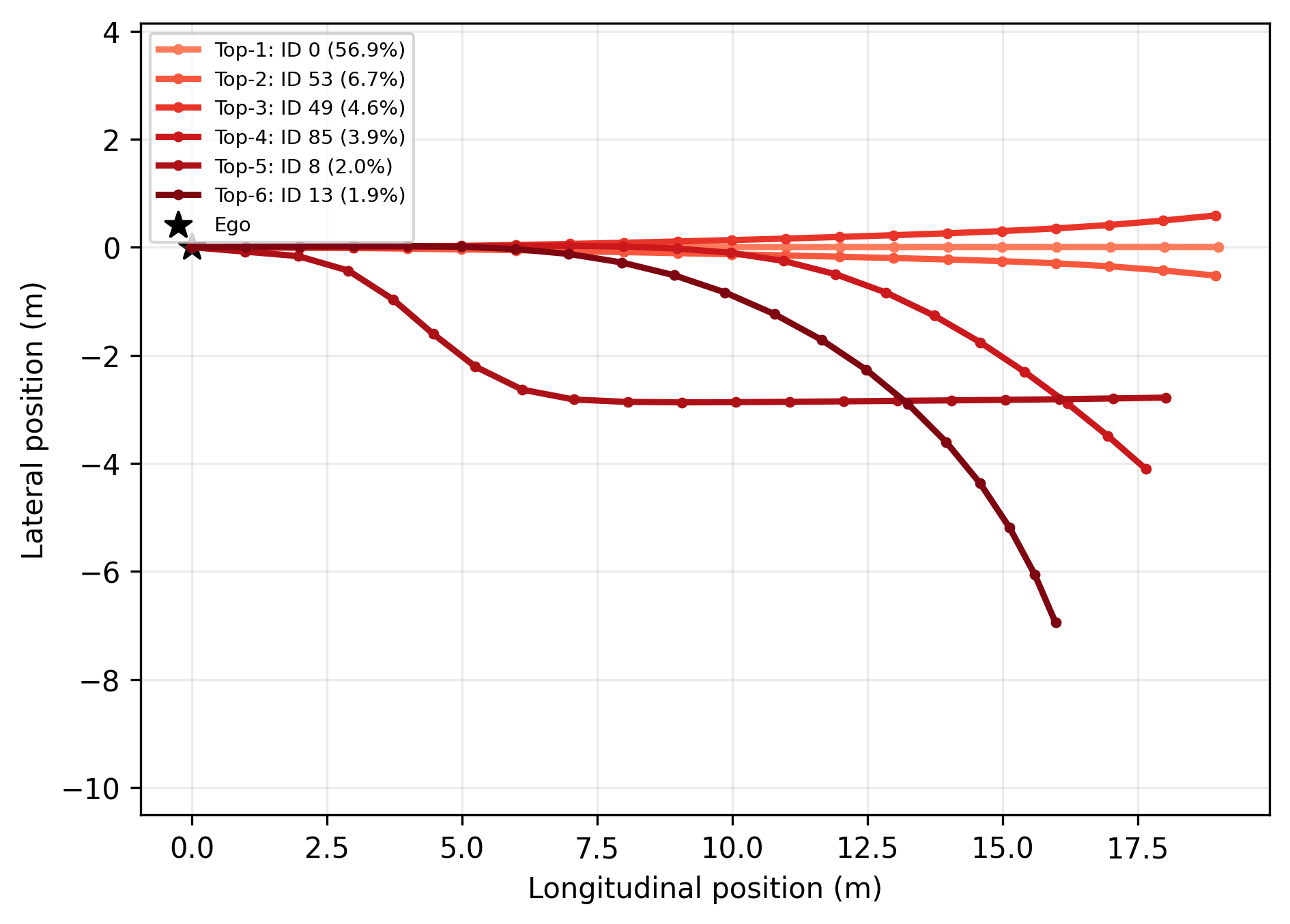}
        \\[0.5mm]

        \anchorrowlabel{Lane Change}
        &
        \anchorpanel{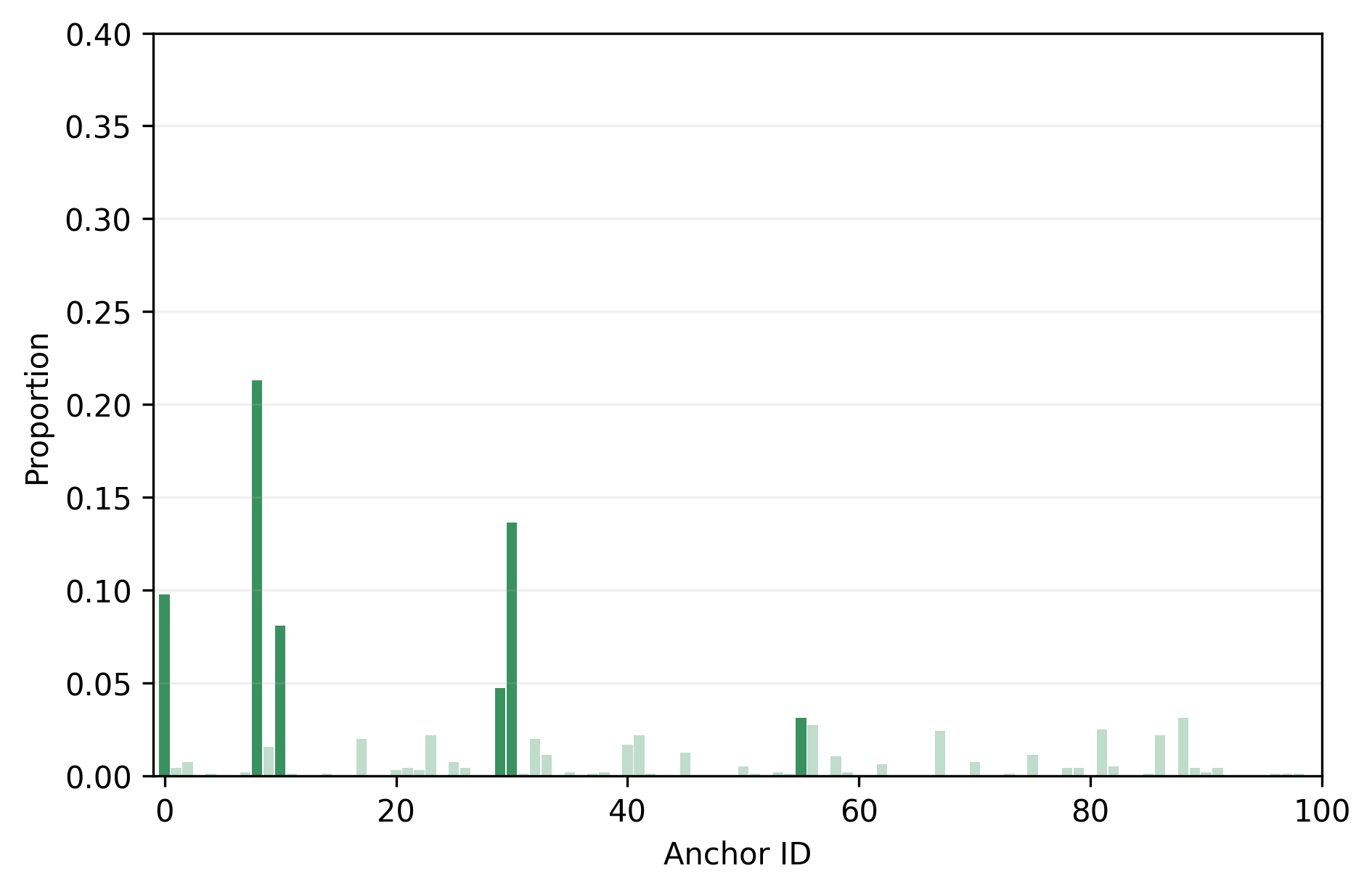}
        &
        \anchorpanel{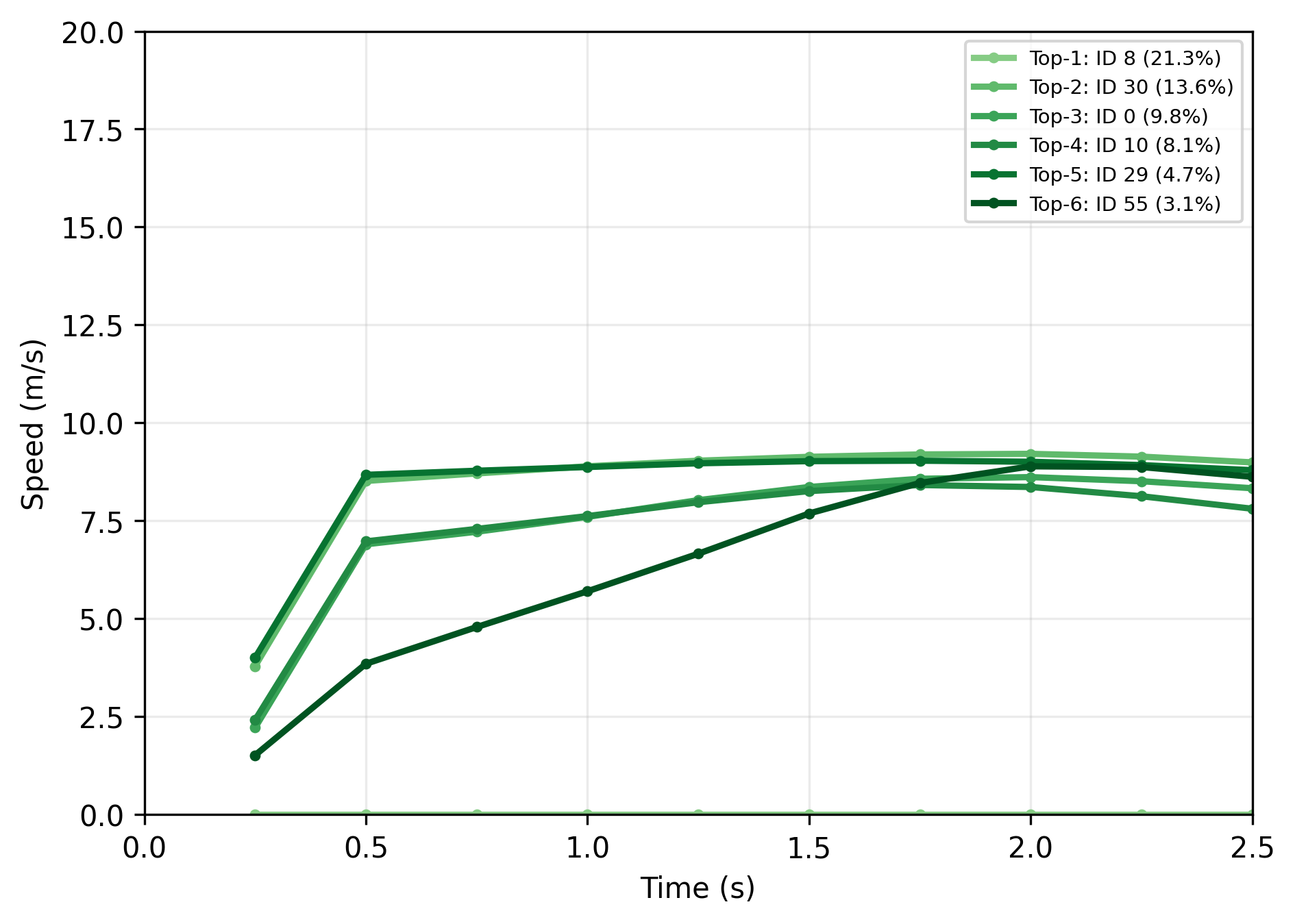}
        &
        \anchorpanel{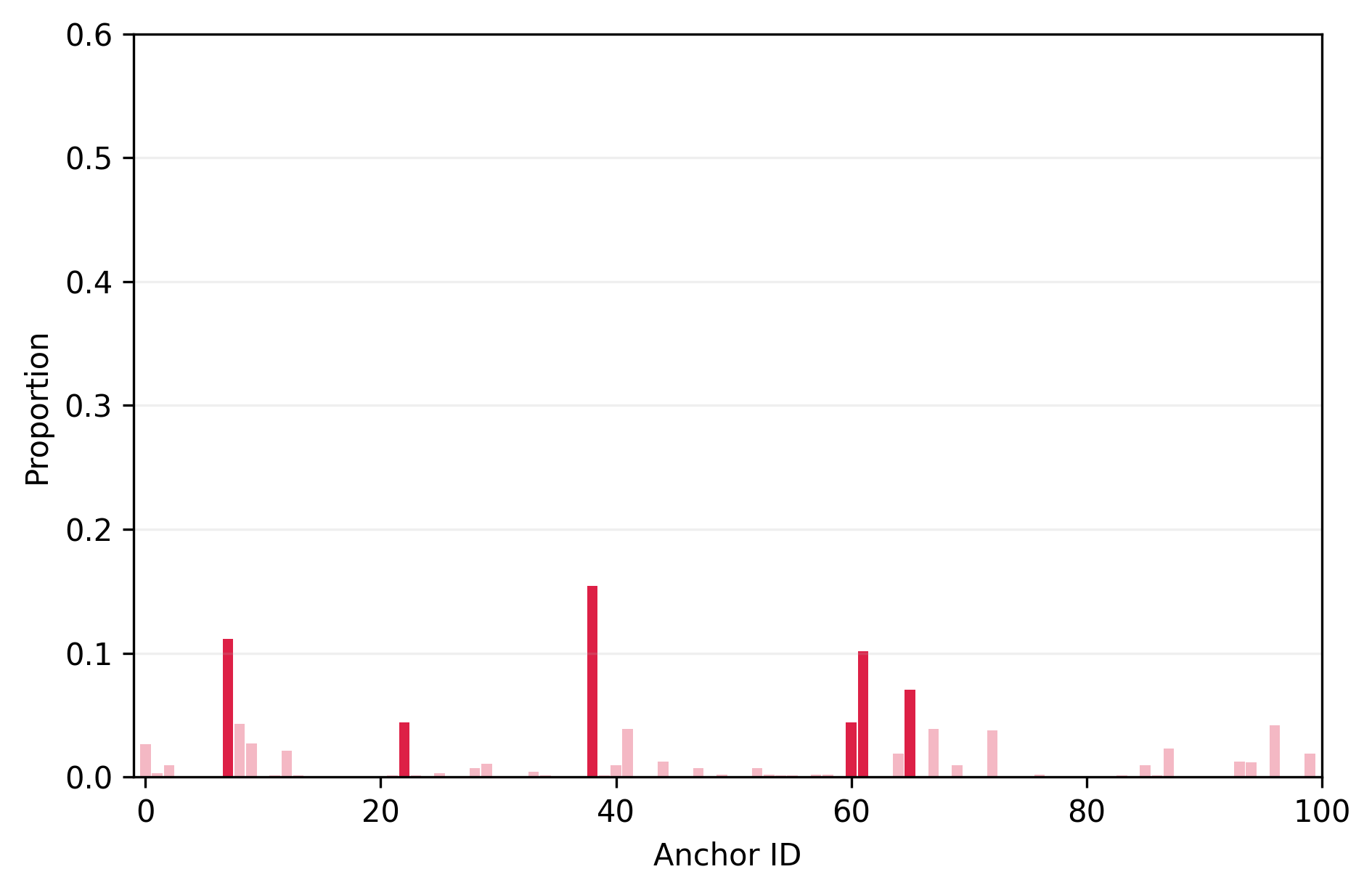}
        &
        \anchorpanel{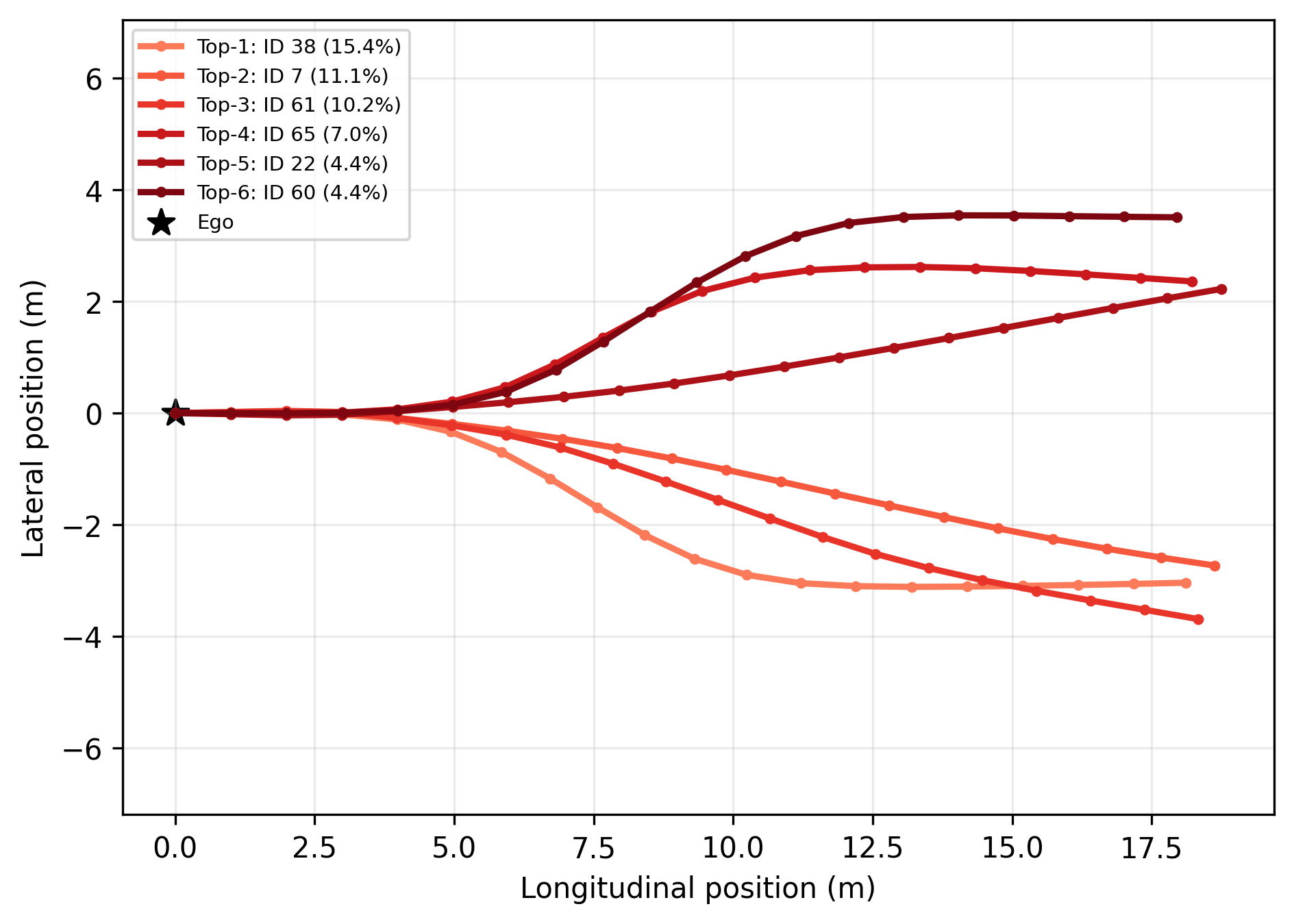}
        \\
    \end{tabular}

    \endgroup

    \caption{
        Instruction-conditioned distributions and representative patterns of
        predicted trajectory anchors on the Action Dreaming dataset.
        The histograms show normalized top-1 selection frequencies, while the
        adjacent panels visualize the six most frequently selected Velocity and
        Direction anchors as speed profiles and spatial paths, respectively.
    }
    \label{fig:instruction_conditioned_anchor_analysis}
\end{figure*}

\section{Analysis of Trajectory-Pattern Anchors as Decision Interfaces}
\label{sec:anchor_decision_analysis}

We analyze whether the predicted trajectory-pattern anchors serve as explicit
decision interfaces between high-level VLA reasoning and continuous trajectory
execution. We examine this role in two complementary settings: instruction-conditioned
analysis on the Action Dreaming dataset and decision-to-execution consistency
analysis during actual Bench2Drive closed-loop rollouts.

\subsection{Instruction-Conditioned Decision Sensitivity}
\label{subsec:instruction_conditioned_anchors}

We first examine whether anchor prediction responds systematically to high-level
driving commands. We consider four representative commands from the Action
Dreaming dataset: \emph{stop}, \emph{slower}, \emph{faster}, and
\emph{lane change}. Only samples labeled as
\texttt{<INSTRUCTION\_FOLLOWING>} are included, thereby excluding cases in
which safety constraints override the original instruction. After filtering,
the \emph{stop}, \emph{slower}, \emph{faster}, and \emph{lane change} subsets
contain 1,187, 1,423, 1,398, and 953 samples, respectively, resulting in a
total of 4,961 samples.

For each command, we compute the normalized distribution of the top-1 anchor
selected by DAAR. We additionally visualize the six anchors most frequently
selected as top-1 predictions under each command, together with their selection
proportions. Velocity anchors are converted into speed profiles over the
$2.5\,\mathrm{s}$ prediction horizon, whereas Direction anchors are visualized
as spatial paths.

As shown in Figure~\ref{fig:instruction_conditioned_anchor_analysis}, the
predicted anchor distributions exhibit clear instruction-conditioned and
component-specific patterns. The \emph{stop}, \emph{slower}, and
\emph{faster} commands primarily alter the Velocity-anchor distribution. Their
dominant anchors correspond to distinct speed-evolution patterns, including
stopping, deceleration, and acceleration. Meanwhile, the Direction-anchor
distributions remain comparatively stable across these longitudinal commands,
as the commands do not directly modify the desired path geometry.

In contrast, the \emph{lane change} command activates a substantially different
subset of Direction anchors. Its most frequently selected anchors exhibit
pronounced lateral motion patterns, while the corresponding Velocity anchors
describe the temporal speed evolution during the maneuver. This
component-specific response shows that anchor prediction is sensitive to the
behavioral content of the high-level command: longitudinal decisions are
primarily reflected in Velocity anchors, whereas lateral decisions are
primarily reflected in Direction anchors. These results support the role of
trajectory-pattern anchors in receiving and representing high-level VLA
decisions.

\subsection{Closed-Loop Decision-to-Execution Consistency}
\label{subsec:closed_loop_decision_consistency}

We next examine whether this decision-interface property is preserved during
actual Bench2Drive closed-loop execution. Unlike the Action Dreaming setting,
where an explicit behavior command is provided, the VLA in Bench2Drive derives
its driving decision from the current scene observation and navigation target
through chain-of-thought reasoning. This setting therefore evaluates whether
the predicted anchors remain consistent with decisions autonomously produced
by the VLA and whether these decisions are subsequently carried into continuous
trajectory generation.

\begin{figure*}[t]
    \centering
    \includegraphics[width=\textwidth]{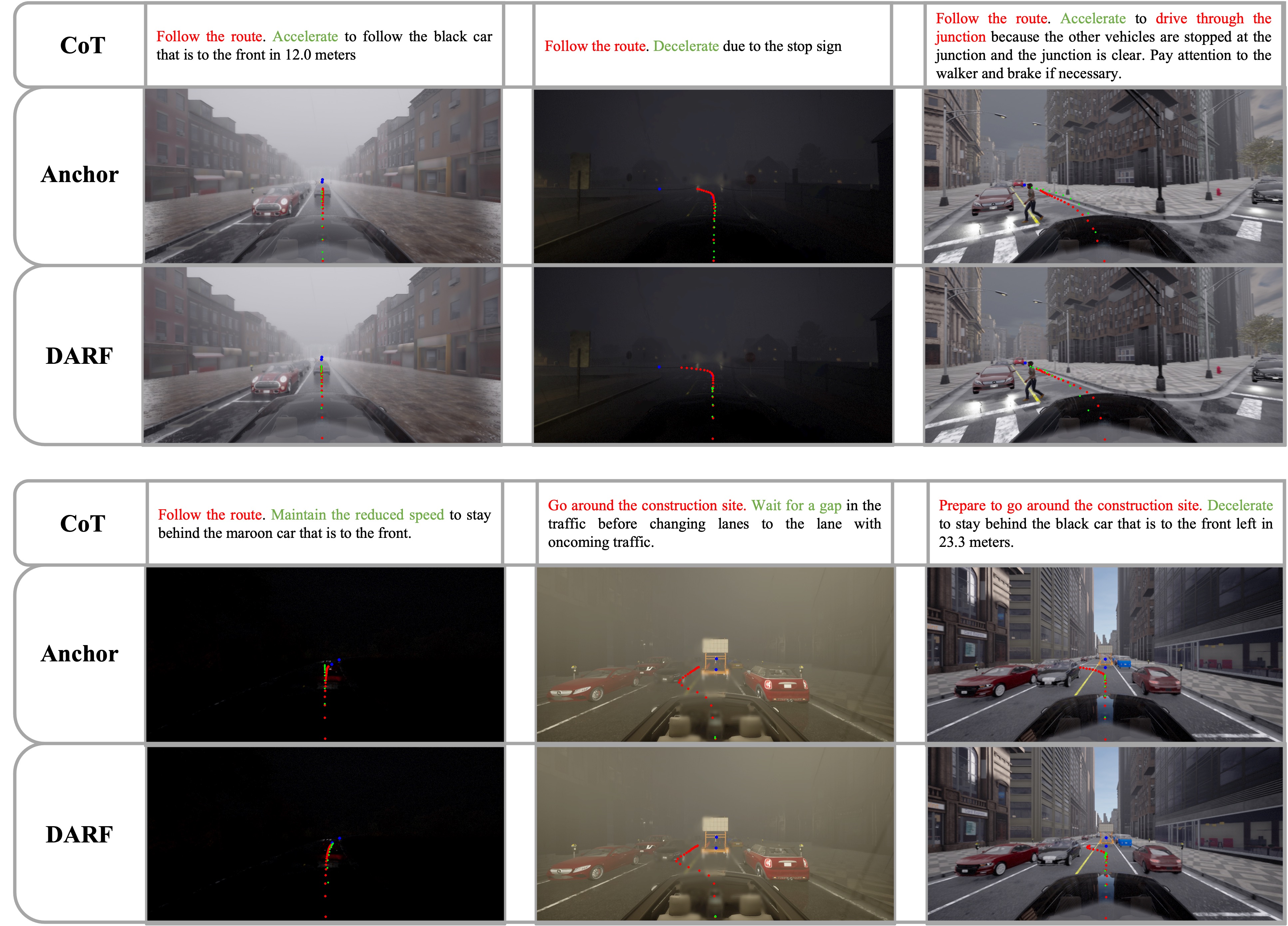}
    \caption{
    Closed-loop decision-to-execution visualization of AnchorVLA on
    Bench2Drive. Each case presents the CoT-based driving decision, the
    Direction and Velocity anchors predicted by the anchor-decision model, and
    the corresponding continuous planning outputs refined by
    \textbf{Decision-Anchored Residual Flow} (DARF). The predicted anchors are
    consistent with the behavior-level decisions produced by the VLA and
    determine the coarse motion patterns, while DARF introduces
    context-dependent refinements within the corresponding anchor-defined
    residual spaces.
    }
    \label{fig:qualitative_visualization}
\end{figure*}

Figure~\ref{fig:qualitative_visualization} presents representative closed-loop
examples of the CoT-based driving decisions, the predicted Direction and
Velocity anchors, and the continuous planning outputs refined by DARF. The
selected anchors capture the coarse spatial and temporal motion patterns
associated with the corresponding VLA decisions, while DARF introduces
context-dependent continuous refinements within the anchor-defined residual
spaces. Importantly, the refined outputs preserve the behavior-level patterns
represented by the selected anchors, illustrating how the anchor transfers the
high-level decision into continuous execution.

The two analyses provide complementary evidence for anchors as decision
interfaces. The instruction-conditioned analysis shows that anchor
distributions respond systematically to externally specified high-level
commands. The closed-loop analysis further shows that anchors align with
decisions autonomously generated by the VLA from scene observations and
navigation targets, and that these anchor-level decisions are preserved in the
continuous outputs produced by DARF. Together, they support the complete
decision-to-execution pathway of AnchorVLA:
high-level command or CoT decision $\rightarrow$ trajectory-pattern anchor
$\rightarrow$ continuous trajectory refinement.

\subsection{Failure Case Analysis}
\label{subsec:failure_cases}

\begin{figure*}[t]
    \centering
    \includegraphics[width=\textwidth]{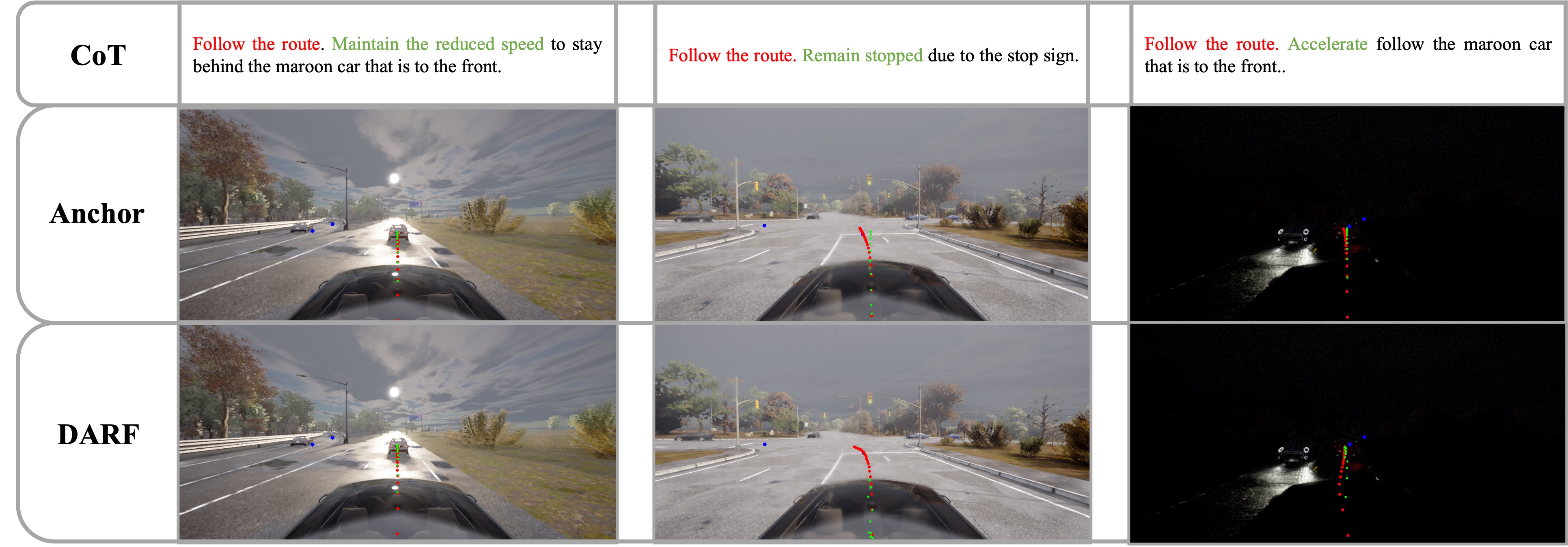}
    \caption{
    Representative failure cases from the Bench2Drive closed-loop evaluation.
    The first two cases illustrate anchor-level decision errors, including an
    off-route anchor and an anchor inconsistent with the VLA decision. The
    third case illustrates a residual-level execution error, where inaccurate
    DARF refinement introduces local geometric deviations despite a reasonable
    anchor. An incorrect anchor propagates an undesired coarse motion pattern
    to the final output, whereas an imperfect residual may affect local
    execution despite a correct high-level decision.
    }
    \label{fig:failure_cases}
\end{figure*}

We further analyze representative failure cases from the Bench2Drive
closed-loop evaluation. As shown in Figure~\ref{fig:failure_cases}, the first
two cases correspond to anchor-level decision errors, including an off-route
anchor prediction and an anchor that is inconsistent with the VLA decision.
These incorrect anchors propagate undesired coarse motion patterns to the final
planning outputs. The third case corresponds to a residual-level execution
error, where inaccurate DARF refinement introduces local geometric deviations
despite a reasonable anchor-level decision. These examples reveal the two
principal failure sources of the proposed decision-to-execution decomposition:
high-level anchor selection and continuous residual refinement.

\section{Quantitative Analysis of the Trajectory-Pattern Codebook and DARF}
\label{sec:quantitative_analysis}

Based on the disentangled path-and-speed representation described above, we provide two
complementary quantitative analyses. First, we study how the
trajectory-pattern codebook size affects anchor coverage, geometric
representation, and closed-loop driving performance. Second, we analyze how
DARF refines the predicted decision anchors and compare the transport lengths
of full-trajectory and residual-flow generation. All open-loop metrics are
computed over 36,720 evaluation samples from the SimLingo evaluation dataset.

\subsection{Trajectory-Pattern Codebook Size}
\label{subsec:codebook_size}

We evaluate trajectory-pattern codebooks with
$K\in\{50,100,200\}$ for both the Direction and Velocity output components while
keeping the number of predicted candidate anchors fixed at $M=6$.
Recall@$m$ measures whether the globally nearest codebook anchor is included
in the predicted top-$m$ candidates. minADE@$m$ measures the minimum trajectory
displacement error between the ground-truth trajectory and the predicted
top-$m$ anchors.

\begin{table*}[t]
    \centering
    \renewcommand{\arraystretch}{1.15}
    \setlength{\tabcolsep}{7pt}

    \resizebox{\textwidth}{!}{
    \begin{tabular}{cc|ccc|ccc|cc}
        \toprule
        \makecell[c]{Codebook\\Size}
        & \makecell[c]{Output\\Component}
        & \makecell[c]{Recall@1\\(\%)}
        & \makecell[c]{Recall@3\\(\%)}
        & \makecell[c]{Recall@6\\(\%)}
        & \makecell[c]{minADE\\@1}
        & \makecell[c]{minADE\\@3}
        & \makecell[c]{minADE\\@6}
        & \makecell[c]{Driving\\Score}
        & \makecell[c]{Success\\Rate}
        \\
        \midrule

        50
        & Direction
        & \textbf{92.80}
        & \textbf{98.32}
        & \textbf{99.35}
        & 0.2042
        & 0.1824
        & 0.1758
        & \multirow{2}{*}{\textbf{90.43}}
        & \multirow{2}{*}{74.43}
        \\
        50
        & Velocity
        & \textbf{80.92}
        & \textbf{96.70}
        & \textbf{99.33}
        & 0.9142
        & 0.7654
        & 0.7418
        &
        &
        \\
        \midrule

        100
        & Direction
        & 90.96
        & 97.47
        & 98.78
        & 0.1600
        & 0.1407
        & 0.1335
        & \multirow{2}{*}{89.92}
        & \multirow{2}{*}{\textbf{77.28}}
        \\
        100
        & Velocity
        & 79.04
        & 95.35
        & 98.92
        & 0.7548
        & 0.6250
        & 0.5956
        &
        &
        \\
        \midrule

        200
        & Direction
        & 88.06
        & 96.49
        & 98.09
        & \textbf{0.1365}
        & \textbf{0.1177}
        & \textbf{0.1096}
        & \multirow{2}{*}{89.51}
        & \multirow{2}{*}{73.64}
        \\
        200
        & Velocity
        & 73.35
        & 91.29
        & 96.96
        & \textbf{0.6928}
        & \textbf{0.5484}
        & \textbf{0.5035}
        &
        &
        \\
        \bottomrule
    \end{tabular}
    }
    \caption{
        Effect of the trajectory-pattern codebook size on open-loop anchor
        coverage and Bench2Drive closed-loop performance. The indicated
        codebook size is used for each output component. Component-specific
        codebooks are employed because the Direction and Velocity components
        have different sampling domains and prediction horizons. Recall and
        minADE are diagnostic open-loop metrics computed using logged
        ground-truth targets, while Driving Score and Success Rate are measured
        for the complete SimLingo planning output through closed-loop
        evaluation.
    }
    \label{tab:codebook_size_analysis}
\end{table*}

As shown in Table~\ref{tab:codebook_size_analysis}, increasing the codebook
size improves its geometric representation resolution. From $K=50$ to
$K=200$, minADE@$6$ decreases from 0.1758 to 0.1096 for the Direction
component and from 0.7418 to 0.5035 for the Velocity component. However,
finer discretization also makes anchor prediction more challenging.
Recall@$6$ decreases from 99.35\% to 98.09\% for Direction and from 99.33\%
to 96.96\% for Velocity.

The closed-loop results exhibit a non-monotonic relationship with codebook
size. The coarse $K=50$ codebook facilitates anchor prediction and achieves
the highest Driving Score of 90.43, but its limited motion-pattern resolution
results in a lower Success Rate of 74.43. Increasing the codebook size to
$K=100$ improves motion-pattern resolution while maintaining reliable
top-$M$ candidate coverage, yielding the highest Success Rate of 77.28.

When the codebook is further enlarged to $K=200$, its open-loop geometric
accuracy continues to improve, but its closed-loop Success Rate decreases to
73.64. The finer partition distributes the prediction probability over a
larger number of nearby and geometrically similar anchors, potentially
reducing the diversity of the top-$M$ candidate set and making closed-loop
anchor selection more difficult. Overall, these results support $K=100$, as
used in our main experiments, as a balanced setting between motion-pattern
resolution and reliable decision-anchor prediction.

\subsection{DARF Refinement and Transport Analysis}
\label{subsec:darf_transport}

We next analyze whether DARF performs continuous refinement around the
selected decision anchors and whether residual-flow generation reduces the
empirical transport length. For the predicted top-1 anchor
$a_{\hat{k}}$ and its DARF-refined trajectory $\hat{\tau}$, we define the
relative anchor-to-output deviation as
\begin{equation}
    \delta
    =
    \frac{\|\hat{\tau}-a_{\hat{k}}\|_2}
    {\|a_{\hat{k}}\|_2+\epsilon},
    \label{eq:anchor_output_deviation}
\end{equation}
where $\epsilon$ is a small constant for numerical stability. We report the
median relative deviation because near-stationary Velocity anchors may have
norms close to zero and produce disproportionately large relative values. We
further compare the RMS transport lengths of full-trajectory flow and
residual flow under the same noise distribution.

\begin{table*}[t]
    \centering

    \renewcommand{\arraystretch}{1.15}
    \setlength{\tabcolsep}{14pt}

    \begin{tabular}{cc|ccc}
        \toprule
        \makecell[c]{Codebook\\Size}
        & \makecell[c]{Output\\Component}
        & \makecell[c]{Median Relative\\
          Anchor-to-Output Deviation (\%)}
        & \makecell[c]{Full-Trajectory\\RMS Transport}
        & \makecell[c]{Residual-Flow\\RMS Transport}
        \\
        \midrule
        50
        & Direction
        & 1.41
        & 48.83
        & 1.88
        \\
        50
        & Velocity
        & 15.77
        & 30.16
        & 5.14
        \\
        \midrule
        100
        & Direction
        & 0.81
        & 48.82
        & 1.46
        \\
        100
        & Velocity
        & 12.58
        & 29.92
        & 4.08
        \\
        \midrule
        200
        & Direction
        & 0.62
        & 48.81
        & 1.19
        \\
        200
        & Velocity
        & 11.81
        & 29.83
        & 3.60
        \\
        \bottomrule
    \end{tabular}
    \caption{
        Quantitative analysis of DARF refinement and transport length using
        the predicted top-1 anchors. Relative Anchor-to-Output Deviation
        measures the magnitude of DARF refinement relative to the selected
        anchor and is reported as the median percentage. Full-Trajectory and
        Residual-Flow RMS Transport measure the empirical transport lengths
        under the same noise distribution.
    }
    \label{tab:darf_quantitative_analysis}
\end{table*}

As shown in Table~\ref{tab:darf_quantitative_analysis}, DARF performs moderate
continuous refinement around the predicted decision anchors. For the Direction
component, the median relative anchor-to-output deviation ranges from 0.62\%
to 1.41\%. For the Velocity component, the median relative deviation ranges
from 11.81\% to 15.77\%. These results show that the refined outputs
preserve the coarse motion patterns represented by the anchors while DARF
adjusts their continuous execution according to the current driving context.

The transport results further support the residual-flow formulation. For the
Direction component, the RMS transport length is reduced from approximately
48.8 in full-trajectory space to 1.19--1.88 in the corresponding
anchor-defined residual spaces, representing a reduction of approximately
96.1--97.6\%. For the Velocity component, the RMS transport length is reduced
from approximately 30.0 to 3.60--5.14, representing a reduction of
approximately 83.0--87.9\%. These results demonstrate that trajectory-pattern
anchors provide compact references for residual generation and substantially
shorten the empirical transport path.

\section{Details of Decision-Anchored Residual Flow}
\label{sec:darf_decoder}

For each SimLingo output component, given the top-$M$ trajectory-pattern
anchors predicted by DAAR, the \textbf{Decision-Anchored Residual Flow}
(DARF) decoder generates a continuous residual for each candidate anchor and
predicts an anchor--context compatibility score for candidate ranking.

As introduced above, the Direction component is represented by 20 geometric
path waypoints indexed by traveled distance, whereas the Velocity component
is represented by 10 temporal speed waypoints indexed by elapsed time. For
clarity, we describe the DARF decoder below for a generic output component
with $T$ points and omit the component superscript. The same formulation is
instantiated for both components.

For each candidate anchor $a_k$, $k\in\mathcal{I}_M$, the decoder receives the
noisy residual state $z_t^k\in\mathbb{R}^{T\times2}$, the flow timestep $t$,
the candidate anchor $a_k\in\mathbb{R}^{T\times2}$, and the multimodal context
features $H_x$ extracted from the visual, language, and navigation inputs.
All top-$M$ candidates within the same output component share the decoder
parameters. The Direction and Velocity components use the same decoder
architecture, with separate input and output projections to accommodate their
different numbers of points and sampling domains.

\paragraph{Input encoding.}
The noisy residual and candidate anchor are encoded independently. Each 2D
point coordinate is mapped to a sine--cosine embedding. The embeddings over
the $T$ points are then flattened and projected into the decoder hidden
dimension using MLPs. This produces the noisy-residual feature $h_z^k$ and
two anchor features, $h_{a,k}^{\mathrm{pos}}$ and
$h_{a,k}^{\mathrm{conf}}$, which are used for residual-flow prediction and
confidence prediction, respectively. The flow timestep is encoded using a
sinusoidal time embedding followed by an MLP, producing the time feature
$h_t$.

\paragraph{Residual-flow branch.}
The residual-flow branch initializes its hidden state with $h_z^k$. At each
decoder layer, the hidden state is conditioned on the flow timestep through
feature-wise affine modulation:
\begin{equation}
    \mathrm{Mod}(h,h_t)
    =
    h\odot\left(1+s(h_t)\right)+b(h_t),
    \label{eq:feature_modulation}
\end{equation}
where $s(\cdot)$ and $b(\cdot)$ are learned MLP projections. The anchor
feature is subsequently added as a query bias:
\begin{equation}
    q_l^k
    =
    \mathrm{Mod}(h_l^k,h_t)+h_{a,k}^{\mathrm{pos}}.
    \label{eq:anchor_query_bias}
\end{equation}
The query $q_l^k$ attends to the multimodal context $H_x$ through
cross-attention, followed by a residual connection, LayerNorm, and an FFN.
After the final decoder layer, the prediction head outputs the flow velocity
field:
\begin{equation}
    \hat{v}_{\theta}^{k}
    =
    \hat{v}_{\theta}(z_t^k,t,a_k,H_x)
    \in\mathbb{R}^{T\times2}.
    \label{eq:decoder_velocity_output}
\end{equation}
During training, the predicted flow velocity field is supervised by the flow
matching objective defined in the main paper. During inference, the velocity
field is integrated to generate the continuous residual in the corresponding
anchor-defined space. The residual is then added to the candidate anchor to
reconstruct the refined output.

\paragraph{Confidence prediction branch.}
The confidence branch assigns an anchor--context compatibility score to each
candidate proposed by DAAR. It adopts the WTA supervision defined in the main
paper, following the candidate-selection strategy used by
DiffusionDrive~\cite{liao2025diffusiondrive}.

Because the confidence score depends on the candidate anchor and the driving
context, this branch does not use the noisy residual state $z_t^k$ or the flow
timestep $t$. It initializes its hidden state with
$h_{a,k}^{\mathrm{conf}}$ and attends to the same multimodal context $H_x$.
The resulting feature is updated through residual connections, LayerNorm,
and FFNs. Finally, the confidence head predicts a scalar score for each
candidate:
\begin{equation}
    s_\phi^k
    =
    s_\phi(a_k,H_x)
    \in\mathbb{R}.
    \label{eq:decoder_confidence_output}
\end{equation}

At inference time, candidates are ranked within each output component
according to the predicted confidence scores. Both components are refined
and scored using the same multimodal driving context. The selected Direction
output provides the geometric path used by the lateral controller to
determine steering, whereas the temporal displacements of the selected
Velocity waypoints are converted into the desired-speed signal used for
throttle and braking control, following the original SimLingo execution
interface.

The residual-flow and confidence branches serve complementary roles. The
residual-flow branch generates continuous refinements in the anchor-defined
residual spaces, while the confidence branch ranks the candidates associated
with the anchors proposed by DAAR.

\paragraph{Decoder refinement setting.}
For computational alignment, the deterministic residual baseline uses a
single residual-prediction step with two successive decoder passes. Full Flow
and DARF use two flow-integration steps, with one decoder pass at each step.
Thus, all variants perform two decoder passes per output component.

\subsection{Additional Hyperparameter Settings}

Table~\ref{tab:additional_hyperparameters} presents additional hyperparameters
used in our experiments.

\begin{table}[t]
\centering
\small
\setlength{\tabcolsep}{4pt}
\caption{Additional hyperparameter settings used in experiments.}
\label{tab:additional_hyperparameters}
\begin{tabular}{p{0.49\columnwidth}p{0.41\columnwidth}}
\hline
\textbf{Hyperparameter} & \textbf{Setting} \\
\hline
Optimizer & AdamW \\
Learning rate & $3\times10^{-5}$ \\
Adam coefficients & $\beta_1=0.9,\ \beta_2=0.999$ \\
Weight decay & 0.1 \\
Learning-rate schedule & OneCycleLR \\
Warm-up fraction & 0.05 \\
Gradient clipping & Global norm of 0.3 \\
Numerical precision & FP16 mixed precision \\
Distributed optimization & DeepSpeed ZeRO Stage 2 \\
FP16 loss scale & 32 \\
LoRA rank & 32 \\
LoRA $\alpha$ & 64 \\
LoRA dropout & 0.1 \\
Image augmentation probability & 0.5 \\
Image-shift augmentation probability & 0.5 \\
Direction soft-label temperature & 0.2 \\
Velocity soft-label temperature & 0.5 \\
Noise schedule & Cosine \\
Training-time discretization & 1,000 \\
\hline
\end{tabular}
\end{table}

\bibliography{aaai2027}
